\documentclass{article}
\usepackage{cite}
\usepackage{amsmath,amssymb,amsfonts, mathtools}
\usepackage{booktabs} 

\usepackage{amsthm}

\newtheorem{statement}{Statement}[subsection]

\newtheorem{theorem}[statement]{Theorem}
\newtheorem{lemma}[statement]{Lemma}
\newtheorem{corollary}[statement]{Corollary}

\newtheorem{proposition}[statement]{Proposition}

\theoremstyle{definition}
\newtheorem{definition}[statement]{Definition}

\theoremstyle{remark}
\newtheorem*{remark}{Remark}          
\newtheorem{remarkn}[statement]{Remark}

\usepackage{graphicx}
\usepackage{textcomp}
\usepackage[dvipsnames]{xcolor}
\usepackage[margin=1in]{geometry}
\usepackage[most]{tcolorbox}

\newtcolorbox{brainbox}{
    colback=gray!10, colframe=black, fonttitle=\bfseries,
    title=What is the brain like?
}

 \usepackage{booktabs}
 \usepackage{array}
 \usepackage{makecell}   
 \usepackage{ragged2e}   
 \usepackage{tabularx}
 \newcolumntype{Y}{>{\centering\arraybackslash}X}
 \newcolumntype{C}[1]{>{\Centering\arraybackslash}p{#1}}

\title{A Theoretical Framework for Time, Space, and Energy Scaling of Neuromorphic Algorithms}
\author{James B Aimone \\
Neural Exploration \& Research Laboratory\\
Sandia National Laboratories\\
Albuquerque, NM 87185\\
jbaimon@sandia.gov}
\date{Revised March 2026}

\begin{document}

\maketitle

\begin{abstract}
Neuromorphic computing (NMC) is increasingly viewed as a low-power alternative to conventional von Neumann architectures such as central processing units (CPUs) and graphics processing units (GPUs), however the computational value proposition has been difficult to define precisely.

Here, we propose a computational framework for analyzing NMC algorithms and architectures. Using this framework, we demonstrate that NMC can be analyzed as general-purpose and programmable even though it differs considerably from a conventional stored-program architecture. We show that the time and space scaling of idealized NMC has comparable time and footprint tradeoffs that align with that of a theoretically infinite processor conventional system. In contrast, energy scaling for NMC is significantly different than conventional systems, as NMC energy costs are event-driven. Using this framework, we show that while energy in conventional systems is largely determined by the scheduled operations determined by the structural algorithm graph, the energy of neuromorphic systems scales with the activity of the algorithm, that is the activity trace of the algorithm graph. Without making strong assumptions on NMC or conventional costs, we demonstrate which neuromorphic algorithm formulations can exhibit asymptotically improved energy scaling when activity is sparse and decaying over time. We further use these results to identify which broad algorithm families are more or less suitable for NMC approaches. 
\end{abstract}

\section{Introduction}

Neuromorphic computing (`NMC') approaches to computation have been proposed for many years \cite{mead1988silicon, mead2002neuromorphic, von2012computer}, and today NMC hardware is increasingly available and can be implemented at scales approaching $10^9$ neurons in silicon  \cite{kudithipudi2025neuromorphic}, although still these systems remain far from the brain's complexity \cite{wang2025neuromorphic}. One implication of the success of implementing large-scale NMC systems is that it is increasingly evident that the biggest challenge facing the neuromorphic field, at least in the near-term, is the identification of which applications are well suited for its use. While scalable NMC platforms were originally motivated, in part, by large-scale brain simulations \cite{merolla2014million, furber2012build, ananthanarayanan2009cat, modha2010network}, the growing need for low-power solutions at both the edge and in high-performance computing systems has increased interest in NMC to address energy challenges in computation broadly. 

Key to their value proposition is that today's digital NMC systems are generically programmable---aside from some constraints of fan-in/fan-out, these systems can implement arbitrary computational graphs. Conceptually, this compatibility includes any algorithm that can be formulated as a threshold gate (TG) circuit, artificial neural network (ANNs), or any other more complex ensemble of neurons. As the ability to implement TGs confers Turing completeness \cite{hopcroft1969formal}, and ANNs confer universal function approximation \cite{hornik1989multilayer}, we can deduce that NMC is both universal in terms of algorithm compatibility and its ability to approximate functions. 

Given this universality, the relevant question of NMC is not ``\textit{can} neuromorphic solve this task?'', but rather ``\textit{should} NMC be used to solve this task?''. While NMC is by definition general purpose in potential, based on the `No Free Lunch' theorem as applied to computer architectures, it should be expected that it will be better at some tasks compared to others \cite{wolpert1997no}. The value proposition of NMC is increasingly important given the increased use of other specialized architectures, such as general-purpose graphics processing units (GPUs). 

This value proposition has been further complicated by the fact that the NMC field itself spans a number of different timescales and levels of technology readiness \cite{christensen20222022,indiveri2011neuromorphic, markovic2020physics, aimone2021roadmap, mead2002neuromorphic}. Unlike conventional architectures that are commercial today and emerging technologies such as quantum computing that will likely remain research platforms for the foreseeable future, neuromorphic research includes technologies that are near-ready for widespread adoption now as well as exploration of novel approaches that are many years out. As hybrid analog--digital NMC systems become increasingly available \cite{dalgaty2024mosaic, richter2024dynap, moradi2017scalable, pehle2022brainscales, wan2022compute, bhattacharyya2023towards}, it is crucial that there exist a mechanism by which different design considerations (e.g., which components should analog devices preferentially target?) can be assessed.

\begin{figure*}[h]
\begin{center}
\includegraphics[width=.9\textwidth]{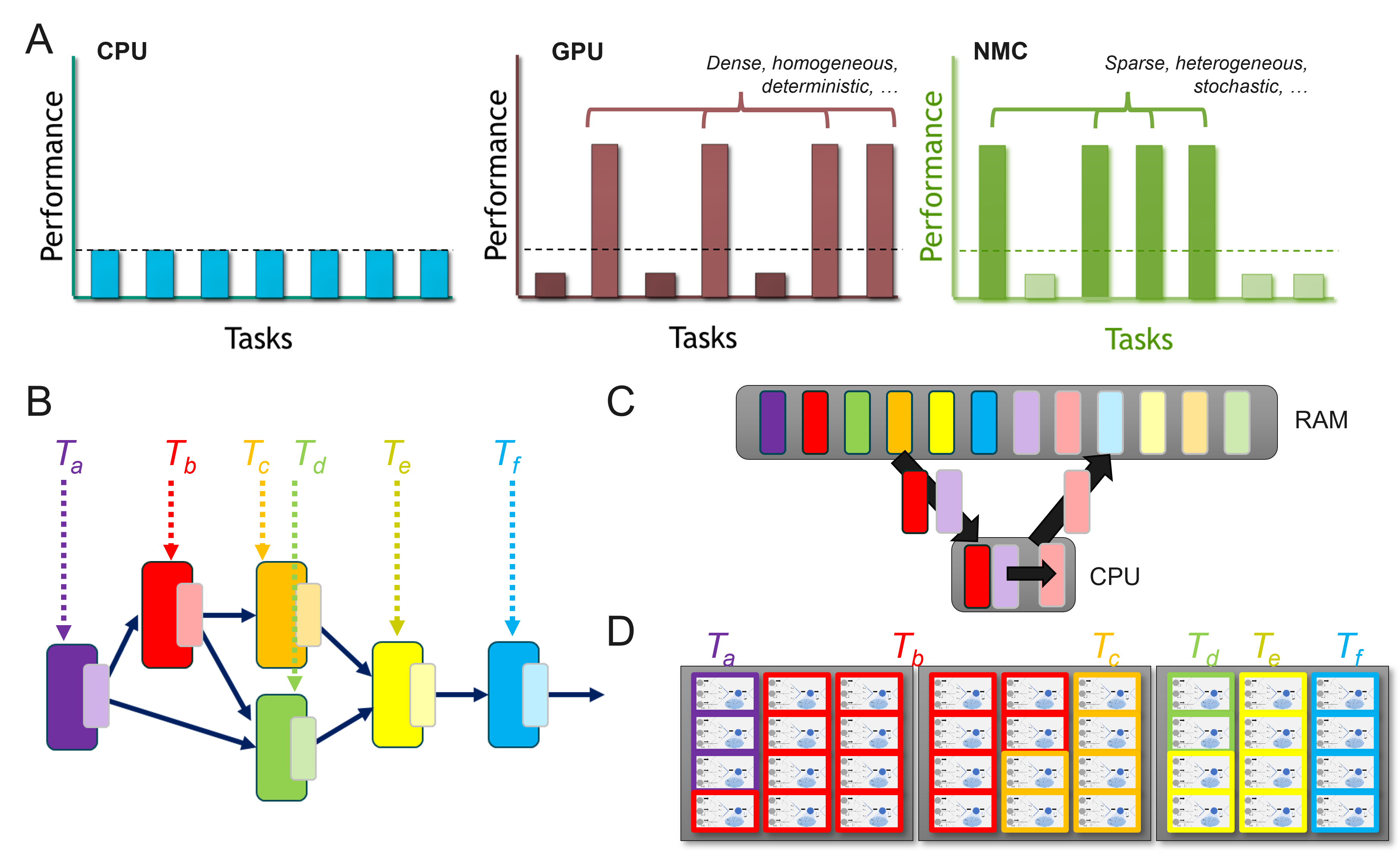}

\caption{(A) NMC hardware is GPU-like in generality, but shows advantageous capabilities in a different set of tasks. Listed are example algorithm features that can lend advantages (which this framework will explore); though note that algorithms can often be reformulated to fit an architecture. (B) An example algorithm graph that has six sets of instructions that require $T_i$ serial operations each. (C) A serial Von Neumann system stores both instructions and data in a single memory structure (RAM), iterating through the algorithm (CPU). (D) An ideal NMC processor spatially distributes the entire algorithm across hardware with a spatial cost proportional to the serial depth, $T_i$, for each operation.}
\label{fig:NMC_GPU}
\end{center}

\end{figure*}

This paper describes how NMC can be viewed as a class of specialized general-purpose architectures that is complementary to GPUs and other types of linear algebra accelerators that have become widespread in computing today. Figure \ref{fig:NMC_GPU} illustrates the notional hypothesis of this analysis: like GPUs, there exist several classes of computations that NMC excels at relative to conventional processors, and moreover that this is a distinct set of applications than what modern accelerators have targeted. Through the next few sections, we will discuss how NMC architecture differ from the conventional Von Neumann approach and we will show how these differences have direct impact on what types of computations NMC excels at relative to more conventional architectural approaches.

\subsection{Previous Work and Contributions}

There have been a number of proposed formal frameworks for neural computation, though they vary considerably in their level of abstraction and goals. Many of these efforts have focused on a formal analysis of how ensembles of neurons can perform computation. For example, early work by Wolfgang Maass and others explored the formal value of spiking neurons (e.g., \cite{maass1996lower, maass1997networks}), which eventually led to explorations of dynamical ensembles \cite{maass2002real, buonomano2009state} and a more formal framework around neural assembly calculus by Papadimitriou \cite{papadimitriou2020brain}. With a slightly different perspective, Chris Eliasmith and colleagues explored the programmability of neural ensembles through a control and dynamical systems perspective \cite{eliasmith2003neural} that has been extended to brain-like circuits \cite{eliasmith2012large} as well as state-space models \cite{voelker2019legendre}. These efforts relate to a broader historic literature looking at the theoretical potential of generic recurrent neural networks \cite{siegelmann1991turing, siegelmann1992computational, siegelmann1994analog, jaeger2001echo}.

These efforts (along with the broader ANN literature) demonstrate that neural algorithms can be constructed with many strategies in mind. While these efforts have been successful in motivating algorithm design (e.g., \cite{jaeger2001echo, maass2002real, voelker2019legendre}), they do not extend to architecture-agnostic cost models suitable for comparing NMC versus conventional approaches. There have been nascent efforts to address this need from a physical computing perspective where the basis of computation is dynamical in nature \cite{jaeger2023toward, pedersen2024neuromorphic, pedersen2025neuromorphic}, the generic implications for analyzing algorithms on brain-inspired hardware remain relatively unexplored \cite{aimone2023brain}. While we and others have explored graph-based neural algorithm design with mappings to spiking NMC hardware as a goal \cite{aimone2019dynamic, aimone2020provable, aimone2021provable, kwisthout2020computational, kay2020neuromorphic, kay2021neuromorphic, hamilton2018neural}, these too have been somewhat restricted by the need for an abstract programming model for analysis.  

Because most of these algorithm-centric frameworks by necessity abstract neural computation to a restricted set of algorithmic primitives, these efforts are not always aligned with the need of developing programmable NMC hardware as a general-purpose resource. This additionally complicates the incorporation of further insights from neurobiology beyond what is commonly used today (e.g., connectionist network design, spiking, Hebbian synaptic plasticity) for which the brain may represent a largely untapped resource \cite{aimone2019neural}. For this reason, the framework developed here defers the exploration of functional algorithm design; rather the goal of the framework is to remain agnostic to specific neural computing strategies in lieu of focusing on how such approaches can be formally analyzed with respect to neuromorphic hardware and neurobiological processes.  

Accordingly, the framework presented here makes the following contributions:

\begin{itemize}
    \item Defines a common neural algorithm and architecture representation $G_N=(N,S)$ and execution trace $G_N(t)$ that is amenable to formal analysis across neuroscience, neuromorphic hardware, and conventional computation.
    \item Shows that in the fully-parallel limit, NMC time and footprint tradeoffs align with classical parallel circuit complexity (e.g., Brent-style bounds); i.e., there is no generic asymptotic space / time advantage implied by "being neuromorphic."
    \item Expresses a trace-based energy scaling model for NMC and proves sufficient conditions for an asymptotic energy complexity separation between activity-sparse NMC execution and a dense scheduled conventional baseline.
    \item Derives digital and analog extensions to this NMC trace-based energy model, showing how for dynamical-system computations on digital NMC a microstate-derivative interpretation connects to the step-by-step variation in represented state, and hence the magnitude of modeled dynamics state changes.
    \item Introduces structural metrics (algorithm reuse $R_T$, degree/fan-out $K$, within-step homogeneity $H$) that predict when workloads are more NMC-aligned versus SIMD-aligned.
    \item Identifies activity trace metrics (activity intensity $\Phi_S$, decay, variability $CV_S$) that characterize when event-driven execution can be beneficial.
\end{itemize}

\section{Defining Neuromorphic Architectures in the Context of Von Neumann}

NMC is typically described as a non-Von Neumann architecture, but rarely is it specified what type of architecture that it is. Here, we will discuss what the implications of being non-Von Neumann are and how this affects how we should view today's NMC platforms. However, first it is useful to briefly characterize what a Von Neumann architecture is and why it is so powerful.

\subsection{Strengths and weaknesses of Von Neumann}

At a very simple level, the Von Neumann architecture refers to a \textit{stored-program} architecture wherein a memory external to the processor includes the instructions that represent a program as well as the data that will be processed. At the lowest level, the processor has specialized logic such as arithmetic logic units (ALU) that consist of a spectrum of specialized low-level circuitry to perform specific operations. From those basic operations, programs can be constructed to implement more sophisticated calculations. For example, a simple ALU has dedicated hardware to implement basic binary arithmetic (e.g., addition, subtraction), comparison operations (e.g., greater than, less than), and logical operations (e.g., bitwise logical AND, XOR). A Von Neumann program is simply a series of instructions that consist of which hardware-level operation to use, where in memory to pull inputs from, and  where to store the outputs. 

Von Neumann architectures are ubiquitous today for good reason. By separating processing and memory, each aspect can be designed and improved independently of the other. For instance, a powerful general-purpose processor, like a CPU, can dedicate considerable resources to having a wide range of increasingly sophisticated calculations hardware accelerated, while a more specialized processor can prioritize a subset of operations (as a GPU does for linear algebra), and a Reduced Instruction Set Computing (RISC) architecture may only have a lightweight set of instructions to maximize efficiency. Similarly, the separation of memory has allowed industry to focus on increasing density and access speeds, and with 64-bit addressing, there is effectively no limit to program or data size. This separation has also led to an important feature that has further entrenched the architecture---a serial program written fifty years ago in principle can run on today's hardware and vice versa. As a result, arguably this feature entrenched Von Neumann programs as the catalyst of the first Hardware Lottery \cite{hooker2021hardware}. 

The downside of Von Neumann is that scaling to larger systems with more memory and more powerful compute elements physically results in moving computation further away from memory. Stated differently, a bigger memory or a bigger processor requires that information---both data and the program itself---be moved longer distances. For this reason, the vast majority of the energy cost of computers today is in memory accesses \cite{horowitz20141}, and much of modern computer architecture research focuses on this challenge \cite{hennessy2019new}. This is one reason why Moore's Law was so critical, so long as transistor sizes were getting smaller, more memory could easily be placed nearer to the processing which effectively bounded the time and energy costs of having to go off chip. 

\subsection{Formal model definitions for neuromorphic computing}

\begin{figure}
    \centering
    \includegraphics[width=0.95\linewidth]{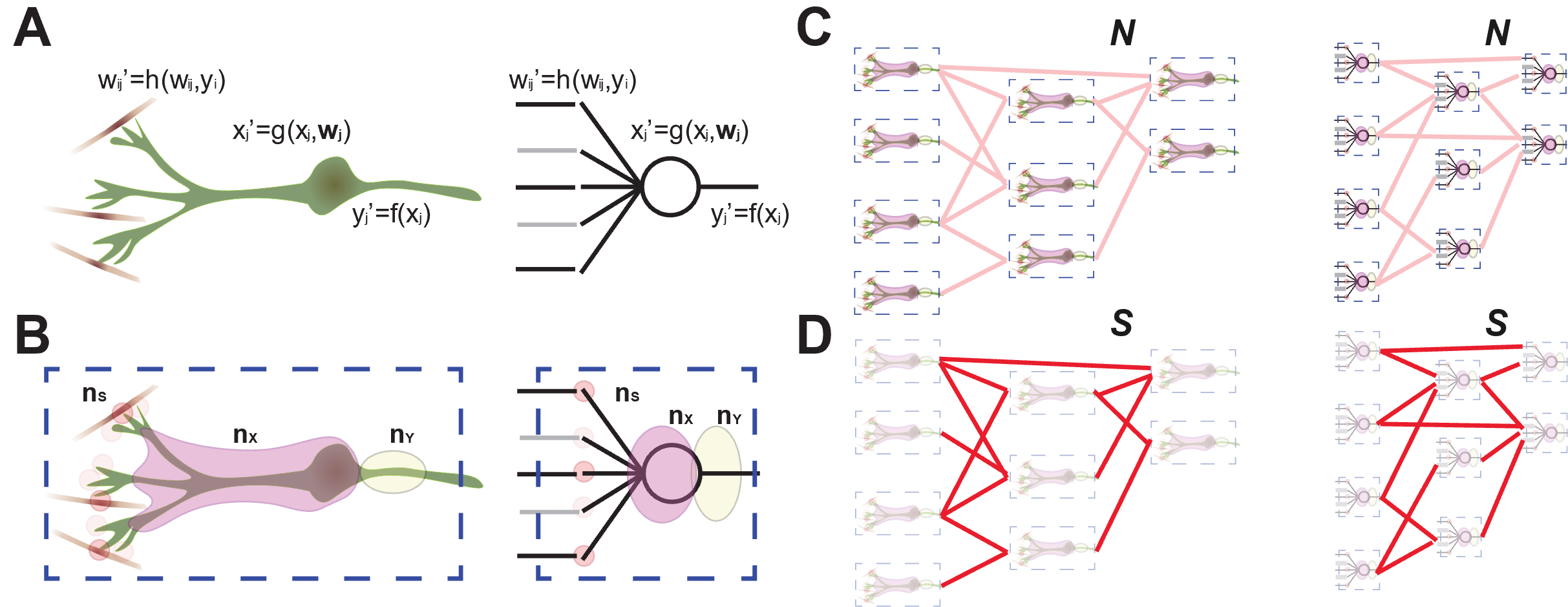}
    \caption{Illustration of model framework. (A) Biological and artificial neurons generally have the same components: synapses which exhibit synaptic dynamics and learning; dendrites and the soma which exhibit internal state dynamics, and spike generation at the soma / axon boundary. (B) The framework treats each of these dynamics as a graph of composable computational elements within each neuron. All neurons own their input synaptic dynamics (both pre- and post-synaptic components; $n_S$), all dendritic and somatic dynamics ($n_X$), and spike generation ($n_Y$). (C) The set of all neuron parts, referred to as the neurons, $N$, represents the totality of the dynamics of a neural algorithm. (D) The set of all connections between neurons, referred to here as the synapses, $S$, represents the interconnect between neurons. A neural algorithm is described as an algorithm graph of these neurons (vertices) and synapses (edges), $G_N=(N,S)$}.
    \label{fig:framework_overview}
\end{figure}

Here, we briefly summarize the framework that is expanded in full detail within the Methods \ref{sec:alg_prelim} and is summarized in Figure \ref{fig:framework_overview}.

Central to the framework is its treatment of neurons (computational elements), and synapses (communication elements). In the brain and many neural algorithms, both neurons and synapses are responsible for computation and exhibit dynamics at multiple scales, with the distinction between communication and computation remaining blurry. This framework accounts for this by consolidating all dynamics and computational transformations into the neurons proper. Each neuron itself is viewed as a graph of interconnected compute elements, permitting synapse dynamics (e.g., stochastic transmission, learning), neuronal dynamics (e.g., dendrites, leakiness), and spike generation to be arbitrarily simple or complex. We term these different classes of dynamics the neuron's governing functions, however the specifics of these functions are largely inconsequential for the complexity analysis of a neural algorithm. In contrast to neurons, synapses are highly simplified in this model, representing the directed graph of how neurons communicate through discrete (i.e., spiking) channels. Synapses accordingly have no dynamics or computation (all synaptic weights and dynamics are handled within the neuron) and exist only as communication channels with a binary state of $0$ or $1$.

This separation has several benefits. The most important is that this separation helps isolate the costs of non-local (i.e., between neuron) communication, which often dominates computing architectures, from that of the computation within neurons, which can vary tremendously between algorithms and architectures. Such separation allows measures of synapse scaling complexity to be assessed distinctly from that of neurons, which is critical because the constant factors associated with neuron complexity are quite variable. Finally, while we will not explore hardware implications much in this study, the restriction of dynamics into isolated neuron graphs allows the implications of alternative hardware strategies, such as analog devices, to be more deliberately quantified.

While this framework aims for broad generality, there are a few neurobiological cases which are not immediately compatible with the current formulation and as such would require extensions if desired. These include
\begin{itemize}
    \item Extra-synaptic connections between neurons, such as gap junctions (violates the isolation of neuron data structures). 
    \item Retrograde synaptic transmission from dendrite to axon (violates the directedness of synapse data structure).
    \item Structural plasticity, such as adult neurogenesis, for which pre-allocation of synapses is not possible (violates the absence of dynamics in synapse data structure)
\end{itemize}

The Turing completeness and universality of neural circuits is for the most part independent of the governing functions \footnote{For example, universal function approximation is proven so long as $f(.)$ is continuous, bounded, and non-linear \cite{hornik1991approximation}.}. Therefore, for any desired target function, for different governing functions there will be different neural algorithms. Understanding this neural algorithm equivalency is critical for mapping an NMC algorithm onto the governing functions available on a given architecture, and likewise is necessary for defining the computational trade-offs for evaluating the advantages or disadvantages of using an NMC-compatible algorithm. Stated differently, the equivalent leaky integrate-and-fire (i.e., `spiking') algorithm for an ANN model will be a different computational graph that may require more or fewer neurons and synapses.

\subsection{Ideal versus realizable neuromorphic architectures and algorithms}

The framework treats both neural architectures and neural algorithms as computational graphs, with neurons as vertices responsible for computation and synapses as edges responsible for communication. As will be shown, this formulation allows direct comparison of neuromorphic algorithms to conventional algorithms which are also often described as computational graphs.

Because neuromorphic hardware is an in-memory approach, a neural algorithm can be viewed as a parameterized subgraph of a neural architecture, and as such a generic neural algorithm can potentially be mapped to suitable architecture. This view is consistent with the often claimed (but rarely formalized) notion that the brain’s architectures and algorithms are one and the same. 

It should be noted that not all neuromorphic architectures are programmable in this generic manner, and all realized neuromorphic systems will have some constraints (e.g., synaptic fan-in / fan-out; neuron capacity) that will eventually lead to added embedding costs or ultimately disallow algorithm use. That said, the following analysis remains agnostic to the constraints of today’s systems and assumes that eventual neural architectures will be able to accommodate the algorithmic scaling costs determined here.

Definition \ref{def:architecture} defines an NMC architecture in a generic manner, however we can further distinguish an ideal NMC architecture from those that are practically realizable. This is useful because while the ideal NMC architecture is a useful strawman of sorts, we can explore the computational advantages of an ideal NMC approach and then back off of those with the costs (and benefits) of backing away from this ideal architecture to a practical approach. 

Our definition of an ideal NMC architecture relates to the parallel structure of the architecture itself. Like the brain, every neuron in an ideal NMC system would be physically distinct from every other neuron and every synaptic connection would similarly exist physically distinct from all others (i.e., point to point connectivity). In contrast, today's NMC architectures leverage a conventional-like hierarchy of processing elements and network routing. For example, Intel Loihi cores have local memory that stores the state variables of many neurons and their associated synapses and leverage specialized circuitry to rapidly update those neurons' states \cite{davies2018loihi}. In effect, below the core level the architecture itself is not neuromorphic as much as it is a highly-specialized RISC stored program architecture. From an algorithm perspective it does not strictly matter that at the lowest level an NMC architecture has cores that are shared by many neurons; but this distinction from the ideal architecture does introduce notable savings (e.g., space) and costs (e.g., time, embedding constraints) that we must eventually consider.

What we do not include in the definition of an ideal architecture are the governing functions themselves (e.g., the activation functions of neurons or the computation within synapses). In part we do this to separate the potential functionality of the components from the architecture itself. This is rather fundamental as the brain is not programmable in the same sense as engineered hardware---the circuit is what it is, the functions are what they are. However, there is a more fundamental reason for dissociating the two. Modifications to the governing functions require a change the algorithms themselves \footnote{For intuition, consider a compute graph $G_{sigmoid}$ of sigmoid activated neurons; an equivalent spiking neural network ($G_{spiking}$) and an equivalent rectified linear neuron network ($G_{ReLU}$) can be constructed, but these computational graphs will be different.}; whereas the ideal versus practical distinction here should not impact a neural algorithm inasmuch as it relates to the embedding and performance of that algorithm. Unless stated otherwise, for the remainder of this analysis we will explore NMC specifically in the context of an ideal NMC architecture.

\subsection{Overview of analysis}
The goal of this paper is to make the value proposition of neuromorphic computing precise enough to be useful. To do that, we take a deliberately simple approach: we focus on how the \emph{time}, \emph{footprint}, and \emph{energy} of a neuromorphic formulation scale as the underlying algorithm grows. The intent is not to predict exact performance of any specific chip, but to provide a common language for reasoning about which algorithm classes are (and are not) naturally aligned with neuromorphic execution.

Time and space are tightly coupled in any parallel architecture, and neuromorphic is no exception. The idealized model studied here shows that NMC does not intrinsically offer any time--space advantages over alternative parallel architecture approaches, a fact that is perhaps self-evident to theoretical computer scientists but is not widely appreciated (Section~\ref{sec:NMCadvantage}). Energy, however, is where neuromorphic departs most sharply from conventional stored-program execution. Because neuromorphic computation is in-memory, asynchronous, and event-driven, energy is not naturally tied to a fixed schedule of operations and memory accesses, and therefore it generally cannot be inferred from structure alone.

For this reason, the analysis makes a distinction between the \emph{structure} of an algorithm and its \emph{execution}. The structural object is the instantiated neuromorphic algorithm graph, $G_N=(N,S)$, which largely determines footprint and bounds time. The execution object is the activity trace,
\begin{equation}
    \Delta G_N(t) = (\Delta N(t), \Delta S(t)),    
\end{equation}

which records which neuron-owned state updates and which synaptic communication events actually occur at each time step. This trace is input- and regime-dependent: the same instantiated graph can exhibit very different activity depending on the data being processed and on where the computation is operating (e.g., transient versus converged dynamics). The advantage of introducing this distinction is that it makes the energy question well-posed: in an event-driven system, energy scales with realized activity, so identifying neuromorphic-aligned algorithms reduces to identifying when useful computations can be expressed with favorable structure \emph{and} favorable traces.

\section{The Benefits and Costs of Moving to Neuromorphic from Von Neumann} 

Even though NMC is not Von Neumann, if it is to be used in a general-purpose manner, it ultimately must satisfy many of the same requirements, such as programmability, scaling, and reliability. As such, it has a distinct set of tradeoffs it must deal with. One-by-one, we use the framework described above to qualitatively walk through some of the positive and negative implications of NMC being non-Von Neumann.

\subsection{Neuromorphic computing is programmable, but it is \textbf{not} a stored-program architecture}

Once programmed, NMC systems generally do not have an external memory from which it serially retrieves the next step of a program continuously during operation. Rather, the program, as such, is directly encoded in the construction of the neural circuit on which information will be computed, a fact which both confers benefits but introduces some direct and indirect challenges. 

The most immediate implication of this is that while NMC is programmable, it is initially best to  consider that it has a fixed program for the duration of its operation. 
The program being fixed means that the whole program must be spatially deployed across the neural hardware. This physical realization of a model does have very real implications on what NMC can be used for. Generally, on today's scalable NMC platforms \cite{kudithipudi2025neuromorphic} the program is defined as the computational graph of neurons and synapses that directly represents the algorithm being implemented \cite{aimone2019composing}, as in definition \ref{def:algorithm}. However, it is important to note that because the program is fully instantiated in hardware, the feasible program size is constrained by the available hardware."

Of course, this physical realization comes with a benefit as well---the program does not need to be retrieved from memory. This immediately cuts down on a substantial part of the cost of a serial program. 
As such, NMC can potentially benefit significantly by identifying calculations that can largely reuse a set of pre-defined calculations. In neural algorithms, this re-use arises in feedback connections between neurons, referred to as \textit{recurrence}. Such recurrence is a defining feature of biological neural circuits in the brain, and it has been observed that recurrent neural networks are preferential on NMC \cite{davies2021advancing}. Notably, this use of recurrence is a clear distinction from how algorithms are often 'unrolled' on conventional architectures which enforce acyclic graphs, effectively increasing footprint. 
\subsection{Neuromorphic computing is intrinsically parallel and asynchronous}
\label{section:nmc_mimd}
Inherent in the graph-based description of NMC programs described above is the implication that NMC is intrinsically a parallel architecture. While details differ across NMC platforms, algorithmically NMC is typically be viewed as having every neuron acting independently and asynchronously from one another. It should be emphasized how different this extremely parallel neural programming paradigm is from conventional parallel architectures. 

Parallel computing introduces a whole set of unique challenges at both the algorithm and architectural level, and one significant simplification that has led to significant efficiencies is to focus on ``single-instruction, multiple data'', or SIMD, approaches that shape algorithms to target a number of very powerful compute cores with a set of common instructions that are applied to a large volume of data. For this reason, many parallel architectures, such as GPUs, are SIMD. SIMD-based architectures are powerful with appropriate algorithms, such as linear algebra applications, where the required mathematical operations and memory accesses are highly structured. The downside of SIMD is that if an application lacks structure, such as Monte Carlo simulations with highly divergent trajectories, the uniformity of SIMD becomes a drawback. For heterogeneous applications, NMC effectively provides a unique path to an alternative multiple-instruction, multiple data-, or MIMD-,like architecture---since every calculation is implemented in its own population of neurons, there is no reason that these calculations need to be the same. 

The flip side of spreading out a computation in this manner is that communication costs can become problematic. The brain's solution to this is to use \textit{spiking }to minimize the costs of communication. Spiking refers to the event-driven communication of neurons, whereby a neuron only communicates if its inputs satisfy some condition, such as crossing a threshold. In the brain, this is an all-or-none single-bit---transmit a $\texttt{1}$ or no transmission at all. In today's NMC platforms the spike may consist of more information, but they are always event-driven (see section \ref{sec:NMC_non-deterministic}) and almost always target-agnostic. This creates a data-dependent cost that is not present in conventional computation. If $\texttt{0}$s are effectively free in communication and computation, sparsity becomes a particularly important feature of algorithm design. 

\subsection{Neuromorphic processing is data-dependent and often non-deterministic}
\label{sec:NMC_non-deterministic}

Because communication is event-driven and neurons behave asynchronously, computation effectively becomes event-driven as well. An individual neuron processes information as it arrives. The combined effect of intrinsic stochasticity, distributed computation, and event-driven communication means that neuromorphic computation can easily become non-deterministic and non-repetitive. Stated differently, the same computation may be performed by many different patterns of neural behavior. 

While the non-determinism and adaptability of biological neural circuits makes neural computation richer and potentially more computationally powerful from a theoretical sense, this complexity also risks making NMC more challenging for those familiar with precise and sequential deterministic algorithms. For this reason, we will not examine these architectural benefits in detail here but rather we will focus more generally on the activity-dependent trace of algorithm graphs as a complement to static structural algorithm graph. 

\section{What neuromorphic is good at}
\label{sec:WhatNMCGoodAt}

Here, we will consider three primary theoretical metrics: how an NMC architecture impacts the time ($C_T$), space ($C_S$), and energy ($C_E$) complexity costs of an algorithm. Most models of parallel computation consider the implications of parallel processors sharing a memory (e.g., the PRAM model \cite{fortune1978parallelism, gibbons1989more}), but there is no external shared-memory instruction and data access in a neuromorphic architecture. However, from a processor perspective many of the same frameworks apply. For instance, it is useful to relate our neuromorphic algorithms, $G_N=(N,S)$ to a computationally equivalent fully expanded out conventional algorithm described as a directed acyclic graph (DAG), which we call $G=(\mathcal{V}, \mathcal{E})$ per definition \ref{def:algorithm}. Summarized results of complexity scaling, which will be derived in the following sections, are presented in Table \ref{tab:complexity_scaling}.

We start first by summarizing how the this framework can be used to understand what we can say about how NMC architectures can impact the complexity scaling of neuromorphic algorithms in general, with the full formal analysis provided in Methods \ref{sec:complexityAnalysis}. Later, we will expand how these formalisms can be used to look at specific classes of neuromorphic algorithms.

\begin{table}[h]
    \centering
    \caption{Ideal Time, Space, and Energy Scaling of Neuromorphic and Conventional Systems} 
    \begin{tabular}{ccccc}
        \toprule
        & Time ($T$) & Space / Footprint ($C_S$) & Energy ($E$) & Source \\
        \midrule

        \textbf{Conv.} 
          & $O\!\left(\frac{T_1}{p}+T_\infty\right)$ 
          & $O\!\left(p + C_{\mathrm{mem}}\right)$ 
          & $O\!\left(A_V + A_{acc}\right)$
          & Def.~\ref{def:timeComplexity}, Def.~\ref{def:spaceComplexity}, Def.~\ref{def:energy_vn} \\

        \textbf{NMC} 
          & $O\!\left(T_\infty(G_N^{(T)})\right)$ 
          & $O\!\left(|G_N|\right)$ 
          & $O\!\left(|\Delta G_N[0{:}T)|\right)$
          & Lem.~\ref{lem:ideal_nmc_depth_time}, Def.~\ref{def:spaceComplexity}, Def.~\ref{def:energyComplexity} \\
        
        \bottomrule
    \end{tabular}
    \vspace{0.1cm} 
    \begin{flushleft}
    \small
    For a conventional algorithm graph $G=(\mathcal{V}, \mathcal{E})$, $T_1$ denotes total work and $T_\infty$ denotes critical-path depth. Conventional footprint includes processor resources plus required memory, where $p$ is the number of conventional processors/threads, and $C_{\mathrm{mem}}$ is the memory required to represent the algorithm graph and state. For conventional energy, $A_V:=\sum_t|A_V(t)|$ and $A_{acc} := \sum_t|A_{acc}(t)|$, where $A_V(t)$ and $A_{acc}(t)$ are the cardinality of the sets of operations and memory accesses, respectively.

    For a neuromorphic algorithm graph $G_N=(N,S)$, $|G_N|$ denotes the instantiated graph footprint (e.g., nodes plus edges under a consistent size/footprint measure). Cumulative activity $\Delta G_N[0{:}T)$ is defined in Def.~\ref{def:deltaG} and used in Def.~\ref{def:energyComplexity}.
    \end{flushleft}
    \label{tab:complexity_scaling}
\end{table}

\subsection{Time and space: neuromorphic operates at depth-limits of extreme parallelism but with associated footprint cost}
\label{sec:time_space_summary}

A useful way to think about a neuromorphic algorithm is as a computational graph that has been turned into hardware. Once the ``program'' is instantiated as a graph of neurons and synapses, $G_N=(N,S)$, the computation is no longer driven by a serial instruction stream. Instead, different parts of the graph can proceed whenever their local dependencies are satisfied. In the ideal limit---where there is enough physical parallelism to avoid resource contention—---runtime is governed by the depth of the induced dependency structure, i.e., the critical path $T_\infty$ (Definition~\ref{def:timeComplexity}). Note that this computational-depth limit at which NMC (as a spatially-instantied, in-memory architecture) naturally operates is equivalent to the limits of classical parallel models and circuit complexity.

While operating near the ideal computational depth may be appealing, there is a significant associated footprint cost. In a stored-program machine, a fixed pool of processors can time-multiplex an arbitrarily large computation from a compact description held in memory (in a sense trading space for time). Whereas in neuromorphic, the program \textit{is} the circuit: because the instantiated graph must exist in hardware, the footprint will scale with the size of that graph (Definition~\ref{def:spaceComplexity}). Said differently, neuromorphic shifts effort away from repeated instruction fetch and memory access and toward physically realizing the dependency structure. This makes shifting the depth--footprint tradeoff more an algorithm, rather than architecture, challenge: reducing depth typically means spatializing more of the computation, whereas conserving footprint typically means accepting more sequential depth.

For NMC, recurrence is the most direct mechanism for making this footprint cost tolerable. Many computations do not need a fully unrolled feed-forward circuit; they can be expressed as a fixed recurrent circuit executed repeatedly. In the formal model this shows up as structural reuse which is quantified by $R_T$: when $R_T$ is large, the work of the time-unrolled computation grows with the execution horizon while the instantiated graph stays fixed (Methods~\ref{def:reuse}). This is why iterative and streaming computations tend to map naturally to neuromorphic substrates: the same deployed circuit can be reused many times without expanding the physical program.

Finally, it is worth being explicit about what this does \emph{not} imply. Because neuromorphic algorithms can be classically simulated, any asymptotic time improvement obtained by a neuromorphic formulation implies a corresponding conventional simulation up to overheads associated with representing and scheduling sparse event-driven updates (Proposition~\ref{prop:conventionalSimOfNeuromorphic} and Corollary~\ref{cor:timeAdvTransfers}). The most defensible general conclusion is that NMC \textbf{does not} change the time or space complexity class of an algorithm, but it naturally pushes computation toward the depth limit and shifts the practical bottlenecks toward footprint, locality, and execution-dependent activity.

\subsection{Energy: neuromorphic costs scale with activity, not work}
\label{sec:energy_summary}

While the time and space costs of NMC are aligned with the limits of conventional parallel computation, the energy consumption of NMC presents a qualitatively different picture. In a stored-program architecture, energy is naturally tied to a schedule: executed operations and associated memory accesses all consume power (Definition~\ref{def:energy_vn}). As a result, at a coarse scaling level conventional energy tends to track the total work $T_1$, up to constant factors set by the memory hierarchy and data movement.

NMC execution is different because it is event-driven. Computation and communication occur only when there are events to process: neurons update when relevant inputs arrive (or when their local dynamics require it), and synapses communicate only when spikes are emitted. For this reason, energy cannot be inferred from the static graph alone. The relevant object is the execution trace (Definition~\ref{def:deltaG}): the sets of neuron-owned updates $\Delta N(t)$ and synaptic communication events $\Delta S(t)$ that actually occur at each time step. Under an event-count model, neuromorphic energy is proportional to the cumulative activity
\begin{equation}
    |\Delta G[0{:}T)| \;=\; \sum_{t=0}^{T-1} |\Delta N(t)| \;+\; \sum_{t=0}^{T-1} |\Delta S(t)|,    
    \label{eqn:NMC_energy_main}
\end{equation}

up to platform-dependent constants (Definition~\ref{def:energyComplexity}, Lemmas~\ref{lem:degree_weighted_spike_count}--\ref{lemma:upper_DeltaN}). This is the practical payoff of separating structure from trace: two algorithms can have comparable footprint $|G_N|$ yet exhibit very different activity traces, and as a result have very different energy costs.

This trace-based view has two immediate consequences. First, it allows us to make the worst case for NMC energy costs explicit. If a constant fraction of synapses are active at every step, then cumulative activity is $\Theta(|S|T)$ and energy scales with the full interaction structure over time (Lemma~\ref{lem:nmc_energy_worstcase}). In that regime, neuromorphic does not avoid work---it simply pays for it in an event-driven manner. This is why dense, fully-active computations (e.g., feed-forward layers in ANNs) are not naturally aligned with neuromorphic energy efficiency unless the structure or the execution can be made substantially sparse.

Second, and perhaps more importantly, it identifies when a genuine scaling benefit is possible. When activity is sparse relative to the structural capacity of the graph—formally, when
\begin{equation}
    \sum_{t=0}^{T-1} |\Delta S(t)| \;=\; o(|S|T),
    \label{eqn:sparse_activity}
\end{equation}
and neuron-update activity remains proportional to synaptic activity—then energy scales with realized activity rather than the worst-case structural bound (Theorem~\ref{thm:energy_separation_template}). This is the regime suggested by convergent iterative computations: as a solver approaches a fixed point, fewer sites need to update, so both $|\Delta N(t)|$ and $|\Delta S(t)|$ can decrease substantially (in expectation). In these cases while the structural graph remains static, the trace becomes sparse and total energy falls accordingly. Put differently, NMC energy is proportional to cumulative activity, not merely how large the graph is.

It should be noted that trace-based savings are not exclusive to neuromorphic in principle: conventional systems can, and often do, adopt sparse, event-driven schedules. This framework simply clarifies that such trace-based energy advantages are natural within a neuromorphic system, whereas leveraging them in a conventional stored-program setting can introduce additional control and scheduling overhead.

\subsection{What does $\Delta G$ mean?}
A limitation of energy being a trace-based analysis is that it generally cannot be assessed a priori. Rather, strictly speaking, the only way to know the execution trace is to run the algorithm for all possible inputs, which is generally infeasible and undesirable. While the analysis above focuses on a generic formulation of NMC, we can offset some of this uncertainty by considering what the trace $\Delta G$ is capturing in different classes of neuromorphic hardware. This is important not only to help frame our interpretation of $\Delta G$, but also because the importance of analog computation is a major open question in the NMC research community.

Methods Section \ref{sec:DigitalAnalog_Extension} illustrates how introducing some specificity to the NMC design can aid in our interpretation of $\Delta G$, though as can be seen the analysis quickly can become architecture specific. The most straightforward analysis is for fully digital NMC approaches, such as Intel Loihi and IBM TrueNorth. In an idealized digital circuit, energy costs scale with changes in the microstate, a bit that does not change ($0 \rightarrow 0$, $1 \rightarrow 1$) in principle should not expend energy. Effectively, digital hardware pays for any $\text{XOR}$ across time at the bit level on the hardware, though different bits may cost different amounts to change. As such, $\Delta N(t)$ and $\Delta S(t)$ can be interpreted as coarse indicators of those microstate bits that are physically changed, as opposed to simply processed in the generic case. This is most evident with spiking: generally, all spikes will have a $0 \rightarrow 1$ followed by a $1 \rightarrow 0$ for every bit associated with routing. However, this analysis also presents implications on digital hardware designs that leverage non-event-driven intrinsic dynamics like leaks and decay.

While this digital case is mostly a reframing of the execution trace, for discrete time-evolving algorithms like dynamical systems or iterative solvers there is a connection between the algorithm derivative and the digital microstate derivative considered here. Intuitively, every update to state variables must correspond to some associated microstate change, so the energy and algorithm trace will be tied to the step-by-step variation o in the represented dynamical state. This exact relationship is dependent on both the nature of the function's time evolution and the efficiency of the algorithm; but in general this does provide an a priori perspective for how an efficient NMC algorithm should be constructed to minimize energy.

Importantly, this story will change for analog architectures. While digital devices are relatively well-defined, there are many types of analog devices, some of which simply represent a higher precision non-volatile state, while others have intrinsic dynamics. As such, a generalization similar to digital is not readily apparent; though we can assume that the energy cost of analog devices is often related to how far each device's state is moved relative to some intrinsic baseline. In particular, we can consider that the trace represented in those $\Delta N(t)$ elements that are analog in implementation will be related to the residual difference from the device's intrinsic steady state. This change can lead to a potential savings if algorithms are appropriately constructed to take advantage of this ``free'' computation (i.e., no additional drive), but at the same time introduces costs in maintaining an particular state steady over an extended time away from the device's equilibrium.

\section{Identifying a Neuromorphic Advantage}

The previous sections outline how the complexity scaling of time, space, and energy  differs on NMC from conventional architectures. The costs of NMC are related to the number of neurons and synapses in the neural algorithm, but in a different way than conventional architectures, whose costs are typically expressed in terms of scheduled work and memory access. This makes it challenging to develop clear apples--to--apples comparisons. 
Combined, the differences in how costs scale between NMC and conventional, summarized in Sections~\ref{sec:time_space_summary}--\ref{sec:energy_summary} (and formalized in the Methods), are broadly what we should expect. Ideal NMC not formally distinct from other idealized parallel architectures from a time--space perspective, though it naturally operates near the depth limit of parallel computation while being spatially expensive because the program must be instantiated. Energy benefits, when they occur, arise from trace behavior, particularly when the realized activity of the algorithm is small relative to the available structural capacity of the instantiated graph. This is consistent with the human brain itself. The brain performs complex calculations at extremely low computational depths (a batter can decide to hit a 100 mph pitch at the equivalent of 10 clock-cycles) with at exceptionally low power cost (estimated at lower than 20 W). However, this time and energy efficiency has a large spatial cost, requiring over $10^{11}$ neurons and $10^{15}$ synapses at levels of complexity far beyond any artificial neurons and synapses.

\begin{figure}
    \centering
    \includegraphics[width=.95\linewidth]{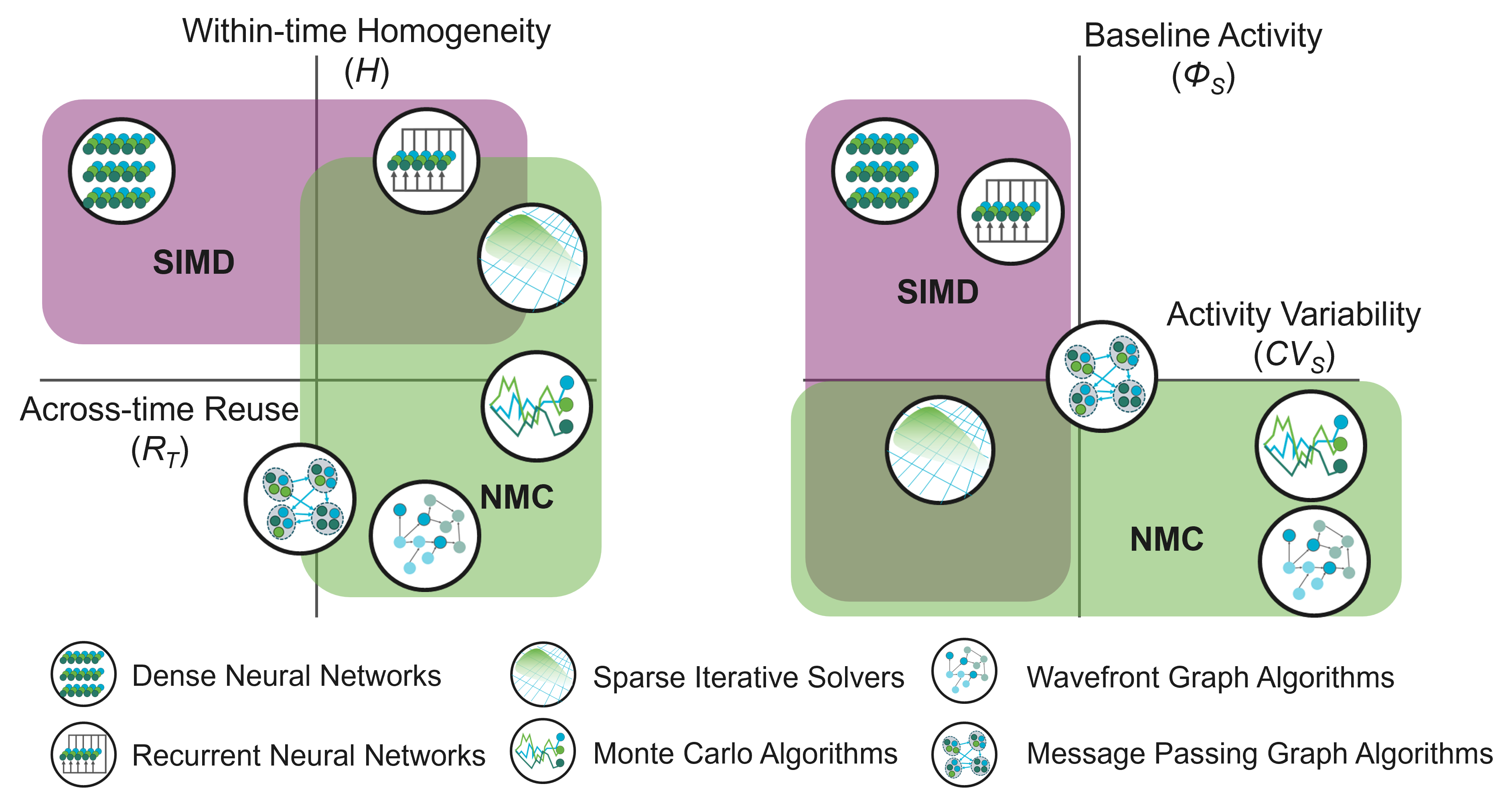}
    \caption{Notional landscape of algorithm classes against framework's structural and trace algorithm metrics. Left: Structural metrics, including within-time homogeneity ($H_T$) and across-time structural reuse ($R_T$), can be used to differentiate SIMD- and NMC-friendly algorithms. Right: Trace metrics, including baseline activity ($\Phi_S$) and activity variability ($CV_S$), provide a separate axis for assessing NMC friendliness.  Note that the placement of these different algorithm classes is notional; as the placement can vary considerably with input data and the specific application.}
    \label{fig:NMCvsSIMD_Algorithms}
\end{figure}

In Methods \ref{sec:NMCadvantage}, we identified several metrics that can characterize whether an algorithm class should be more or less suitable for NMC based on its algorithm structural graph ($G$) and activity graph $\Delta G$. 

Two of the structural metrics focus on the relative homogeneity of the computational graph (Figure \ref{fig:NMCvsSIMD_Algorithms}, left). \textbf{Within-timestep homogeneity}, $H(t)$, is a measure of how much repeated structure exists in the algorithm graph at each timestep $t$ \ref{def:simd_homogeneity}. A large $H(t)$ is highly preferable to systems with SIMD parallelism, as those systems can allow their many processors that share instructions to be used simultaneously. In contrast, an NMC system is rather agnostic to $H(t)$, as its full spatial realization makes it less able to benefit from repeated structure within a timestep. An example of high $H(t)$ would be linear algebra tasks in which all of the elements vector or matrix are updated the same sequence of operations. 

Across-timestep homogeneity represents a complementary, but different, form of homogeneity. An algorithm's \textbf{structural reuse factor}, $R_T(G)$, is defined as a measure of how much of an unrolled computational graph could be `rolled up' to reuse the same components \ref{def:reuse}. A large $R_T(G)$ is highly preferable to NMC systems, as this allows the space footprint to be greatly reduced from that of the otherwise unrolled algorithm. As a result, a high $R_T(G)$ can have a disproportionate benefit to an in-memory architecture, as it directly reduces processor costs, whereas in a stored-program architecture the benefits of recurrence are fairly low (a modest reduction of program-space in memory). Examples of high $R_T(G)$ are solvers that iterates over the same mesh or recurrent neural networks.

An additional structural metric that we consider is the locality of communication. Because NMC algorithms are realized in space, space and energy costs will increase with increased communication distance between compute elements. While the framework defines all within neuron computation and communication as local ($N$), communication between neurons ($S$) may not be. We define $K(G_N)$ as the \textbf{maximum fan-out} in the NMC algorithm graph $G_N=(N,S)$ (Definition \ref{def:boundedFanout}), and note that if $K(G_N)$ is bounded (i.e., independent of algorithm graph size), the number of synapses $|S|$ will scale linearly with the number of neurons. 

We also introduce three measures of trace-based algorithm analysis (Figure \ref{fig:NMCvsSIMD_Algorithms}, right). In some respects, these measures align with the structural measures above. First, we define \textbf{activity intensity}, $\Phi_S$, which describes the normalized synaptic activity of a NMC algorithm's execution (Definition \ref{def:activity_intensity}). There are two additional measures derived from this baseline activity, analogous to the across- and within-timestep structural metrics above. We define \textbf{activity convergence} (Definition \ref{def:activity_trajectory}) as a measure of how activity evolves over time. In particular, a convergent algorithm, whereby the expectation of per-step activity decreases ($\mathbb{E}[a_S(t+1)]<\mathbb{E}[a_S(t)]$), is highly advantageous to NMC as the energy cost will diminish over time. Additionally, we define \textbf{activity variability} ($CV(G)$, Definition \ref{def:activity_variability}) as a measure of how predictable activity is overtime. Like with structural homogeneity, conventional SIMD-architectures benefit from algorithms whose trace is highly homogeneous (low $CV(G)$), whereas NMC approaches are tolerable to high $CV_S(G_N; T)$.

Given these relative costs, we can return to our original motivating question of what classes of algorithms should be well suited for NMC. At first glance, it is evident that algorithms with the following properties may stand to benefit:

\begin{itemize}
    \item \textbf{Graph algorithms} Graph algorithms that can be described as operations over the graph can naturally lend themselves to the $G_N=(N,S)$ construction. Neural graph algorithms that lead to concentration of activity along a wavefront, as in graph search or dynamic programs, can yield a low $\Phi_S$ and high $CV_S(G_N; T)$ that can make NMC architectures preferable, particularly compared to SIMD-like alternatives. 
    \item \textbf{Iterative solvers and discrete optimization} Iterative algorithms by construction often allow high structural reuse (high $R_T$). Further, iterative algorithms often are designed with convergence as a goal, such as in discrete optimization applications, which can exhibit activity that is sparse or decreases over time. Decreasing $|\Delta N(t)|$ and $|\Delta S(t)|$ directly reduces cumulative activity and therefore energy (Definitions~\ref{def:deltaG} and \ref{def:activityDecay}; Theorem~\ref{thm:energy_separation_template}).
    \item \textbf{Probabilistic and sampling algorithms} For algorithms that repeatedly sample a fixed interaction structure, recurrence and reuse can amortize instantiation costs, and energy can track the number of realized events rather than the worst-case size of the state space (e.g., via $|\Delta G[0{:}T)|$). These algorithms also often have a high activity variability, which can it challenging for these algorithms to maximize the parallelization offered by SIMD alternatives.
    \item \textbf{Recurrent neural networks} Recurrent neural algorithms, such as reservoir networks and state-space models, often have a dependency structure can be expressed with cycles allow the same neuron resources to be reused over time, yielding large $R_T$ and mitigating footprint growth relative to a fully unrolled feed-forward implementation (Definition~\ref{def:reuse}). While these approaches can be tailored for sparse operation (low $\Phi_S$) and local communication (low $K_G$), the NMC benefits may be lost if they are operated in a dense activity or high fan-out condition.

\end{itemize}

While these broad algorithm classes can be thought of as potentially ideally suited for NMC architectures, it is worth diving deeper into two example algorithm classes to illustrate the relative benefits and downsides of an NMC approach.

\subsection{A positive example: iterative mesh-based algorithms}

Inspired by previously identified algorithms with claimed neuromorphic advantages \cite{theilman2025solving, smith2022neuromorphic}, we first consider a conceptual model of an algorithm that tracks the behavior of the system over $M_S$ state locations. For example, in a finite element simulation, the state of all $M_S$ locations are updated in an iteration, and in a discrete-time Markov Chain (DTMC) simulation $M_S$ reflects the dimensionality of the transition matrix. Consider specifically the case where each mesh point only communicates at most with a finite number (scale-free) of other locations, which is defined as the mesh maximum fan-out, $K$. The model runs for $M_T$ timesteps.

While it would never fully be fully expanded and unrolled, the computational graph of this problem, $G^{(T)}_{mesh}$, would in fact be a hierarchical graph of graphs consisting of what computations must be performed at each mesh point. We consider that within each model timestep each mesh node consists of a computational graph of $T_{1S}$ computations, consisting of a minimal depth of $T_{\infty S}$ to update the state of that mesh point. As there are $M_S$ meshpoints and since the model consists of $M_T$ timesteps, the fully expanded graph will be replicated $M_S \times M_T$ times.

Prior to any recognition of recurrence or parallelizability, we can quantify the size of $G^{(T)}_{mesh}$ as $T_{1, mesh} = M_S M_T T_{1S}$ and $T_{\inf, mesh} = M_T T_{\infty S}$.

\subsubsection{Conventional Implementation}

In a Von Neumann implementation, the description of the model and its states reside in memory, and a fixed pool of processors executes a schedule over that stored representation. For non-trivial problem sizes, much of the state and connectivity will typically live in a memory hierarchy that ultimately includes off-chip DRAM, and the computation proceeds by repeatedly fetching data, applying local update rules, and writing updated state back to memory.

We can use $G_{mesh}$ to estimate the computational costs of simulating the mesh on $p$ processors under a standard parallel-graph view of execution.

\textit{Time: }
Per Definition \ref{def:timeComplexity}, the time on a parallel computer with $p$ processors is bounded by the depth of the computation and the available parallelism. For this mesh family, the depth is set by the per-site update depth composed across $M_T$ iterations, i.e., $T_{\inf,mesh} = M_T T_{\inf S}$, while the total work scales as $T_{1,\text{mesh}}=M_S M_T T_{1S}$.

\textit{Space / Footprint: }
Under a stored-program model, the processor resources scale with $p$, and the required memory footprint scales with the number of state locations (and any stored representation of the local connectivity). For the present stylized mesh example, we summarize this footprint as scaling linearly with $M_S$:
\begin{equation}
    C_{p,\text{mesh}} = c_p p + c_{\text{mem}} M_S .
    \label{eqn:size_mesh_conv}
\end{equation}

\textit{Energy: }
To remain architecture-agnostic, we adopt the scheduled work/access baseline (Definition~\ref{def:energy_vn}), under which conventional energy scales with the number of executed operations and memory accesses. For this mesh example, this implies an energy scaling that tracks the total work (up to platform-dependent constants and memory-access factors), yielding:
\begin{equation}
    E_{p,\text{mesh}} \approx e_{\text{op}}\, T_{1,\text{mesh}},
    \quad
    T_{1,\text{mesh}} = M_S M_T T_{1S}.
    \label{eqn:energy_mesh_conv}
\end{equation}

\subsubsection{Neuromorphic Implementation}

\textit{Time: }
In the idealized execution model, we can interpret the runtime of an ideal NMC implementation of the mesh as being governed by the depth of the induced computation (Definition~\ref{def:timeComplexity}). For this mesh family, the critical path scales with the per-site depth across $M_T$ iterations, i.e., $T_{\inf, mesh} = M_T T_{\inf S}$ (up to constant factors associated with the chosen update semantics).

\textit{Space: }
We define $N_{mesh}$ as the number of neurons per meshpoint and we recognize that $N_{mesh} \approx O(T_{1 S})$. We know from the problem definition that the timesteps of $G_{mesh}$ can be represented without unrolling time into additional footprint (i.e., as a recurrent layer). As such, the instantiated size of the NMC deployment scales as
\begin{equation}
    C_{N, mesh} = c_N N_{mesh} M_S \approx O(T_{1 S} M_S).
\end{equation}

\textit{Energy: }
The key distinction in the framework is that neuromorphic energy depends on the execution trace, not only on the structural graph. Let $\Delta S(t)$ denote the set of synaptic communication events at time $t$ and $\Delta N(t)$ denote the set of neuron-owned updates at time $t$ (Definition~\ref{def:deltaG}). Under the event-count model, total energy over $M_T$ steps is proportional (up to constants) to the cumulative activity (Definition~\ref{def:energyComplexity})
\begin{equation}
    |\Delta G[0{:}M_T)| \;=\; \sum_{t=0}^{M_T-1} |\Delta N(t)| \;+\; \sum_{t=0}^{M_T-1} |\Delta S(t)|.
    \label{eqn:mesh_activity_def}
\end{equation}

For the mesh family, bounded fan-out $K$ implies that the structural number of synapses scales linearly with the number of sites, $|S| = O(K M_S)=O(M_S)$. In the worst case, a constant fraction of synapses are active at each step, i.e., $|\Delta S(t)|=\Theta(|S|)=\Theta(M_S)$ for all $t$, which yields cumulative synaptic activity $\sum_t |\Delta S(t)|=\Theta(|S|M_T)=\Theta(M_S M_T)$ (Lemma~\ref{lem:nmc_energy_worstcase}). In this high-activity regime, the event-driven energy model does not change the scaling: energy is proportional to the full interaction structure over time.

The interesting case is when the mesh dynamics are \textit{iterative} and convergent, and the neuromorphic formulation is designed so that activity decreases as the system approaches steady-state. In the framework language, this corresponds to executions for which $|\Delta S(t)|$ (and comparably $|\Delta N(t)|$) is sparse relative to the available structural capacity $|S|$, for example satisfying
\begin{equation}
\sum_{t=0}^{M_T-1} |\Delta S(t)| = o(|S|M_T).
\end{equation}
In that regime, the framework predicts that total energy scales with realized activity rather than with the worst-case bound (Theorem~\ref{thm:energy_separation_template}). Intuitively, the instantiated mesh graph remains available, but fewer sites (and therefore fewer communication edges) need to update as convergence progresses, so the trace becomes progressively sparser.

\subsection{A less-promising example: dense linear algebra-based ANNs}

As a contrast, it is useful to consider a numerical computing example that is not inherently advantageous for neuromorphic. ANNs are typically defined in terms of linear algebra functions, such as vector, matrix, and tensor multiplications. This linear algebra formulation of ANNs motivated the initial use of GPUs for ANNs \cite{raina2009large, krizhevsky2012imagenet} and ultimately the explosive development of large-scale linear algebra-based ANNs \cite{hooker2021hardware}.

While this linear algebra approach is seen as a natural fit to crossbar-based neural accelerators \cite{yang2013memristive, cai2019fully}, promising results for scalable spiking neuromorphic platforms on ANNs have been limited to narrow cases \cite{davies2021advancing}. The benefits of an event-driven neuromorphic system described above are less clear-cut.

To illustrate this, we consider the computational graph of an ANN, and since ANNs are typically a sequence of non-identical layers, for simplicity we focus on the graph of a single feed-forward layer, termed $G_{ff}$. A single layer of a neural network performs the following steps:
\begin{equation}
\begin{aligned}
    \text{\textbf{x}}_{j} &= W \text{\textbf{y}}_i \\
    \text{\textbf{y}}_{j} &= f(\text{\textbf{x}}_{j})
\end{aligned}
\label{eqn:feedforwardANN}
\end{equation}
whereby $f(\cdot)$ is a simple non-linearity, as in Table \ref{tab:neuron_types}.

This graph classically is described in two stages: a vector--matrix multiply followed by a non-linearity operated over a vector. However, for a computational implementation it is useful to view it as three distinct stages: a layer consisting of every synaptic operation ($s_{ij}=w_{ij} y_i$), of which there are $S_{total} = N_i N_j$ operations, every neuron summation ($x_j = \Sigma_i s_{ij}$), of which there are $N_j$ operations, and every neuron non-linearity ($y_j = f(x_j)$), of which there are also $N_j$ operations. The minimal depth of this circuit is $T_{\infty}=3$ and the total operation count, $T_1$, scales with the number of synapses, i.e., $T_1 = \Theta(N_i N_j)$.

\subsubsection{Conventional implementation}

Conventional implementations of equation \ref{eqn:feedforwardANN} are straightforward, but have been highly optimized over recent years due to the rise of ANN applications. As with the prior example, the time and space costs follow directly from standard parallel analysis. For energy, we adopt the scheduled work/access baseline (Definition~\ref{def:energy_vn}); for this dense layer the scheduled work scales with the number of multiply-accumulate-like operations, yielding
\begin{equation}
    E_{p, ff} \approx O(N_i N_j).
\end{equation}

\subsubsection{Neuromorphic implementation}

\textit{Time: }
As with the prior example, the idealized NMC time cost of a feed-forward ANN layer is driven by the depth of the circuit, in this case $T_{\infty} \approx 3$ (Lemma~\ref{lem:ideal_nmc_depth_time}; see also Methods, Section \ref{sec:NMCadvantage}).

\textit{Space: }
Unlike the prior example which considered iterative algorithms on a fixed mesh, the dense structure of ANNs represents a challenge for NMC implementations. In the framework, instantiated footprint scales with the size of the algorithm graph (Definition~\ref{def:spaceComplexity}), and for a dense feed-forward layer the inter-neuron communication skeleton is dense: $|S|=\Theta(N_i N_j)$ (Definition~\ref{def:denseFF} in the Methods). Since external memory is not available in the idealized model, this connectivity must be realized in the instantiated graph. Likewise, for feed-forward networks there is no recurrence structure in $G_{ff}$ that allows reuse across time (i.e., $R_T = O(1)$).

\textit{Energy: }
The dense structure of feed-forward ANNs is similarly problematic for energy scaling on NMC as well. Under the event-count model, neuromorphic energy is governed by the trace activity (Definitions~\ref{def:deltaG} and \ref{def:energyComplexity}). For a single forward evaluation, if a constant fraction of synapses are active in each step of the layer, then the synaptic activity satisfies $|\Delta S(t)|=\Theta(|S|)$ (in expectation) and therefore the cumulative activity scales as $\sum_t |\Delta S(t)|=\Theta(|S|T)$, yielding energy proportional to the dense connectivity (Lemma~\ref{lem:nmc_energy_worstcase}). In particular, under sustained activation the Methods formalize that no activity-sparsity advantage is expected for this dense layer family (Proposition~\ref{prop:denseNegative}):
\begin{equation}
    E\!\left[C_E(G_{ff};T)\right] = \Omega(T|S|)=\Omega(T N_i N_j).
\end{equation}

As with space, this dense scaling of energy is likely prohibitive for realizing an ANN advantage of NMC hardware without substantial restructuring. In the framework language, feed-forward dense layers are not naturally well-suited to minimize the execution trace $|\Delta G[0{:}T)|$, because a large fraction of the graph remains active throughout computation. Thus, for dense feed-forward layers, an ideal NMC implementation should not be expected to exhibit an asymptotic energy scaling advantage over conventional implementations unless activity can be made sparse.

While this analysis shows that \textit{direct} mappings of ANNs to NMC platforms should not be expected to provide scaling advantages, the prior analysis illustrates that advantages may be obtained if the computational graph of ANNs can be reshaped suitably. For this reason, research into SNN mappings likely should focus on identifying isomorphic computational graphs that can shift the neuromorphic costs. For instance, an NMC implementation of an ANN that enforces bounded fan-in/fan-out, or induces scale-independent activation sparsity (i.e., $|\Delta S(t)| \ll |S|$), could shift the ANN costs closer to the mesh costs. This goal is similar to conventional methods focused on ANN quantization, pruning, and regularization to minimize conventional resource costs, which can be factored into ANN training methods, however the specific approaches would likely need to be specific to NMC.

An alternative is that NMC solutions should minimize the use of feed-forward layers and focus on recurrent networks. This is consistent with a growing belief in the NMC community that recurrent ANN models, such as echo-state networks \cite{jaeger2001echo}, liquid-state machines \cite{maass2002real}, and more recently state-space models \cite{voelker2019legendre, meyer2025diagonal}, may be more suitable for NMC than feed-forward models. Recurrent networks naturally enable the reuse of components (mitigating the high space costs of NMC), however as described above, it is likely necessary to have either scale-free connectivity (bounded degree) or a convergence of network activity to observe scalable energy advantages. To date it remains an open question whether such recurrent networks can be realized with these energy characteristics while providing competitive performance.

\section{From Ideal to Reality}
The above analysis focuses on ideal implementations of both NMC and conventional computing. While this is reasonable for understanding the theoretical computing landscape, there are significant considerations when looking at the reality of how these NMC architectures are constructed.
The biggest difference between today's NMC systems and the ideal system explored in this framework is how neural computation is realized in a digital architecture. As described in section \ref{section:nmc_mimd}, NMC is conceptually an MIMD architecture; similar conceptually to a massive multi-core CPU system. In practice, today's scalable NMC systems leverage a hierarchy of neuromorphic cores that each are responsible for a certain number of neurons and synapses and these cores generally are moderately programmable with local memory defining the neurons and synapse parameters. Nonetheless, within a core the simulation of neurons and synapses is often serialized. If we consider a system that has $M_{cores}$ separate neuromorphic cores, we can define $N_{core}$ as the number of neurons per core, and $S_{core}$ as the number of synapses per core. In the worst case of synchronous per-tick evaluation within a core, this changes our top speed of neuromorphic system from Lemma \ref{lem:ideal_nmc_depth_time} to the heuristic

\begin{equation}
    T_{inf} \leq T_N \leq N_{core} T_{inf}
\end{equation}

as every irreducible step through the model may require that all neurons on each core are cycled at least once, though it is important to note that these bounds are heuristic and intended only to illustrate how within-core multiplexing can trade footprint for serialization; precise bounds depend on the core’s event-processing model and mapping. At the same time, the total space requirement for NMC will decrease by a comparable factor. Rather than the physical space being directly proportional to the size of the compute graph (Definition \ref{def:spaceComplexity}), it can be scaled down since $N_{core}$ neurons share the same physical compute structure, allowing the space cost to be amortized along the following heuristic,

\begin{equation}
        c_N \frac{T_1}{T_{\infty} N_{core}} \leq C_N \leq c_N T_1 / N_{core}.
    \label{eqn:reality_Space_NMC}
\end{equation}

A related difference that is is that today, in practice, the inter-neuron communication, $S$, of a neuromorphic architecture is not realized as a collection of dedicated point-to-point wires (like axons in the brain); but rather spike events are typically transported over a shared routing infrastructure adapted from conventional networks-on-chip. There are several approaches to this routing, including most commonly varied forms of address-event representation (AER). In practice, shared routing introduces additional constraints and non-idealities that can affect both time and energy scaling, particularly as synaptic activity increases. While these effects are architecture-specific, they can be incorporated into the framework by refining the event-count model that is used to guide complexity analysis. 

Finally, the energy of realized NMC systems depends on numerous factors, including routing hierarchies and embedding. In the idealized framework energy scales with cumulative activity (Definitions~\ref{def:deltaG} and \ref{def:energyComplexity}), but in realized systems the per-event cost can vary substantially depending on whether communication is local (within-core) or routed more globally (cross-core/cross-chip). As such, embedding and communication locality become first-order practical considerations.

\section{Conclusions}
\subsection{Common thread of neuromorphic advantages}

The above examples highlight several important points to consider in the context of the broader analysis described here.

First, there is likely no intrinsic \textit{algorithm-independent} theoretical advantage of NMC. From a complexity perspective, the worst-case energy costs of NMC are equivalent to the expected costs of conventional and the time advantage of neuromorphic is directly related to the number of processors deployed (which in theory could be similarly scaled up for a conventional system). This should be expected---in a very real sense our framework for NMC is equivalent to an ideal parallel computer. That is not to say that there would not be an empirical advantage with scaling; we effectively are contrasting NMC to a PRAM model where all conventional processors have a shared memory that has a scale-free access cost. Nevertheless, as we are considering an ideal NMC system it is reasonable to overlook this limitation of parallel conventional systems.

Second, the illustration of iterative algorithms highlights that a significant theoretical benefit is possible if the task is suitable and the NMC algorithm is appropriately designed. Because dynamic energy in the event-count model scales with realized activity (Definitions \ref{def:energyComplexity}), an NMC algorithm tailored so that its activity decreases as the solver converges (e.g., decreasing $|\Delta N|$ and decreasing $|\Delta S|$ over time, in expectation)can see real benefits in total energy consumption. This not only suggests benefits for iterative numerical methods as suggested here, but has potential benefits for other tasks such as machine learning and optimization algorithms.

Third, this analysis demonstrates that NMC provides at worst comparable scaling to conventional when the computational graphs are equivalent, which is always an option for NMC. While not extensively explored, it has been shown that neuron-inspired primitives (e.g., threshold circuits) can yield more depth-efficient computational graphs \cite{aimone2020provable}. For instance, even simple TG models can permit a logarithmic advantage in depth \cite{parekh2018constant}, which would be directly realized in this analysis as both a time and energy advantage. It is likely that incorporating more complexity into neuron and synapse dynamics can similarly enable the realization of more compact computational graphs. These benefits would build on the baseline advantages and characteristics of NMC scaling demonstrated here. For example, in \cite{smith2022neuromorphic, smith2020solving}, the computational graph developed for Monte Carlo sampling of DTMC models leverages a distinct neuromorphic algorithm to solve the same math. Not only are the empirical advantages seen there due to the decay of activity ($|\Delta N| \rightarrow 0$) as the Monte Carlo simulation progresses, they also benefit from random number generation being similarly local to the processing---a feature that may be further be amplified with the incorporation of true random number capabilities in hardware \cite{misra2023probabilistic}.

Finally, this analysis has been to be fully agnostic to the underlying materials and device implementations of an NMC system. Moving from digital to analog or from silicon to non-silicon devices will have immediate consequences on many of the scaling factors in the above analysis, but this should not change the underlying architectural Big-O scaling. That being said, the analysis above targets an ideal NMC system, and the evolution of NMC towards this ideal parallelism and connectivity may require less conventional materials and devices than those used today. Additionally, moving away from digital devices will almost certainly allow the extension of the low-level capabilities of neurons and synapses, opening the door towards algorithmic advantages (i.e., more efficient $G$). Importantly, this analysis demonstrates the importance that any such exploration of advanced low-level capabilities be made in conjunction with architectural analysis such as this and co-design with algorithm development.

\subsection{Limitations of this analysis}
This analysis focuses primarily on the time, space, and energy complexity of ideal neuromorphic and conventional architectures. While we do briefly consider the implications of realistic implementations, specifically digital NMC systems that leverage neuromorphic cores contrasted to GPUs, this analysis could be extended considerably. Specifically, we only briefly discuss the costs associated with hierarchies of communication in neuromorphic cores and chips and conventional shared and local memories. Memory hierarchies have a significant effect on time and energy costs in conventional systems and thus are an area of considerable research today. Likewise, NMC hierarchies will similarly increase communication and impact time and energy costs for real NMC systems. While we do not expect the core complexity findings of this analysis will be impacted by these hierarchies, further analyses specific to each NMC system will be necessary to understand the points where these costs become prohibitive.

Additionally, this analysis does not consider precision of a computation. In digital systems, precision manifests itself directly in the computational graph $G$, however it can be expected to be an expansion of $G$ that is compressible and thus easily handled within realized conventional systems if required precision is homogeneous. The impact of precision on NMC systems is an open question in NMC algorithm design and represents a potential cost that must be considered its suitability. Further, any move to leverage analog or non-conventional components in NMC systems must consider the impact on precision. Importantly, the precision demands of numerical algorithms are already being challenged in machine learning applications, and it is possible that the increased use of probabilistic approaches identified here may also mitigate some of the need for extreme high precision.

This analysis proposes a concrete hypothesis that the computational graph of an algorithm can be used to identify NMC suitability. This hypothesis merits a deeper investigation both from an empirical perspective---do the actual energy costs of today's NMC systems fit this framework?---and from a theoretical perspective---are the concepts of graph compressibility over space and time explored here suitable to characterize for classes of real applications? Empirical validation could involve benchmarking neuromorphic systems against GPUs for iterative optimization algorithms, measuring energy costs, and analyzing how sparsity impacts performance. The benchmarking of NMC has been a small but growing endeavor \cite{vineyard2022neural, yik2025neurobench}, however these early efforts have been more application-driven as is typical in machine learning \cite{reddi2020mlperf} and it remains an open question how to account for the fundamental architectural differences of NMC. 

Lastly, as mentioned in section \ref{sec:NMC_non-deterministic}, we intentionally do not fully explore stochasticity and learning in this analysis. Ultimately, both stochasticity and learning can be viewed as algorithm design choices, but they may interact uniquely with the fully parallel and event-driven nature of NMC hardware, providing theoretical benefits far beyond what is identified here. For instance, when viewed in terms of the computational graphs analysis above, both stochasticity and learning could be viewed as mechanisms in which a NMC approach can uniquely compress the computational graph. This question of how this non-determinism of NMC algorithms relates to NMC architectures is a significant open question to explore going forward.

The addressing of these limitations is a significant challenge, but future research clarifying these questions will help refine a theoretical framework for NMC that can provide a useful guide to identifying the true potential for neuromorphic approaches to computation.

\subsection{A path forward for neuromorphic computing, NeuroAI, and the brain}

Despite its recent successes in achieving brain-like scales \cite{kudithipudi2025neuromorphic}, NMC still faces a number of challenges to become widely adopted as a complementary platform for numerical computing and AI. NMC systems remain difficult to program, with no easily accessible equivalent to PyTorch or a lower-level interface like CUDA. While there are emerging efforts to identify common programming frameworks \cite{bekolay2014nengo, pedersen2024neuromorphic, eshraghian2021training, aimone2019composing}, the rapid growth of the field---both in terms of hardware and algorithms---makes it difficult for the broader community to converge on a common programming models that are robust to new hardware and applications yet make neural computation more broadly accessible.

Additionally, the diversity of NMC hardware looks likely to continue to grow. Due to the slowing of Moore's Law, conventional CMOS scaling looks unlikely to continue indefinitely, meaning that non-CMOS materials and alternative architectural approaches, including analog and optical, may be required to achieve human-brain scales at reasonable space and energy footprints. While the analysis here suggests that these approaches are unlikely to change the fundamental theoretical advantages of neuromorphic, they may make it more likely to realize the idealized benefits described here in future systems.

Despite these challenges going forward, the analysis in this paper demonstrates that there is a benefit to begin identifying algorithms that can benefit from NMC. Strengthening this fundamental understanding of what makes a strong NMC algorithm will be critical going forward to help overcome the future hardware and programming challenges mentioned above. The potential necessity for probability theory and graph analytics to influence this algorithm design places NMC research as much in the domain of computer science and applied mathematics as in its traditional fields of electrical engineering and physics. Additionally, from a neuroscience perspective it can be argued that we are at an early stage of leveraging brain-inspiration for computing and AI \cite{aimone2019neural}. These opportunities suggest that through suitable algorithm design and focused brain-inspiration, NMC can help influence a revival of energy-efficient numerical computing and AI solutions. 

Unlike proposed theoretical frameworks seeking to globally relate the brain's dynamics and anatomy to computational goals \cite{friston2010free, hawkins2021thousand}, the framework proposed here does not aim to make predictions or claims about the brain's function. Rather, one of its goals is to provide a mechanism by which biological processes can potentially be analyzed with rigor and related to computation proper. While there are a few exceptions, the framework's partitioning of biophysical dynamics into isolated neuron components is capable of directly representing neurobiological processes far more complex than the neuron and synapse dynamics in today's NMC hardware. For example, biological learning is immensely complex and is often coupled to a diverse set of neuromodulators often ignored in today's NMC algorithms \cite{aimone2019neural}. Such processes could be handled by a richer and heterogeneous synaptic processing elements ($n_S$). As this framework demonstrates, such design choices should not change the overall structure of the analysis (algorithm activity trace accounting) though they can potentially open doors for more powerful neural algorithm design leveraging increased neuron capabilities (Section \ref{sec:NMCadvantage}). Furthermore, those processes that are not immediately realizable---such as astrocytes, gap junctions, and developmental and adult neurogenesis---likely can be accounted for with extensions of the present formalism. 

With the rise of large-language models and the perceived urgency of energy-efficient AI at scale, the debate about how much neural fidelity is required for NMC to reach its potential has been increasing in recent years \cite{indiveri2025neuromorphic, ou2022overview, yan2024reconsidering}. As this debate has largely lacked a formal foundation, there remain unresolved disagreements about how to prioritize emerging hardware, how to evaluate spike-based AI algorithms, whether to encode information temporally or spatially in neurons, and how much to prioritize scale compared to biological fidelity. While the framework presented here does not propose answers to these ongoing debates, the framework presents a principled path by which the advantages of brain-inspiration can be investigated.

\clearpage
\section*{Methods}
\addcontentsline{toc}{section}{Methods}

\makeatletter

\newcounter{subsubsubsection}[subsubsection]

\renewcommand\thesubsubsubsection{\thesubsubsection.\arabic{subsubsubsection}}
\newcommand\subsubsubsection{\@startsection{subsubsubsection}{4}{\z@}%
  {3.25ex \@plus 1ex \@minus .2ex}%
  {1.5ex \@plus .2ex}%
  {\normalfont\normalsize\bfseries}}

\renewcommand{\thesubsection}{M\arabic{subsection}}
\renewcommand{\thesubsubsection}{M\arabic{subsection}.\arabic{subsubsection}}
\renewcommand{\thesubsubsubsection}{%
  M\arabic{subsection}.\arabic{subsubsection}.\arabic{subsubsubsection}%
}
\setcounter{subsection}{0}

\renewcommand{\theequation}{\thesubsection.\arabic{equation}}

\makeatother
\setcounter{secnumdepth}{4}
\renewcommand{\thefigure}{\thesubsection.\arabic{figure}}
\renewcommand{\thetable}{\thesubsection.\arabic{table}}

\setcounter{equation}{0}
\setcounter{statement}{0}

\subsection{Neuromorphic Algorithm Preliminaries} 

\label{sec:preliminaries}

\subsubsection{Neurons and Synapses}
\label{sec:preliminaries_NeuronsSynapses}
First, we provide generic formal definitions of the two primary classes of objects in a NMC system, \textit{neurons} and \textit{synapses}. Neurons and synapses are explicit physics concepts in the brain, but these terms have taken a broader meaning in computing fields. A basic aim of this framework is to preserve generality and ensure that specific models, regardless of being from the ANN, NMC, or neurobiological perspectives, can be represented. 

\begin{figure}
    \centering
    \includegraphics[width=.95\linewidth]{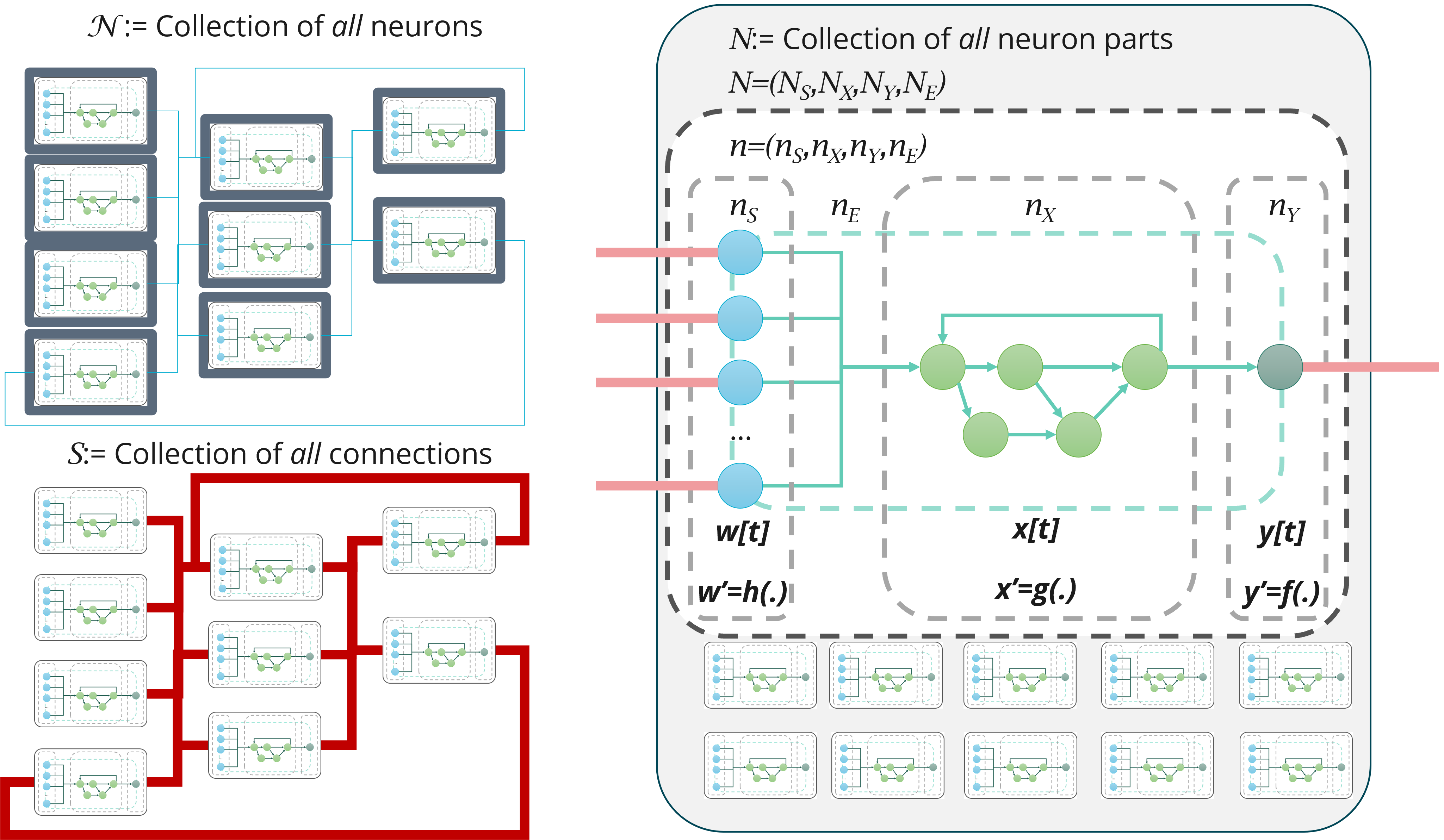}
    \caption{Illustration of structural model, $G_N=(N,S)$ of neural graphs used in framework. For a neural algorithm $G_N=(N,S)$, $\mathcal{N}$ (top left) represents the set of all neurons, $S$ (bottom left) represents the set of all synaptic connections, and $N$ (right) represents the set of the neuron parts which are responsible for \textit{all} of the computation in the neural algorithm, including $N_S$ (all synaptic computation), $N_X$ (all internal neuronal dynamics), and $N_Y$ (all spiking dynamics), as well as $N_E$ which describes internal neuronal communication.}
    \label{fig:structure_G_N}
\end{figure}

Figure \ref{fig:structure_G_N} illustrates these concepts.

\begin{definition}[Neuron]
    A \textbf{neuron} is a compute element represented as a (potentially cyclic) directed subgraph 
    \(n= (n_S, n_X, n_Y, n_E) \), where: 
    
    \begin{itemize}
        \item \( n_S \): Synaptic input nodes, representing incoming spiking signals from other neurons (\( \textbf{s}[t] \)) and exhibiting synaptic dynamics related to the representation ($w[t]$), and evolution ($h(\cdot)$) of synaptic weight, synaptic delay, stochasticity, and synaptic plasticity.
        \item \( n_X \): Internal compute nodes describing the neuron's internal state variables ($x[t]$) and the neuron's internal state dynamics ($g(\cdot)$).
        \item \( n_Y \): Output nodes, representing the neuron's observable output ($y[t]$) and the generation of that output ($f(\cdot)$). For spiking neurons, \(y[t] \in \{0,1\}\).
        \item \( n_E \): Internal edges, representing directed connections between \( n_S \), \( n_X \), and \( n_Y \). The internal edges \(n_E\) may contain self-loops and directed cycles to represent intrinsic dynamics (e.g., decay, refractoriness, or adaptation).
    \end{itemize}
    
    The neuron's computation is characterized by a non-linear transfer function \(f(\cdot)\), such that \(y[t]=f(x[t])\), and the internal state \(x[t]\) evolves according to some (discrete-time) dynamic, such as
    \[
        x[t+1] = g(x[t], \textbf{w}[t]),
    \]
    where \(\textbf{w}[t]\) describes the set of synaptic spike-event inputs for the neuron at time step \(t\) represented at the $n_S$ nodes.

    Synaptic weight and all analog synaptic behavior are considered post-synaptic components belonging to the neuron and are represented within the neuron's synaptic input nodes \(n_S\) (and, more generally, within the neuron's internal state). Learning represents the change of the post-synaptic synaptic state over time, e.g.,
    \[
    \frac{d w_{ij}(t)}{d t} = h_{\text{learn}}(w_{ij}(t), y_i(t), x_j(t)),
    \]
    or, in discrete time,
    \[
    w_{ij}[t+1] = h_{\text{learn}}(w_{ij}[t], y_i[t], x_j[t]).
    \]
    \label{def:neuron}
\end{definition}

This definition of neuron is generic and inclusive of any artificial neuron type. For instance, for TGs or McCulloch--Pitts neurons \(f(\cdot)\) is a Heaviside function; for modern ANNs \(f(\cdot)\) is often a sigmoid or rectified linear function. The complexity of \(f(\cdot)\) and \(g(\cdot)\) determines the size and structure of the neuron subgraph \(n\). For simple neuron models (e.g., LIF neurons), \(n_X\) may consist of a single node, while for more complex models, \(n_X\) may represent a larger graph of internal dynamics.

As clarity, we refer to the set of isolated neuron subgraphs as \(\mathcal{N}\), where each \(n \in \mathcal{N}\) is an individual neuron, and \(N\) represents the collection of neuron components. The set of isolated neuron subgraphs \(\mathcal{N}\) fully describes the graph of neuron compute nodes \(N\), such that:
\[
N_S = \biguplus_{n\in\mathcal{N}} n_S,\quad
N_X = \biguplus_{n\in\mathcal{N}} n_X,\quad
N_Y = \biguplus_{n\in\mathcal{N}} n_Y,\quad
N_E = \biguplus_{n\in\mathcal{N}} n_E.
\]

Finally, we define the total number of neurons as $M$, such that
\[
M := |\mathcal{N}|
\]

\begin{definition}[Synapse]
    Let \( \mathcal{N} \) represent the set of isolated neuron subgraphs in a neuromorphic system, where each \textbf{synapse} is a directed synaptic communication channel from a source neuron \( n_i \in \mathcal{N} \) to a target neuron \( n_j \in \mathcal{N} \). We take the set of synaptic communication channels to be a set of directed edges
    \[
        S \subseteq N_Y \times N_S,
    \]
    where \(N_Y=\bigcup_{n\in\mathcal{N}} n_Y\) and \(N_S=\bigcup_{n\in\mathcal{N}} n_S\).
    
    Each synapse connects an output node \( n_{y_i} \in n_{i,Y} \) of the source neuron \( n_i \) to a synaptic input node \( n_{s_{ij}} \in n_{j,S} \) of the target neuron \( n_j \) via a directed edge \( s_{ij} \in S \), where:
    \begin{itemize}
        \item \( n_{y_i} \): The output node of the source neuron \( n_i \), representing the observable spiking output \( y_i[t] \in \{0,1\} \).
        \item \( n_{s_{ij}} \): The synaptic input node of the target neuron \( n_j \), representing post-synaptic state associated with the connection, including synaptic weight \(w_{ij}[t]\) and any synaptic dynamics (e.g., delay, stochasticity, filtering, and plasticity).
        \item \( s_{ij} \): The directed edge connecting \( n_{y_i} \) to \( n_{s_{ij}} \), representing discrete spike-event transmission between neurons, \(s_{ij}[t] \in \{0,1\}\).
    \end{itemize}
    The set of synaptic communication channels \(S\) fully describes the inter-neuron communication edges in the neural system.
    \label{def:synapse}
\end{definition}

The spike-event transmission on a synaptic communication channel, \(s_{ij}[t]\), is typically defined as a function of the pre-synaptic neuron's observable output \(y_i[t]\), such as \(s_{ij}[t] = y_i[t]\) (or, more generally, \(s_{ij}[t] = \tau(y_i[t])\) to account for communication constraints). 

This definition of a synapse is similarly generic and inclusive of any spiking synaptic communication model; for example, multi-bit communication between two neurons may be represented by multiple synaptic communication channels (i.e., multiple edges) between the neurons, though this is generally not desirable. In this framework, synapses are treated as discrete communication channels in \(S\); all synapse-specific state and analog dynamics are post-synaptic and belong to the target neuron's synaptic input nodes \(n_S\).

For later accounting, we represent the neuron-owned portion as a typed directed graph \(N=(N_S,N_X,N_Y,N_E)\), where \(N_S\), \(N_X\), and \(N_Y\) are synaptic input, internal state, and output nodes, and \(N_E\) contains edges internal to neuron subgraphs. Neuron subgraphs may be cyclic to capture stateful dynamics. By construction, each synaptic communication channel targets a unique synaptic input node, so \(|N_S|=|S|\).

\begin{remarkn}(Broadcast spike semantics)
For a spiking neuron with observable output \(y_i[t]\in\{0,1\}\), we assume that all outgoing synaptic communication channels sourced at the same output node \(n_{y_i}\in N_Y\) carry the same binary event at time \(t\), i.e., for all targets \(j\) with \(s_{ij}\in S\), \(s_{ij}[t]=\tau(y_i[t])\). Thus, spike transmission is broadcast/multicast: when \(y_i[t]=1\), all outgoing channels from \(n_{y_i}\) are simultaneously active; when \(y_i[t]=0\), none are.
\label{rmk:broadcast}
\end{remarkn}

\begin{remarkn}(Delay-augmented spike communication)
Although we treat inter-neuron communication as discrete and reliable spike events ($s_{ij}[t]\in\{0,1\}$), many spiking neuromorphic algorithms additionally exploit \emph{programmable transmission delay} to encode problem structure or numerical values in spike timing (e.g., shortest-path and dynamic-programming constructions \cite{aimone2019dynamic,aimone2021provable}). To accommodate this without changing the discrete-event nature of communication, we allow each synaptic communication channel $s_{ij}\in S$ to carry an associated nonnegative integer delay parameter $\delta_{ij}\in\mathbb{N}$ (or, equivalently, we allow the post-synaptic synaptic input node $n_{s_{ij}}$ to implement a delay line as part of neuron-local state). In this paper we treat $\delta_{ij}$ as a hardware- and implementation-dependent parameter with bounded resolution and range; detailed models of delay precision, jitter, and routing-dependent latency are beyond the scope of this initial analysis.
\end{remarkn}

Table \ref{tab:neuron_types} shows how governing functions often used for neurons and synapses fit this framework. Note that the framework described here is inclusive of most standard governing functions for neurons and synapses such so long as they can be localized within individual neurons.

Additionally, a note about stochasticity, which is common in many neuromorphic algorithms and architectures. Like all other computational dynamics, in this model stochasticity is entirely represented \textit{within} neurons, whether stochastic synaptic release (included within $n_S$) or neuronal dynamics (included within $n_X$). All communication in $S$ is fully deterministic in this framework.

\begin{definition}[Local stochasticity primitive]
\label{def:local_stochasticity}
A neuromorphic algorithm \(G_N=(N,S)\) has \emph{local stochasticity} if each neuron subgraph \(n\in\mathcal{N}\) may access an independent local random source \(r_n[t]\) (true or pseudo-random), and neuron/synapse updates may depend on \(r_n[t]\) without requiring any additional inter-neuron communication beyond \(S\). Formally, for each neuron \(n_j\),
\[
x_j[t+1]=g(x_j[t], w_j[t], r_{n_j}[t]), \qquad
y_j[t]=f(x_j[t], r_{n_j}[t]),
\]
and any stochastic synaptic rule is implemented post-synaptically within \(n_{j,S}\) and \(n_{j,X}\).
\end{definition}

\begin{table}[t]
\centering
\caption{Neuron and synapse functions for different neural computing frameworks.}
\label{tab:neuron_types}
\renewcommand{\arraystretch}{1.25}
\setlength{\tabcolsep}{4pt}
\small
\begin{tabularx}{\linewidth}{>{\Centering\arraybackslash}p{2.6cm} Y Y Y}
\toprule
\textbf{Model} &
\makecell{\textbf{Output}\\\(y'[t]=f(\cdot)\)\\(\(n_Y\))} &
\makecell{\textbf{State update}\\\(x'[t]=g(\cdot)\)\\(\(n_X\))} &
\makecell{\textbf{Post-synaptic learning}\\\(w'[t]=h(\cdot)\)\\(\(N_S\))} \\
\midrule

\makecell{Threshold\\gate (TG)} &
\makecell{\(y_j=\mathbf{1}[x_j>\theta]\)} &
\makecell{\(x_j=\sum_i w_{ij}s_{ij}\)} &
\makecell{\(---\)} \\
\addlinespace

\makecell{Artificial\\neural network} &
\makecell{\(y_j=\phi(x_j)\)\\\(\phi(x)=\max(0,x)\) or \(\tanh(x)\)} &
\makecell{\(x_j=\sum_i w_{ij}s_{ij}\)} &
\makecell{\(w'_{ij}=w_{ij} - \eta \delta_jy_i\)} \\
\addlinespace

\makecell{Leaky\\integrate-and-fire} &
\makecell{\(y_j[t]=\mathbf{1}[x_j[t]>v_{\mathrm{th}}]\)} &
\makecell{\(x_j[t{+}1]=\alpha x_j[t]+\sum_i w_{ij}[t]s_{ij}[t]\)\\
if \(x_j[t]\le v_{\mathrm{th}}\);\\
\(x_j[t{+}1]=v_{\mathrm{reset}}\) otherwise\\
(\(\alpha=e^{-\Delta t/\tau}\))} &
\makecell{\(w'_{ij}[t]=\eta(y_j[t-d_{ij}]tr_i[t]\)\\\(-y_i[t-d_{ij}]tr_j[t])\)} \\
\bottomrule
\end{tabularx}
\end{table}

\subsubsection{Neuromorphic algorithms and architecture}
\label{sec:preliminaries_algorithms}

While there is variability in the programmability of NMC systems, we propose here that a generic NMC architecture is one which restricts the computation of neurons and synapses to a targeted set of functions. Such NMC systems are generically programmable, but they are not stored-program architectures in the von Neumann sense. While hardware specifics vary, once configured an NMC system generally does not fetch and execute a serial instruction stream from an external memory during operation. Instead, the ``program'' is directly encoded in the structure and parameters of the instantiated neural circuit: the neurons and synaptic communication channels that realize the algorithm. 

This distinction has two immediate implications. First, the full algorithm must be spatially realized across the hardware fabric, so the footprint of an NMC implementation is tied to the size and structure of the instantiated graph rather than to a fixed processor pool. Second, because the circuit is fixed during execution, the dominant overhead associated with programmability is the cost of instantiating/loading the circuit; consequently, calculations that repeatedly reuse the same circuit structure can amortize this cost. In neuromorphic algorithms such reuse arises naturally through feedback connections between neurons (recurrence), which allows the same neuron resources to be reused over many computational steps without ``unrolling'' the computation into a larger feed-forward graph. The implications of recurrence will be described in Section \ref{sec:NMCadvantage}.

We can define neural architectures in two ways: as a collection of isolated neurons \(\mathcal{N}\) and synaptic communication channels \(S\), or as collections of neuron components \(N\) and synaptic communication channels \(S\). In this analysis, we focus on the latter representation, as increasingly NMC architectures are flexible in how neurons are defined from a fixed pool of configurable neuron components. Additionally, isolating inter-neuron communication entirely within \(S\) provides a modular framework for analyzing synaptic communication and energy scaling, which will be explored in detail later.

\begin{definition}[Neuromorphic Computing Architecture]
    A \textbf{neuromorphic computing architecture} is a distributed computing system capable of representing a set of neurons \(\mathcal{N}\) and a set of synaptic communication channels \(S\) as a directed graph \(G_{N,\text{arch}} = (N_{\text{arch}}, S_{\text{arch}})\), where:
    \begin{itemize}
        \item \(N_{\text{arch}}\): The union of neuron components of the set of isolated neuron subgraphs \(\mathcal{N}\), where each neuron \( n \in \mathcal{N} \) is represented as \( n = (n_S, n_X, n_Y, n_E) \), as defined in Definition \ref{def:neuron}.
        \item \(S_{\text{arch}}\): The set of directed synaptic communication channels between neuron subgraphs, where each \( s_{ij} \in S_{\text{arch}} \) connects a source output node \(n_{y_i}\) to a target synaptic input node \(n_{s_{ij}}\), as defined in Definition \ref{def:synapse}.
    \end{itemize}
    The architecture consists of specialized circuitry to directly compute the governing functions for neurons (\( f(\cdot) \), \( g(\cdot) \)) and to store and update all required neuron-local state, including post-synaptic synaptic state (e.g., \(w_{ij}[t]\)) and any learning rules (e.g., \(h(\cdot)\)). All required memory for parameterization of neurons and synapses is co-located with the circuitry required for their computation.
    \label{def:architecture}
\end{definition}

Even though their governing functions can vary, for any set of governing functions we can define a neural algorithm as a graph of neurons and synaptic communication channels. 

\begin{definition}[Neuromorphic Algorithm]
    A \textbf{neuromorphic algorithm} is an algorithm implemented on a neuromorphic computing architecture \(G_{N,\text{arch}} = (N_{\text{arch}}, S_{\text{arch}})\). The algorithm is specified by (i) a selection of an algorithm subgraph \(G_{N,\text{alg}} = (N_{\text{alg}}, S_{\text{alg}})\) and (ii) a parameterization of the selected neuron-local state and update rules, where:
    \begin{itemize}
        \item \(N_{\text{alg}}\): The combined graph of selected neuron subgraphs, representing the computational nodes of the algorithm.
        \item \(S_{\text{alg}}\): The selected set of synaptic communication channels, representing the directed edges between distinct neurons.
    \end{itemize}
    The algorithm graph \( G_{N,\text{alg}} = (N_{\text{alg}}, S_{\text{alg}}) \) represents the ideal computational graph for the algorithm, with
    \[
        N_{\text{alg}} \subseteq N_{\text{arch}}, \quad S_{\text{alg}} \subseteq S_{\text{arch}}.
    \]
    \label{def:algorithm}
\end{definition}

For the purposes of this analysis, unless noted otherwise $G_N=(N,S)$ will refer to a neuromorphic algorithm.

\subsubsection{Neuromorphic Analysis Preliminaries}
\label{sec:alg_prelim}
Because both the target hardware and the algorithm representation change between conventional and neuromorphic systems, we introduce an intermediate (cross-platform) representation to enable fair comparative analysis. Our intent is to remain agnostic to the physical substrate of neurons/synapses and to specific neuron/synapse models, while retaining a graph-level description sufficient for time, space, and energy scaling.

Let \(G_N=(N,S)\) be a neuromorphic algorithm graph. To align with conventional notation \(G=(\mathcal{V},\mathcal{E})\), we may equivalently write \(G_N=(\mathcal{V}_N,\mathcal{E}_N)\) with
\[
\mathcal{V}_N = N_S \uplus N_X \uplus N_Y, \qquad \mathcal{E}_N = N_E \uplus S,
\]
where \(N_E\) are edges internal to neuron subgraphs and \(S\) are inter-neuron communication channels.

Given a conventional reference algorithm \(G=(\mathcal{V},\mathcal{E})\) and a computationally isomorphic neuromorphic algorithm \(G_N=(N,S)\), we define:
(i) a conventional simulation \(G^*=(\mathcal{V}^*,\mathcal{E}^*)\) of \(G_N\), and
(ii) a neuromorphic simulation \(G_N^*=(N^*,S^*)\) of \(G\).
In general there may be many valid choices for \(G^*\) and \(G_N^*\).

\begin{definition}[Intermediate Computational Graph]
\label{def:intermediateGraph}
Let \(G=(\mathcal{V},\mathcal{E})\) be a conventional computational graph and \(G_N=(N,S)\) a computationally isomorphic neuromorphic graph. An \textbf{intermediate computational graph} \(G^*=(\mathcal{V}^*,\mathcal{E}^*)\) is a conventional computational graph that simulates the dynamics of \(G_N\) and is computationally isomorphic to both \(G\) and \(G_N\).
\end{definition}

We define time, space, and energy complexity measures in the following sections. To compare computationally isomorphic algorithms across architectures, we use ratios:

\begin{definition}[Relative Complexity Gain]
\label{def:relativeGain}
Let \(G\) and \(G'\) be computationally isomorphic algorithms executed on a common problem family (and horizon \(T\), where applicable). For \(X\in\{T,S,E\}\), define the \textbf{relative complexity gain}
\[
\rho_X(G,G') := \frac{C_X(G')}{C_X(G)}.
\]
\end{definition}

\begin{definition}[Neuromorphic Advantage]
\label{def:nmc_advantage}
Let \(G\) and \(G_N\) be computationally isomorphic algorithms for conventional and neuromorphic hardware, respectively. We say \(G_N\) exhibits a \textbf{neuromorphic advantage} on a problem family if
\[
\rho_S(G,G_N)\in O(1),\qquad \rho_T(G,G_N)\in O(1),\qquad \rho_E(G,G_N)\in O(1),
\]
and for at least one \(X\in\{T,S,E\}\),
\[
\rho_X(G,G_N)\in o(1).
\]
\end{definition}

\newpage
\subsection{Complexity Analysis of Neuromorphic}
We now use the framework from Section~\ref{sec:alg_prelim} to make concrete statements about how neuromorphic formulations scale in time, footprint, and energy relative to conventional stored-program execution. The first set of results is intentionally simple: in the idealized in-memory limit, neuromorphic execution is best understood through the same lens as parallel computation, so the natural quantities for time and footprint are work and depth. We then address the part that motivates neuromorphic computing broadly---energy. In an event-driven neuromorphic system, energy is governed by which updates and communication events actually occur (the execution trace), not by a predetermined instruction schedule, and this trace dependence is where the most meaningful distinctions emerge.

\label{sec:complexityAnalysis}
\subsubsection{Time--space complexity on neuromorphic}
\label{sec:time_space}
We first look to use the framework to understand the relationship of neuromorphic time--space scaling with conventional parallel time--space scaling. As an in-memory computing system, neuromorphic hardware enables highly distributed processing; however, as a classical computing substrate, it does not in general provide extreme (e.g., super-polynomial) time or space complexity advantages compared to conventional computation. Rather, neuromorphic hardware should be understood in relation to parallel computation limits and the tradeoffs between time and space (resources).

To frame this analysis, we first define time- and space- complexity for the purposes of our analysis.

\begin{definition}[Time Complexity]
Let $G = (\mathcal{V}, \mathcal{E})$ represent a computational graph (conventional or neuromorphic). Let $T_1(G)$ denote the total work (serial time) and let $T_{\infty}(G)$ denote the depth (critical-path length) of the computation for a chosen execution semantics. For a parallel system with $p$ processors, the time complexity $C_T(G;p)$ (parallel time) satisfies the lower bound
\[
C_T(G;p) \ge \max\!\left(T_{\infty}(G), \frac{T_1(G)}{p}\right),
\]
and by Brent's Theorem \cite{brent1974parallel}, there exists a schedule achieving the upper bound
\[
C_T(G;p) \le \frac{T_1(G)}{p} + T_{\infty}(G).
\]
\label{def:timeComplexity}
\end{definition}

We can then apply this definition to neurons and synapses, which allows us to see how neuromorphic depth is impacted by Brent's theorem.

\begin{lemma}[Ideal neuromorphic depth time in the fully-parallel limit]
\label{lem:ideal_nmc_depth_time}
Let $G_N$ be a neuromorphic algorithm graph executed for a horizon $T$ under a fixed execution semantics, and let $G_N^{(T)}$ denote its time-unrolled computational graph (Definition~\ref{def:unroll}. Suppose an \emph{idealized} neuromorphic substrate in which (i) each vertex update in $G_N^{(T)}$ has unit time cost, (ii) all vertices at the same dependency level in $G_N^{(T)}$ can be updated concurrently (i.e., sufficient physical parallelism exists to avoid resource contention), and (iii) communication latency is either absorbed into the chosen semantics or bounded by a constant per edge event. Then the runtime of the execution is depth-limited:
\[
C_T(G_N;T) = \Theta\!\left(T_{\infty}\!\left(G_N^{(T)}\right)\right).
\]
This statement is with respect to the specific instantiated neuromorphic graph $G_N$ (via its unrolling $G_N^{(T)}$); it does not assert that $T_\infty(G_N^{(T)})$ is minimal over all possible algorithms for the underlying problem family. In particular, in the fully-parallel limit, $C_T(G_N;T)$ matches the critical-path length of the induced computation.
\end{lemma}

\begin{proof}
Any correct execution must respect the dependency partial order encoded by the time-unrolled graph $G_N^{(T)}$, so no schedule can complete in fewer than $T_{\infty}\!\left(G_N^{(T)}\right)$ unit-time vertex-update rounds. Under assumptions (i)–(iii), a level-by-level schedule that updates all currently-enabled vertices in parallel is feasible and completes in exactly $T_{\infty}\!\left(G_N^{(T)}\right)$ rounds (up to constant factors if edge events incur bounded constant latency). This gives the matching upper bound and therefore the $\Theta(\cdot)$ result.
\end{proof}

\begin{definition}[Resource footprint (space complexity)]
Let $G = (\mathcal{V}, \mathcal{E})$ represent a computational graph (conventional or neuromorphic). The \textbf{resource footprint} (space complexity for the purposes of this analysis) $C_S(G)$ is defined as the physical resources required to represent the computation and its dependencies, including both storage and communication structures:
\[
C_S(G) = c_{\mathcal{V}} |\mathcal{V}| + c_{\mathcal{E}} |\mathcal{E}|,
\]
where $c_{\mathcal{V}}$ is the per-node resource cost (including storage of node-local state and any required compute structures), $c_{\mathcal{E}}$ is the per-edge resource cost (including storage of edge attributes and required communication structures), and $|\mathcal{V}|$ and $|\mathcal{E}|$ represents the total number of elements in $\mathcal{V}$ and $\mathcal{E}$, respectively.
\label{def:spaceComplexity}
\end{definition}

As a convention we will define
\[
|G| := |V| + |E|
\]
for any graph $G=(V,E)$, and accordingly,
\[
|G_N|:=|N_S|+|N_X|+|N_Y|+|N_E|+|S|
\]
for the neuromorphic graphs $G_N=(N,S)$ explored here. The footprint $C_S(G)$ is a weighted version of size.

This definition of resource footprint is generic to whether the underlying architecture is Von Neumann or neuromorphic, however it can be specialized to each. For a Von Neumann architecture, the number of processors $p$ is typically fixed, and the algorithm graph $G$ is represented in memory and executed through a schedule over that fixed processor pool. In contrast, for a neuromorphic architecture the number of physical compute elements grows with the realized neuromorphic graph, so that (in the idealized in-memory limit) effective parallelism can scale with the size of the instantiated graph.

\begin{remark} (Depth-limited computation and spatialization)
In the idealized in-memory limit, neuromorphic time can approach the depth bound \(T_{\infty}(G)\) because effective parallelism scales with the instantiated graph. In architectures where communication delay is a dominant physical constraint (e.g., biological cortex), this naturally encourages \emph{spatialization}: restructuring computations to reduce depth (number of sequential synaptic/communication stages) by expanding the number of concurrent compute elements per stage. This is a classical depth--size tradeoff (circuit depth reduction at increased size), and it does not change computability or worst-case complexity classes; rather, it changes the Pareto frontier between footprint and time in a regime where depth is the primary binding constraint.
\end{remark}

The results below formalize two complementary points. First, any conventional bounded-fan-in circuit can be mapped to a neuromorphic graph with only constant-factor footprint overhead (Proposition \ref{prop:naiveSimulation}), establishing that neuromorphic is algorithmically universal under a fixed primitive class. Second, any neuromorphic execution can be simulated conventionally with at most a multiplicative overhead factor $\gamma(|G_N|)$ associated with representing and scheduling event-driven activity (Proposition \ref{prop:conventionalSimOfNeuromorphic}), implying that asymptotic time improvements are not uniquely neuromorphic (Corollary \ref{cor:timeAdvTransfers}). Together, these results motivate treating ideal neuromorphic execution as operating near the depth limit of parallel computation, with primary differentiators arising from footprint and trace-dependent energy.

\begin{proposition}[Existence of a naive neuromorphic simulation with constant-factor size overhead]
Let $G=(\mathcal{V}, \mathcal{E})$ be a computational graph representing an algorithm on conventional hardware described as a bounded-fan-in Boolean circuit (i.e., $G$ is already given in circuit form). Assume a neuromorphic architecture whose neuron primitives include a constant-size threshold-gate realization with bounded fan-in and whose synaptic communication channels transmit binary spike events. Then there exists at least one (naive) neuromorphic \emph{simulation} of $G$ on neuromorphic hardware, denoted
\[
G_N^* = (N^*,S^*),
\]
that is functionally equivalent to $G$ with size scaling
\[
|G_N^*| \le k\,|G|,
\]
where $k$ is a constant depending only on the chosen conventional gate/operation basis and the neuromorphic primitive set, and $|G|$ and $|G_N^*|$ denote the sizes (number of nodes plus edges, or another consistent size measure) of the computational graphs $G$ and $G_N^*$, respectively.
\label{prop:naiveSimulation}
\end{proposition}

\begin{proof}
We construct a \textit{naive neuromorphic simulation} $G_N^*=(N^*,S^*)$ of the conventional circuit $G=(\mathcal{V},\mathcal{E})$ by mapping each operation node $v\in\mathcal{V}$ to a constant-size collection of neuromorphic primitives that implement an equivalent bounded-fan-in threshold gate (or equivalent primitive in the assumed basis). Because the neuromorphic architecture supports a constant-size realization of the chosen primitive with bounded fan-in, each $v$ can be implemented by at most $k_0$ neuron-owned nodes/edges in $N^*$ for some constant $k_0$, and each conventional dependency edge $e\in\mathcal{E}$ can be implemented by at most $k_1$ synaptic communication channels (spike edges) in $S^*$ for some constant $k_1$.

Therefore the total size of the realized neuromorphic graph satisfies
\[
|G_N^*| \le k_0|\mathcal{V}| + k_1|\mathcal{E}| \le k(|\mathcal{V}|+|\mathcal{E}|),
\]
for $k=\max(k_0,k_1)$. Since $|G|$ is proportional to $|\mathcal{V}|+|\mathcal{E}|$ under a consistent size measure, we obtain $|G_N^*|\le k|G|$.
\end{proof}

Note that Proposition~\ref{prop:naiveSimulation} is a \textit{circuit-to-circuit embedding} result: it shows that, under a fixed neuron primitive class, neuromorphic graphs can simulate bounded-fan-in Boolean circuits with only constant-factor size overhead. Importantly, it does \emph{not} claim that an arbitrary conventional (RAM/Von Neumann) algorithm can be converted to such a circuit without additional overhead; RAM-to-circuit transformations may incur nontrivial (even polynomial) overhead depending on word size, uniformity, and memory-access assumptions.

\begin{proposition}[Conventional simulation of a neuromorphic algorithm (time overhead)]
\label{prop:conventionalSimOfNeuromorphic}
Let \(G_N=(N,S)\) be a neuromorphic algorithm executed for \(T\) discrete time steps under a chosen execution semantics. Let \(|G_N|\) denote a consistent size measure of the instantiated neuromorphic graph. Then there exists a conventional computational graph \(G^*=(\mathcal{V}^*,\mathcal{E}^*)\) that simulates the execution of \(G_N\) such that its parallel time on \(p\) processors satisfies
\[
C_T(G^*;p)\;=\;O\!\left(\frac{T_1(G^*)}{p}+T_{\infty}(G^*)\right),
\]
with
\[
T_1(G^*) \;=\; O\!\left(T\cdot |G_N|\cdot \gamma(|G_N|)\right),
\]
where \(\gamma(\cdot)\) captures the overhead of representing and scheduling sparse event-driven updates on a conventional platform (e.g., \(\gamma\in O(1)\) for dense synchronous scans, or \(\gamma\in O(\log |G_N|)\) for priority-queue or event-queue based implementations).
\end{proposition}

\begin{proof}
A conventional simulator can represent the neuron-local state associated with \(N\) and the inter-neuron connectivity \(S\) in memory. For each discrete step \(t\), the simulator must (i) update the neuron-local dynamics for the relevant neuron components and (ii) deliver any spike events along edges in \(S\). In the worst case (dense synchronous evaluation), this corresponds to performing \(O(|G_N|)\) constant-time update/communication operations per step, yielding \(T_1(G^*)=O(T|G_N|)\). More generally, when using data structures to process sparse event sets, an additional overhead factor \(\gamma(|G_N|)\) is incurred per processed update/event (e.g., due to indexing, queue management, indirect addressing), yielding \(T_1(G^*)=O(T|G_N|\gamma(|G_N|))\). 
\end{proof}

\begin{corollary}[Asymptotic time improvements are not uniquely neuromorphic]
\label{cor:timeAdvTransfers}
Let \(G\) be a conventional baseline algorithm for a problem family, and let \(G_N\) be a correct neuromorphic algorithm for the same problem family executed for a horizon \(T\). If \(G_N\) achieves an asymptotic time complexity improvement over \(G\) (under comparable input/output specifications), then a conventional simulation \(G^*\) of \(G_N\) also achieves the same asymptotic improvement up to the simulation overhead factor \(\gamma(|G_N|)\) of Proposition ~\ref{prop:conventionalSimOfNeuromorphic}. In particular, whenever \(\gamma(|G_N|)\) is subpolynomial and does not asymptotically dominate the improvement, the existence of a faster neuromorphic algorithm implies the existence of a faster conventional algorithm.
\end{corollary}

\begin{proof}
By Proposition~\ref{prop:conventionalSimOfNeuromorphic}, \(G_N\) can be simulated on a conventional system with at most a multiplicative overhead \(\gamma(|G_N|)\) (and constant-factor representation overhead) relative to executing the corresponding update/event operations. Therefore any asymptotic improvement in time complexity achieved by \(G_N\) transfers to \(G^*\) up to this overhead; if \(\gamma(|G_N|)\) does not asymptotically dominate, the improvement remains asymptotic.
\end{proof}

\begin{remark} (Constant factors and why neuromorphic can still matter for time).
Corollary~\ref{cor:timeAdvTransfers} emphasizes that asymptotic time improvements are algorithmic rather than uniquely neuromorphic, since neuromorphic algorithms are classically simulable. However, practical time-to-solution depends critically on constant factors and on whether an algorithm aligns with the dominant costs of the execution substrate (e.g., control/scheduling overhead and data movement on conventional platforms). As in fast matrix multiplication (e.g., Strassen-type methods), asymptotically favorable algorithms can be practically unattractive on conventional systems due to large constants or unfavorable memory-access patterns. Neuromorphic architectures may expose different constant-factor regimes because computation is spatially instantiated and event-addressed: highly parallel local updates, wavefront/frontier computations, and recurrent reuse can reduce conventional scheduling and data-movement overheads even when asymptotic time classes are unchanged.
\end{remark}

\subsubsection{Energy analysis of neuromorphic}
\label{sec:energy}
In this analysis we seek to break down the energetic costs of neuron computation and synapse communication into analyzable components. With the breakdown of $G_N$ in the previous sections, we next consider what it means for information to be processed on the neural graph. Energy expenditure in electronic systems arises primarily from two sources: computation and communication. Both of these can be related to a change of state: computation is the change of state within algorithm nodes, and communication is the change of state in algorithm edges.

To begin, we first define a baseline conventional energy model based on scheduled work and memory access. 

\begin{definition}[Energy Complexity (Scheduled Work and Access Model)]
Let $G=(\mathcal{V},\mathcal{E})$ be a conventional computational graph executed on a conventional platform over a horizon $T$ under some schedule. Let $A_{\mathcal{V}}(t)\subseteq \mathcal{V}$ denote the set of operations executed at step $t$ and let $A_{\text{acc}}(t)$ denote the set of distinct memory/register accesses performed at step $t$. The conventional energy is modeled as
\[
C_E^{\text{vN}}(G;T)= e_{\text{op}}\sum_{t=0}^{T-1} |A_{\mathcal{V}}(t)| \;+\; e_{\text{acc}}\sum_{t=0}^{T-1} |A_{\text{acc}}(t)|,
\]
where $e_{\text{op}}$ and $e_{\text{acc}}$ are platform-dependent constants capturing the average energy per operation and per access.
\label{def:energy_vn}
\end{definition}

As is readily apparent, while this memory-based energy model may be appropriate for conventional hardware, it is not suitable for neuromorphic hardware. At a high level, conventional platforms are typically \emph{address-addressed} (work is triggered by computing and dereferencing addresses, incurring access/control costs even for sparse workloads), whereas neuromorphic platforms are naturally \emph{activity-based} (events route computation directly to the relevant local state), motivating the trace-based energy model below. For this reason, we need to identify a novel energy model to describe event-driven neuromorphic hardware. 

In this section we adopt an event-count energy model. Specifically, we assume that each spike transmission event on an inter-neuron communication edge has a fixed energy cost. In practice, specific hardware implementations may induce non-uniform spike costs due to communication hierarchies (e.g., within-core, cross-core, cross-chip, cross-board routing). Such architectural details are beyond the scope of this initial analysis.

\begin{figure}
    \centering
    \includegraphics[width=.95\linewidth]{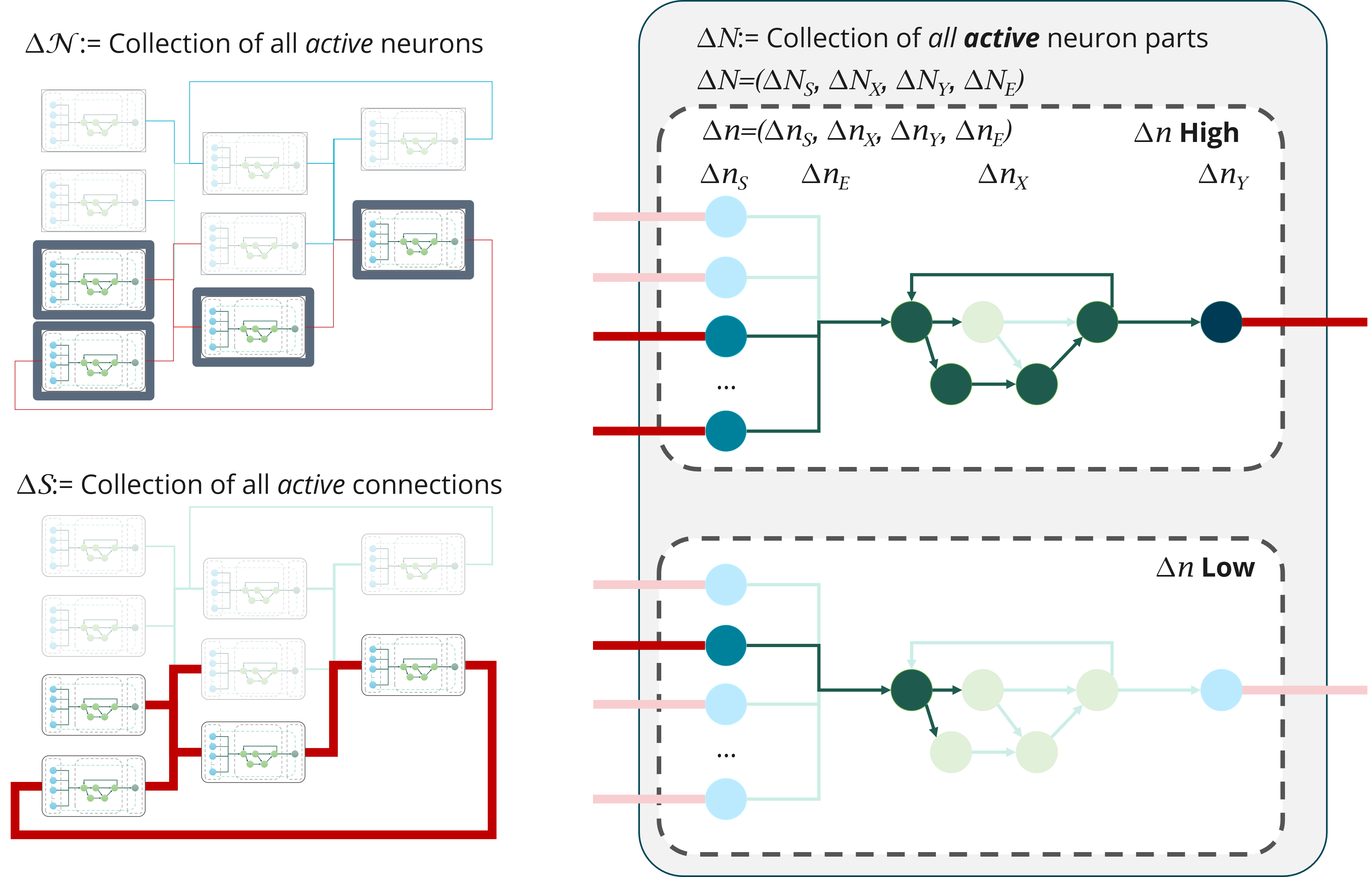}
    \caption{Illustration of trace model, $\Delta G_N=(\Delta N, \Delta S)$ of neural graphs used in framework. For a neural algorithm $G_N=(N,S)$, $\mathcal{\Delta N}$ (top left) represents the set of all neurons whose state has changed at a point in time, $\Delta S$ (bottom left) represents the set of all synaptic connections whose state has changed, and $\Delta N$ (right) represents the set of the neuron parts which consist of \textit{all} of the computation elements that have changed in the neural algorithm, including $\Delta N_S$ (all synaptic computation that has changed), $\Delta N_X$ (all internal neuronal dynamics that has changed), and $\Delta N_Y$ (all spiking dynamics that has changed), as well as $\Delta N_E$ which describes internal neuronal communication that has changed.}  
    \label{fig:trace_G_N}
\end{figure}

Accordingly, it is useful to introduce a new concept, which is illustrated in Figure \ref{fig:trace_G_N}:

\begin{definition}[Graph Activity and Change in Graph State]
Let $G=(\mathcal{V},\mathcal{E})$ represent a computational graph whose nodes and edges may have time-dependent computational states. Let time be discrete, indexed by $t \in \{0,1,\dots,T-1\}$. For each time step $t$, define:
\begin{itemize}
    \item $A_{\mathcal{V}}(t) \subseteq \mathcal{V}$ as the set of nodes that are \textbf{active} (i.e., whose state is updated) at time $t$;
    \item $A_{\mathcal{E}}(t) \subseteq \mathcal{E}$ as the set of edges that are \textbf{active} (i.e., on which a communication event occurs) at time $t$.
\end{itemize}
We define the \textbf{change in graph state} (or \textbf{graph activity}) at time $t$ as:
\[
\Delta G(t) = \big(A_{\mathcal{V}}(t),\,A_{\mathcal{E}}(t)\big),
\]
and the cumulative activity over a horizon $T$ as:
\[
\Delta G[0{:}T) = \left(\biguplus_{t=0}^{T-1} A_{\mathcal{V}}(t),\ \biguplus_{t=0}^{T-1} A_{\mathcal{E}}(t)\right),
\]
where we will primarily use the total counts $\sum_t |A_{\mathcal{V}}(t)|$ and $\sum_t |A_{\mathcal{E}}(t)|$ in subsequent bounds.
\label{def:deltaG}
\end{definition}

Based on our above formulation of $G_N=(N,S)$, we similarly define the activity of a neuromorphic algorithm at time $t$ as
\[
\Delta G_N(t) = (\Delta N(t), \Delta S(t)),
\]
where $\Delta S(t)\subseteq S$ is the set of synaptic communication channels that transmitted a spike at time $t$, and $\Delta N(t)\subseteq N$ is the set of neuron-owned components that are active (i.e., updated) at time $t$. Additionally, the change in $N$ can be further decomposed as
\[
\Delta N(t) = \big(\Delta N_S(t), \Delta N_X(t), \Delta N_Y(t), \Delta N_E(t)\big),
\]
where each component denotes the subset of those neuron-owned elements that are active at time $t$.

\begin{definition}[Energy Complexity (Event-count Model)]
Let $G=(\mathcal{V},\mathcal{E})$ represent a computational graph executed for $T$ discrete time steps. The \textbf{energy complexity} $C_E(G;T)$ is defined as the total energy required to execute the algorithm represented by $G$ under an event-count model:
\[
C_E(G;T) = e_{\mathcal{V}} \sum_{t=0}^{T-1} |A_{\mathcal{V}}(t)| \;+\; e_{\mathcal{E}} \sum_{t=0}^{T-1} |A_{\mathcal{E}}(t)|,
\]
where $e_{\mathcal{V}}$ is the energy cost per active node update and $e_{\mathcal{E}}$ is the energy cost per active edge communication event.
\label{def:energyComplexity}
\end{definition}

From Definition \ref{def:deltaG}, we can note lower- and upper-bounds on the magnitude of $C_E(G_N;T)$. We would like to get this in terms of $|S|$ and $M$ (and, more precisely, in terms of the activity $|\Delta S(t)|$). Our strategy here is to leverage the fact that neural algorithms consist of many isolated graphs that are connected through $S$.

\begin{lemma}[Degree-weighted spike count for synaptic activity]
\label{lem:degree_weighted_spike_count}
Assume \emph{broadcast spike semantics}: for each output node \(n_{y_i}\in N_Y\) and every outgoing synaptic communication channel \(s_{ij}\in S\) sourced at \(n_{y_i}\), the transmitted event satisfies
\[
s_{ij}[t] = \tau(y_i[t]),
\]
where \(y_i[t]\in\{0,1\}\) and \(\tau:\{0,1\}\to\{0,1\}\) (typically \(\tau\) is the identity).

Define the set of spiking output nodes with nonzero out-degree at time \(t\) as
\[
Y_1(t) := \{\, n_y\in N_Y : \tau(y[t]) = 1 \ \wedge\ \deg^+_{S}(n_{y})\ge 1 \,\}.
\]
Then the number of active synaptic communication events at time \(t\) satisfies
\[
|\Delta S(t)| \;=\; \sum_{n_{y}\in Y_1(t)} \deg^+_{S}(n_{y}),
\]
where \(\deg^+_{S}(n_y)\) denotes the out-degree of \(n_y\) with respect to the inter-neuron channel set \(S\).
\end{lemma}

\begin{proof}
Let \(S^+(n_y) := \{\, s\in S : s \text{ is sourced at } n_y \,\}\) denote the set of outgoing channels from an output node \(n_y\in N_Y\), so \(|S^+(n_y)|=\deg^+_{S}(n_y)\). By broadcast spike semantics, for any \(n_y\) and any \(s\in S^+(n_y)\), the edge event at time \(t\) satisfies \(s[t]=\tau(y[t])\). Therefore:
\begin{itemize}
    \item If \(n_y\in Y_1(t)\), then \(\tau(y[t])=1\) and all channels in \(S^+(n_y)\) are active at time \(t\), so \(S^+(n_y)\subseteq \Delta S(t)\).
    \item If \(n_y\notin Y_1(t)\), then either \(\tau(y[t])=0\) or \(\deg^+_{S}(n_y)=0\), in which case no channel in \(S^+(n_y)\) is active at time \(t\), so \(S^+(n_y)\cap \Delta S(t)=\emptyset\).
\end{itemize}
Since the sets \(S^+(n_y)\) for distinct output nodes \(n_y\) are disjoint (each edge has a unique source output node), it follows that \(\Delta S(t)\) is the disjoint union of outgoing-channel sets over \(Y_1(t)\):
\[
\Delta S(t) \;=\; \biguplus_{n_y\in Y_1(t)} S^+(n_y).
\]
Taking cardinalities yields
\[
|\Delta S(t)| \;=\; \sum_{n_y\in Y_1(t)} |S^+(n_y)|
\;=\; \sum_{n_y\in Y_1(t)} \deg^+_{S}(n_y),
\]
as claimed.
\end{proof}

\begin{lemma}[Per-step correspondence of spike events and post-synaptic synaptic-input activity]
For a neuromorphic graph $G_N=(N,S)$, at any time step $t$,
\[
|\Delta N_S(t)| = |\Delta S(t)|.
\]
\label{lemma:DeltaNS_equals_DeltaS}
\end{lemma}

\begin{proof}
Each synaptic communication channel $s_{ij}\in S$ is a directed edge from a unique source output node $n_{y_i}\in N_Y$ to a unique post-synaptic synaptic input node $n_{s_{ij}}\in N_S$. Therefore each active edge event in $\Delta S(t)$ activates exactly one post-synaptic synaptic input node in $\Delta N_S(t)$, and no two distinct edges share the same target synaptic input node. Hence the sets have the same cardinality.
\end{proof}

Similarly, we can define an upper bound as follows:

\begin{lemma}[Per-step upper bound on neuron-owned activity]
Let $G_N=(N,S)$ be a neuromorphic graph with isolated neuron subgraphs $\mathcal{N}$. Define
\[
c_x = \max_{n\in\mathcal{N}} |n_X|,\qquad
c_y = \max_{n\in\mathcal{N}} |n_Y|,\qquad
c_e = \max_{n\in\mathcal{N}} |n_E|.
\]
Then, for any time step $t$,
\[
|\Delta N(t)| \;\le\; 2|\Delta S(t)| \;+\; c_n M,
\]
where $c_n$ is a constant that depends only on neuron-internal structure (e.g., $c_n = c_x + c_y + c_e$).
\label{lemma:upper_DeltaN}
\end{lemma}

\begin{proof}
We bound each component of $\Delta N(t)=(\Delta N_S(t),\Delta N_X(t),\Delta N_Y(t),\Delta N_E(t))$.

From Lemma \ref{lemma:DeltaNS_equals_DeltaS}, $|\Delta N_S(t)| = |\Delta S(t)|$.

For each neuron $n$, at most $|n_X|\leq c_X$ internal-state nodes and $|n_Y|\leq c_y$ output nodes can be active at a step, hence $|\Delta N_X(t)|\leq c_X M$ and $|\Delta N_Y(t)|\leq c_Y M$.

Similarly, each neuron has at most $c_y$ output nodes; therefore $|\Delta N_Y(t)| \le |N_Y| \le c_yM$.

Finally, $\Delta N_E(t)$ includes internal edge activations associated with processing synaptic inputs and internal state updates. Because each neuron's internal edge set has size at most $c_e$, we have $|\Delta N_E(t)| \le |\Delta N_S(t)| + c_eM = |\Delta S(t)| + c_eM$.

Summing these bounds yields
\[
|\Delta N(t)| \le |\Delta N_S(t)| + |\Delta N_X(t)| + |\Delta N_Y(t)| + |\Delta N_E(t)|
\le 2|\Delta S(t)| + (c_x+c_y+c_e)M.
\]
\end{proof}

\begin{remarkn}(Bounded neuron complexity)
Although Definition~\ref{def:neuron} permits arbitrarily complex internal neuron subgraphs, meaningful scaling comparisons require fixing a neuron primitive class. Accordingly, we treat the maximum neuron subgraph size \(\kappa=\max_{n\in\mathcal{N}}(|n_S|+|n_X|+|n_Y|+|n_E|)\) as an architecture-dependent constant (analogous to a gate basis in circuit complexity). Increasing \(\kappa\) corresponds to providing more powerful neuron-local primitives, which can reduce required inter-neuron communication \(|S|\) and synaptic activity \(\sum_t|\Delta S(t)|\) at the expense of larger per-neuron constants.
\label{rmk:boundedNeuron}
\end{remarkn}

From Lemma \ref{lemma:DeltaNS_equals_DeltaS} and Lemma \ref{lemma:upper_DeltaN}, we can conclude that for all $t$,
\begin{equation}
|\Delta S(t)| \le |\Delta N(t)| \le 2|\Delta S(t)| + c_n M.
\label{eqn:DeltaN_bounds}
\end{equation}

Of note, the analysis here considers neurons with relatively simple internal computation, such as a LIF neuron. However, Definition \ref{def:neuron} permits arbitrarily complex graphs of internal neural dynamics. The relative costs and benefits of increasing neuronal complexity are an active area of research in the community today; in this framework, increased neuronal complexity is captured by larger constants $c_x$, $c_y$, and $c_e$.

Combining Definition \ref{def:energyComplexity} with the per-step bounds above implies that neuromorphic energy over a horizon $T$ scales with cumulative activity:
\begin{equation}
C_E(G_N;T) = O\!\left(\sum_{t=0}^{T-1}|\Delta S(t)| \;+\; M\,T\right),
\label{eqn:Energy_BigO}
\end{equation}
where the first term captures event-driven synaptic activity and the second term captures the (worst-case) possibility of neuron-internal updates across all neurons at every step. 

Because event-driven neurons do not update unless needed, we can further restrict equation \ref{eqn:Energy_BigO} in asynchronous systems to improve this upper bound to

\begin{equation}
C_E(G_N;T) = O\!\left(\sum_{t=0}^{T-1}|\Delta S(t)| \;+\; \sum_{t=0}^{T-1}|\Delta \mathcal{N}(t)|\right),
\label{eqn:Energy_BigO_aynchronous}
\end{equation}

where

\[
\sum_{t=0}^{T-1}|\Delta \mathcal{N}(t)|\leq M\,T.
\]

\begin{lemma}[Worst-case neuromorphic energy upper bound]
\label{lem:nmc_energy_worstcase}
Let \(G_N=(N,S)\) be a neuromorphic algorithm executed for \(T\) discrete time steps under the event-count energy model (Definition~\ref{def:energyComplexity}). Assume that each synaptic communication channel can generate at most a constant number \(b_s\) of spike events per time step (for standard binary spike channels, \(b_s=1\)). Then for all \(t\),
\[
|\Delta S(t)| \le b_s |S|,
\]
and therefore
\[
\sum_{t=0}^{T-1} |\Delta S(t)| \le b_s |S|T.
\]
Consequently, neuromorphic energy is bounded by
\[
C_E(G_N;T)=O(|S|T + MT),
\]
with constants depending on \(b_s\) and neuron-internal structure (via Lemma~\ref{lemma:upper_DeltaN}).
\end{lemma}

\begin{proof}
By assumption, at most \(b_s\) spike events can occur on each of the \(|S|\) channels per time step, so \(|\Delta S(t)|\le b_s|S|\). Summing over \(t\in\{0,\dots,T-1\}\) yields \(\sum_t |\Delta S(t)|\le b_s|S|T\).

From Lemma~\ref{lemma:upper_DeltaN}, for each \(t\),
\(
|\Delta N(t)| \le 2|\Delta S(t)| + c_nM
\),
hence
\[
\sum_t |\Delta N(t)| \le 2\sum_t |\Delta S(t)| + c_nMT = O(|S|T + MT).
\]
Finally, applying Definition~\ref{def:energyComplexity} gives
\(
C_E(G_N;T)=O\!\left(\sum_t |\Delta S(t)| + \sum_t |\Delta N(t)|\right)=O(|S|T+MT).
\)
\end{proof}

Based on these results, we can consider the alternative of the penalty that an event-driven neuromorphic algorithm receives by use on conventional hardware. 

\begin{lemma}[Conventional simulation of event-driven activity incurs control/access overhead]
Let $G_N=(N,S)$ be a neuromorphic algorithm executed for $T$ steps with activity sets $\Delta S(t)$ and $\Delta N(t)$. Any conventional simulation $G^*$ must (at minimum) represent and update the state of the active neurons and deliver the active spike events to their targets. Consequently,
\[
C_E^{\text{vN}}(G^*;T) \;\ge\; \Omega\!\left(\sum_{t=0}^{T-1} |\Delta S(t)|\right)
\]
and typically includes additional overhead terms associated with managing sparse event sets (e.g., indexing, queues, indirect addressing), which are absent in an ideal in-memory/event-driven implementation.
\label{lem:vn_sim_overhead}
\end{lemma}
\begin{proof}
In any conventional simulation \(G^*\), each spike-event on an active channel \(s_{ij}\in\Delta S(t)\) must be represented and processed at least once (to deliver the event to the appropriate post-synaptic state update), implying at least constant energy per event; therefore \(C_E^{\mathrm{vN}}(G^*;T)\ge \Omega\!\left(\sum_{t=0}^{T-1}|\Delta S(t)|\right)\). The additional control/access overhead terms arise from maintaining and traversing sparse event representations (e.g., queues or indirect-addressing structures).
\end{proof}
Building on these results, we can then identify the condition on which event-driven neuromorphic algorithms have an explicit energy advantage over a conventional approach.

\begin{theorem}[Energy scaling separation under sparse activity]
\label{thm:energy_separation_template}
Consider a family of neuromorphic executions of algorithms \(G_N=(N,S)\) over a horizon \(T\) for which:
\begin{align}
\sum_{t=0}^{T-1} |\Delta S(t)| &= o(|S|T), \label{eq:sparse_activity_assump}\\
\sum_{t=0}^{T-1} |\Delta N(t)| &= O\!\left(\sum_{t=0}^{T-1} |\Delta S(t)|\right). \label{eq:neuron_activity_not_dominate}
\end{align}
Under the event-count neuromorphic energy model,
\[
C_E(G_N;T)=O\!\left(\sum_{t=0}^{T-1} |\Delta S(t)|\right),
\]
i.e., energy is activity-proportional.

In contrast, consider any conventional implementation of the same logical interaction structure that performs a dense scheduled evaluation in the sense that there exists a constant \(\alpha>0\) such that, at each step \(t\),
\[
|A_{\mathcal{V}}(t)| + |A_{\text{acc}}(t)| \ge \alpha |S|,
\]
where \(A_{\mathcal{V}}(t)\) and \(A_{\text{acc}}(t)\) are as in Definition \ref{def:energy_vn}. Then the conventional energy satisfies
\[
C_E^{\mathrm{vN}}(G;T)=\Omega(|S|T)
\]
(up to platform-dependent constants). Therefore, for such families, neuromorphic execution can exhibit asymptotically smaller energy scaling than dense scheduled conventional execution.
\end{theorem}

\begin{proof}
By Definition \ref{def:energyComplexity}, the neuromorphic event-count energy is
\[
C_E(G_N;T) = e_{\mathcal{V}}\sum_{t=0}^{T-1}|\Delta N(t)| + e_{\mathcal{E}}\sum_{t=0}^{T-1}|\Delta S(t)|.
\]
Applying the assumption \eqref{eq:neuron_activity_not_dominate} and absorbing positive constants into Big-O gives
\[
C_E(G_N;T) = O\!\left(\sum_{t=0}^{T-1}|\Delta S(t)|\right).
\]

For the conventional implementation, Definition \ref{def:energy_vn} yields
\[
C_E^{\mathrm{vN}}(G;T)= e_{\mathrm{op}}\sum_{t=0}^{T-1} |A_{\mathcal{V}}(t)| \;+\; e_{\mathrm{acc}}\sum_{t=0}^{T-1} |A_{\mathrm{acc}}(t)|.
\]
Under the dense scheduled evaluation condition, for some \(\alpha>0\),
\[
|A_{\mathcal{V}}(t)| + |A_{\mathrm{acc}}(t)| \ge \alpha |S|\quad \text{for all } t,
\]
and since \(e_{\mathrm{op}},e_{\mathrm{acc}}>0\) are constants, it follows that
\[
C_E^{\mathrm{vN}}(G;T)=\Omega\!\left(\sum_{t=0}^{T-1} |S|\right)=\Omega(|S|T).
\]

Finally, combining \eqref{eq:sparse_activity_assump} with the derived bounds implies
\[
C_E(G_N;T)=o(|S|T)\quad\text{while}\quad C_E^{\mathrm{vN}}(G;T)=\Omega(|S|T),
\]
which establishes the claimed asymptotic separation for this family.
\end{proof}

Theorem~\ref{thm:energy_separation_template} is a \emph{conditional separation} result. It does not claim that conventional implementations must always consume \(\Omega(|S|T)\) energy, since a conventional platform may also employ event-driven data structures and sparse schedules. Rather, the theorem states that when a conventional implementation performs a dense scheduled evaluation that touches a constant fraction of the \(|S|\) interaction terms each step, its energy scales as \(\Omega(|S|T)\), whereas neuromorphic energy scales with realized activity \(\sum_t |\Delta S(t)|\), which can be \(o(|S|T)\) for activity-sparse algorithm families. Independently, Lemma~\ref{lem:nmc_energy_worstcase} shows that neuromorphic energy admits a worst-case upper bound proportional to \(|S|T\) (up to additional neuron-update terms), so sparse-activity regimes represent a strict improvement over this worst-case bound.

\newpage

\subsection{Conditions for a Neuromorphic Advantage}
\label{sec:NMCadvantage}

A useful way to interpret the above framework is through the lens of parallel computation. Neuromorphic architectures are intrinsically parallel and heterogeneous: in an idealized model, each neuron operates as an independent compute element and, because computation is instantiated directly in the circuit, there is no requirement that different parts of the algorithm execute the same instruction at the same time. In this sense, neuromorphic execution is naturally aligned with a multiple-instruction, multiple-data (MIMD) paradigm, in which different parts of the computational graph may implement different local update rules and operate asynchronously.

The previous sections outline how the time, space, and energy scaling of neuromorphic computing (NMC) differs from conventional architectures. In particular, ideal NMC can approach the infinite-processor limit of parallel computation, while its energy costs are naturally described by event-driven activity. However, because both the algorithm representation and the target architecture differ between conventional and neuromorphic systems, comparative claims must be stated in terms of algorithm families and in terms of properties of the induced execution traces.

Building on the framework's foundation, we can begin to explore what aspects of algorithms, with a particular goal of identifying which algorithm classes are most advantageous for neuromorphic. To do this, we will introduce formal algorithm properties (recurrence, locality, and activity decay) that are sufficient to explain when a neuromorphic advantage may or may not be expected. While conventional algorithm energy analysis is typically expressed in terms of scheduled work and memory / communication complexity (i.e., \textit{graph}-based); as section \ref{sec:energy} shows, the energy complexity of neuromorphic is heavily dependent on realized, and not scheduled, activity (i.e., \textit{trace }based). 

This suggests that evaluating whether a class of algorithms is suitable for neuromorphic, it is important to consider both the structure of the graph ($G$) and the expected execution of the graph ($\Delta G$. In this section, we will first define a graph-based landscape that can be evaluated directly from $G_N$ (or, if appropriate, a time-unrolled representation $G_N^{(T)}$) without executing the algorithm, and then define a trace-based landscape evaluated from $\Delta G_N$ which is based on how activity evolves during execution. 

To preview, our graph-based landscape (Section \ref{sec:Graph_landscape}) will consider three characteristics: 
\begin{enumerate}
    \item Within-time step structural homogeneity, $H(G)$, which is a measure of how many operations share a common instruction template within a time slice of a time-unrolled graph.
    \item Across-time step structural reuse, $R_T(G)$, which is the extent to which a finite instantiated graph can represent a larger unrolled computation via recurrence.
    \item Locality of communication, $K(G)$, which describes the extent to which communication is bounded and local.
\end{enumerate}

Additionally, our trace-based landscape (Section \ref{sec:Trace_landscape}) will consider three characteristics:
\begin{enumerate}
    \item Activity intensity, $\Phi_S$, which describes the normalized synaptic activity over the algorithm's execution.
    \item Activity trajectory, $a_s(t)$, which describes whether this activity is stationary, increases, or decreases over the course of execution
    \item Activity variability, $CV_S$, which describes how activity of the network varies across either different inputs or internal noise sources.
\end{enumerate}

For intuition, the graph-based landscape can provide an a priori assessment of how an algorithm will map to different architecture types based on the structure of the algorithm graph. The features explored here consider different structural characteristics that are often prioritized or de-prioritized by alternative methods of hardware parallelization, such as single-instruction multiple-data (SIMD) architecture like GPUs. 

In contrast, the trace-based landscape is input dependent and thus cannot be assessed precisely a priori, however it can often be estimated. This landscape assesses the extent to which the fact that neuromorphic energy is activity-dependent (as compared to work-dependent) can be expected to provide energy savings.

\subsubsection{Implication of algorithm structure}
\label{sec:Graph_landscape}
Our structural analysis of neuromorphic algorithms aims to quantify the extent to which an algorithm is best suited for a SIMD, NMC, or perhaps some alternative form of hardware acceleration. At a high level, SIMD architectures, such as GPUs, achieve strong parallelization within an algorithm time step by sharing a restricted set of pre-defined operations across a block of data stored in a shared memory (thus `single instruction multiple data'), whereas neuromorphic hardware achieves strong parallelization by dedicating hardware resources to each operation in the algorithm graph without the need of a shared memory. In this sense, NMC does not benefit to the same extent from homogeneity, but it does preferentially benefit from recursion which allows those operations to be reused over time. 

\subsubsection{Within-time step homogeneity and compressibility}

\label{sec:Graph_homogeneity}

Conventional accelerators such as GPUs primarily obtain efficiency by exploiting \emph{within-step homogeneity}: many operations that share the same instruction template can be executed in lockstep. To express this contrast in a way that remains architecture-agnostic, we introduce a simple notion of within-step homogeneity for a time-unrolled computational graph.

\begin{definition}[Within-step homogeneity (operation-type multiplicity)]
\label{def:simd_homogeneity}
Let \(G^{(T)}=(V^{(T)},E^{(T)})\) be a time-unrolled computational graph whose vertices are partitioned by step,
\[
V^{(T)}=\biguplus_{t=0}^{T-1} V_t.
\]
Assume each vertex \(v\in V^{(T)}\) is labeled by an \emph{operation type} \(\mathrm{type}(v)\) (e.g., add, multiply, threshold, compare). Define the within-step homogeneity as
\[
H(t):=\max_{op}\, |\{v\in V_t:\ \mathrm{type}(v)={op}\}|,
\]
and $H_{\mathrm{avg}}:=\frac{1}{T}\sum_t H(t)$.

Further, let \(G^{(T)}=(V^{(T)},E^{(T)})\) be time-unrolled with \(V^{(T)}=\biguplus_{t=0}^{T-1}V_t\), and let \(A_V(t)\subseteq V_t\) denote the set of vertices that are active at step \(t\) under the chosen execution semantics. Define the \emph{active within-step homogeneity} as
\[
H^\Delta(t) \;:=\; \max_{op}\, \big|\{\,v\in A_V(t):\ \mathrm{type}(v)=op\,\}\big|.
\]

\end{definition}

To help interpret this notion of homogeneity, large \(H(t)\) indicates that step \(t\) contains many operations that share a common set of operations and are therefore amenable to lockstep/SIMD-style execution. Small \(H(t)\) indicates that step \(t\) is heterogeneous in its operations, which can reduce SIMD utilization even when abundant total work exists.

\begin{proposition}[Lockstep lower bound from within-step homogeneity]
\label{prop:lockstep_homogeneity_lb}
Assume a lockstep machine that, at step \(t\), can apply a \emph{single} operation type in parallel to at most \(L_W\) active operations, and cannot mix operation types within the same lockstep group. Then step \(t\) requires at least
\[
\left\lceil \frac{|A_{\mathcal{V}}(t)|}{\min\{L_W,H^\Delta(t)\}} \right\rceil
\]
lockstep groups, where \(H^\Delta(t)\) is the maximum multiplicity of a single operation type among the active operations at step \(t\).
\end{proposition}

\begin{proof}
At step \(t\), any lockstep group can include at most \(L_W\) operations by hardware limitation, and it can include operations of only one type. The largest set of same-type active operations has size \(H^\Delta(t)\). Hence the maximum number of active operations that can be covered by a single lockstep group is at most \(\min\{L_W,H^\Delta(t)\}\). Covering all \(|A_{\mathcal{V}}(t)|\) active operations therefore requires at least
\(\left\lceil |A_{\mathcal{V}}(t)|/\min\{L_W,H^\Delta(t)\}\right\rceil\) groups.
\end{proof}

As can be seen, within-step homogeneity \(H(t)\) is a primary efficiency lever for SIMD accelerators because it allows many identical operations to be executed in lockstep. In the neuromorphic model considered here, computation is \emph{spatially instantiated}: each neuron-owned update and each synaptic event corresponds to a distinct physical element and a distinct state transition. Consequently, within-step homogeneity does not provide an analogous ``packing'' benefit on neuromorphic hardware---if \(M\) neuron updates (or synaptic events) occur at a given step, they must still be realized as \(M\) physical update/event operations regardless of whether their local update rules are identical. Any benefit from homogeneity must therefore arise indirectly, by enabling a more compact instantiated graph (smaller \(|G_N|\)) or a sparser activity trace (smaller \(\sum_t|\Delta S(t)|\)), rather than by lockstep execution. We can formalize this as well.

\subsubsection{Across time homogeneity and recurrence}
\label{sec:graph_structure_Recurrence}

While the previous section was concerned with homogeneity within a time step; algorithms can also have sequential computations that are repeated. While such sequential computations typically cannot be parallelized within a single time step; they do lend themselves to a representation as recurrence, or cycles in the algorithm graph, whereby the same compute resources may be used repeatedly. This is of particular importance in neural circuits where each operation must be physically realized in hardware, so recurrence directly leads to a space savings. For this reason, while conventional algorithms ($G=(\mathcal{V}, \mathcal{E})$ are typically described as directed acyclic graphs (DAGs), neuromorphic algorithm graphs $G_N=(N,S)$ often contain cycles both within neurons and across neurons. To apply standard depth/work reasoning (e.g., Brent-style bounds), it is useful to define a time-unrolled representation over a finite horizon.

\begin{definition}[Time-unrolled computational graph]
    Let $G_N=(N,S)$ be a neuromorphic algorithm executed for $T$ discrete time steps. The \textbf{time-unrolled} graph $G_N^{(T)} = (\mathcal{V}^{(T)},\mathcal{E}^{(T)})$ is a DAG obtained by replicating the neuron-owned state update structure across time, such that each neuron-owned state variable (and each inter-neuron connection) at time $t$ is represented by distinct vertices/edges in the unrolled graph, and dependencies follow the update semantics of $G_N$ from time $t$ to $t+1$.
    \label{def:unroll}
\end{definition}

The time-unrolled graph $G_N^{(T)}$ allows the definitions of total work $T_1(G_N^{(T)})$ and depth $T_{\infty}(G_N^{(T)})$ to be applied to recurrent neuromorphic algorithms. Time-unrolling refers to the structural update dependencies; actual activity is captured separately by $\Delta S(t)$ and $\Delta N(t)$.

\begin{definition}[Structural reuse factor]
    Let $G_N=(N,S)$ be a neuromorphic algorithm executed for $T$ steps. Define the \textbf{structural reuse factor} as
    \[
        R_T(G_N) = \frac{T_1(G_N^{(T)})}{|G_N|},
    \]
    where $|G_N|$ denotes a consistent size measure of the instantiated neuromorphic algorithm (e.g., the number of neuron-owned nodes plus the number of inter-neuron communication channels).
    \label{def:reuse}
\end{definition}

Intuitively, $R_T(G_N)$ captures the extent to which a finite instantiated neuromorphic graph can represent and execute a larger unrolled computation via recurrence. Large $R_T$ corresponds to strong reuse of the same neuron resources across time steps (e.g., iterative algorithms on a fixed graph). The structural reuse factor \(R_T(G_N)\) compares the work of the time-unrolled computation to the size of the instantiated neuromorphic graph. In particular:
\begin{itemize}
    \item For a purely feed-forward computation evaluated once, \(T_1(G_N^{(T)})\) is typically proportional to \(|G_N|\) (up to constants), giving \(R_T(G_N)=O(1)\).
    \item For recurrent/iterative computations executed for many steps on a fixed instantiated graph, \(T_1(G_N^{(T)})\) may grow with \(T\) while \(|G_N|\) remains constant, yielding \(R_T(G_N)\) that grows with \(T\).
\end{itemize}
Thus large \(R_T\) captures algorithmic reuse enabled by recurrence without unrolling.

Unlike stored-program reuse (time-multiplexing general-purpose processors), large \(R_T\) here arises from executing a fixed recurrent circuit over time. This in effect allows in-memory architectures like neuromorphic to capture some of the space savings of stored-program architectures without sacrificing time costs.

There is one final consideration with regard to recurrence and depth for neuromorphic circuits, which is the ability to trade depth for space. As mentioned above in Section \ref{sec:time_space}, neuromorphic architectures allow computations to be restructured to reduce computational depth to minimize sequential communication stages by expanding compute elements per time step, a process known as spatialization. This while neuromorphic architectures by definition fully realize the computational graph (thus space equals serial depth, minus any recurrence reuse), they do allow the algorithm graph to vary along the classic depth--space axis. Here, we can formalize this tradeoff function as it is useful when comparing computationally equivalent neuromorphic algorithms of different depths.

\begin{definition}[Depth--footprint tradeoff function]
\label{def:depth_footprint_tradeoff}
Fix a problem family \(\mathcal{P}\) and an execution semantics. For any instance \(x\in\mathcal{P}\) and depth budget \(D\in\mathbb{N}\), define the minimum instantiated size required to achieve depth at most \(D\) as
\[
S^\star(x;D) \;:=\; \min \left\{\, |G_N| \;:\; G_N \text{ correctly solves } x \text{ and } T_{\infty}(G_N)\le D \,\right\}.
\]
For a family-level summary, define
\[
S^\star(\mathcal{P};D) \;:=\; \sup_{x\in\mathcal{P}} S^\star(x;D),
\]
i.e., the worst-case (over the family) minimal footprint needed to achieve depth at most \(D\).
\end{definition}

Definition~\ref{def:depth_footprint_tradeoff} makes explicit a second, complementary form of ``compressibility'' beyond recurrence. The reuse factor \(R_T(G_N)\) (Definition~\ref{def:reuse}) captures reuse of a fixed instantiated circuit across time, whereas \(S^\star(\mathcal{P};D)\) captures the footprint required to compress a computation into a small number of sequential stages. When physical latency (depth) is tightly constrained, one expects architectures to favor solutions with small \(T_{\infty}\) even when \(S^\star(\mathcal{P};D)\) is large, i.e., extreme spatial parallelism to reduce depth.

\subsubsection{Locality and bounded-degree communication structure}

While the homogeneity and recurrence structural metrics focused on the compressibility of the compute elements within and across time steps, the final structural consideration we focus on is the communication structure. As conventional stored program architectures leverage shared memories (which are typically accessible hierarchically), there is only a modest (e.g., logarithmic) benefit of accessing local information over global information. In contrast, as neuromorphic algorithms are physically realized across an architecture, the distance of communication is of immediate importance, as communication introduces both time (i.e., delays), space, and energy costs. For this reason, many problems of interest to neuromorphic computing have a communication skeleton that is sparse and local (e.g., meshes, graphs with bounded degree, nearest-neighbor couplings). We capture this with a bounded fan-out (degree) condition and, optionally, a placement-sensitive distance model.

\begin{definition}[Bounded fan-out neuromorphic graph family]
A family of neuromorphic algorithms $\{G_N\}$ has \textbf{bounded fan-out} $K$ if for every graph $G_N=(N,S)$ in the family, each neuron output node $n_{y_i}\in N_Y$ is the source of at most $K$ inter-neuron synaptic communication channels in $S$.
\label{def:boundedFanout}
\end{definition}

\begin{corollary}[Bounded fan-out implies synaptic activity is proportional to spike count]
\label{cor:bounded_fanout_spikecount}
Let \(G_N=(N,S)\) satisfy bounded fan-out \(K\) (Definition~\ref{def:boundedFanout}). Under the broadcast spike semantics of Lemma~\ref{lem:degree_weighted_spike_count}, for any time step \(t\),
\[
|\Delta S(t)| \;\le\; K\,|Y_1(t)|,
\]
where \(Y_1(t)\subseteq N_Y\) is the set of spiking output nodes (with nonzero out-degree) at time \(t\).
\end{corollary}

\begin{proof}
By Lemma~\ref{lem:degree_weighted_spike_count},
\[
|\Delta S(t)|=\sum_{n_y\in Y_1(t)}\deg^+_S(n_y).
\]
Under bounded fan-out, \(\deg^+_S(n_y)\le K\) for all \(n_y\), so
\[
|\Delta S(t)|\le \sum_{n_y\in Y_1(t)} K = K|Y_1(t)|.
\]
\end{proof}

While not necessary for idealized asymptotic algorithm analysis, in practice the physical distance of communication will be a limit of neuromorphic algorithm efficiency. As such, we also introduce a measure of locality-weighted synaptic activity to point towards realization costs. 

\begin{definition}[Locality-weighted synaptic activity]
Let $G_N=(N,S)$ be a neuromorphic algorithm and let $\pi$ denote a placement (embedding) of neurons onto a hardware substrate with a distance metric $d(\cdot,\cdot)$. Define the \textbf{locality-weighted} synaptic activity at time $t$ as
\[
a_S^{(d)}(t) = \sum_{s_{ij}\in \Delta S(t)} d\big(\pi(n_i),\pi(n_j)\big).
\]
\label{def:localWeightedActivity}
\end{definition}

We will note that in this paper we primarily adopt the fixed-cost event-count energy model of Section \ref{sec:energy}; Definition \ref{def:localWeightedActivity} is provided as a forward-looking refinement for architectures where interconnect distance materially affects spike energy.

This concept of fan-out is critical to our final metric, which quantifies the magnitude of communication in an algorithm by taking the maximum and average output degree of neurons.

\begin{definition}[Communication locality and degree]
Let \(G_N=(N,S)\) be a neuromorphic algorithm graph. Define the (maximum) fan-out
\[
K(G_N) := \max_{n_y\in N_Y} \deg^+_S(n_y),
\]
and the average fan-out \(\bar{K}(G_N):=|S|/|N_Y|\).
Optionally, given a class of allowable placements \(\Pi\) onto a canonical substrate with distance metric \(d(\cdot,\cdot)\), define the (normalized) globality index
\[
\mathrm{Glob}(G_N) := \inf_{\pi\in\Pi}\ \frac{1}{|S|}\sum_{s_{ij}\in S} d(\pi(n_i),\pi(n_j)).
\]
\end{definition}

Importantly, this measure of algorithm degree may be dependent on the size of the graph, so from a complexity perspective understanding how communication locality scales with input size is critical. Ideally a neuromorphic algorithm's communication skeleton will be scale-agnostic, that is linear, with the number of neurons. Fortunately, we see that this is the case if fan-out is indeed bounded.

\begin{lemma}[Bounded fan-out implies linear synapse scaling]
\label{lem:bounded_degree_linear_S}
Let \(G_N=(N,S)\) be a neuromorphic algorithm graph. Then
\[
|S| \le K_{\max}(G_N)\,|N_Y|.
\]
If a graph family satisfies \(K_{\max}(G_N)=O(1)\) and each neuron has \(O(1)\) output nodes (so \(|N_Y|=\Theta(M)\)), then \(|S|=O(M)\).
\end{lemma}

\begin{proof}
By definition, \(|S|=\sum_{n_y\in N_Y}\deg^+_S(n_y)\le K_{\max}(G_N)|N_Y|\). The second claim follows by substituting \(K_{\max}(G_N)=O(1)\) and \(|N_Y|=\Theta(M)\).
\end{proof}

This bounded-degree perspective also clarifies why dense numerical kernels are structurally misaligned with scalable neuromorphic implementations. When a problem’s natural dependency structure is dense (e.g., all-to-all interactions), the corresponding neuromorphic communication skeleton has $|S|$ that grows superlinearly with problem size, increasing both footprint and the worst-case energy bound of Lemma \ref{lem:nmc_energy_worstcase}. In such cases, any neuromorphic advantage depends on either restructuring the interaction graph (e.g., sparsification/approximation) or achieving trace sparsity (low $\Phi_S$) despite a dense structure.

\subsubsection{Implications of algorithm activity}
\label{sec:Trace_landscape}

Whereas the structural analysis above focuses on static properties of an algorithm graph, the activity analysis focuses on what actual computations occur when an algorithm is instantiated; that is the \textit{trace} of the algorithm graph over time. Activity is by definition input-dependent---if we knew the outputs of computation a priori, that output could be structurally realized---and as such we cannot directly assess the activity without measuring it empirically. Nevertheless, as described in Section \ref{sec:energy}, the energy consumption of event-driven neuromorphic is fundamentally determined by what compute and communication nodes are used, not the structural graph itself, so trace-based measures are required to assess an algorithms' neuromorphic potential.

The measures we will use are all directly related to the activity of the network itself. Specifically, we consider three measures, the average activation of neurons and synapses, how that activity changes over time, and how that activity changes across various inputs. Note that these latter two are roughly analogous to the across-time and within-time structural analysis described above. 

\subsubsection{Activity processes}

Section \ref{sec:energy} defines neuromorphic energy in terms of per-step activity sets $\Delta N(t)$ and $\Delta S(t)$. For convenience, we define the corresponding scalar activity processes:

\begin{definition}[Neuron and synapse activity processes]
Let $G_N=(N,S)$ be a neuromorphic algorithm and let $\Delta G_N(t)=(\Delta N(t),\Delta S(t))$ denote its per-step activity. Define
\[
a_N(t) = |\Delta N(t)|,\qquad a_S(t) = |\Delta S(t)|.
\]
\label{def:activity}
\end{definition}

\begin{definition}[Trace-based activity intensity (normalized activity)]
\label{def:activity_intensity}
Let \(G_N=(N,S)\) execute for \(T\) steps with synaptic activity \(\Delta S(t)\subseteq S\). Define the normalized synaptic activity intensity
\[
\Phi_S(G_N;T) := \frac{1}{|S|T}\sum_{t=0}^{T-1}|\Delta S(t)|.
\]
Optionally, define the normalized neuron-update intensity (in asynchronous semantics) using \(\Delta\mathcal{N}(t)\subseteq\mathcal{N}\):
\[
\Phi_{\mathcal{N}}(G_N;T) := \frac{1}{MT}\sum_{t=0}^{T-1}|\Delta\mathcal{N}(t)|.
\]
\end{definition}

\subsubsection{Activity convergence}
\label{sec:activity_subsection}
\begin{definition}[Activity-decaying algorithm]
A neuromorphic algorithm $G_N$ executed for $T$ steps is an \textbf{activity-decaying algorithm} if its synaptic activity process $a_S(t)$ is non-increasing over time (or non-increasing in expectation), i.e.,
\[
\exists t_0 : \mathbb{E}[a_S(t+1)] \le \mathbb{E}[a_S(t)] \quad \text{for all } t>t_0,
\]
and there exists a limiting regime in which
\[
\lim_{t\to\infty} \mathbb{E}[a_S(t)] \leq a_\infty.
\]

where $a_\infty$ is a small, not necessarily $0$, constant.
\label{def:activityDecay}
\end{definition}

This definition captures the intuition that iterative solvers may naturally reduce activity as they converge, and therefore may reduce per-step energy consumption over time.

\begin{definition}[Trace-based activity decay)]
\label{def:activity_trajectory}
Let \(a_S(t)=|\Delta S(t)|\) be the synaptic activity process (Definition \ref{def:activity}). We say the execution exhibits \textbf{activity decay} if \(a_S(t)\) is non-increasing over time (or non-increasing in expectation), as in Definition~\ref{def:activityDecay}. More generally, we treat the qualitative trajectory class of \(a_S(t)\) (decaying, stationary, or increasing) as a trace-based coordinate.
\end{definition}

Activity decay is relevant because it directly reduces cumulative activity. In particular, if $a_S(t)$ is decaying and $\sum_t(a_S(t)=o(|S|T)$, then the synaptic component of energy is strictly sub-worst-case relative to the bound $\Theta(|S|T)$ (Theorem \ref{thm:energy_separation_template}). Thus, for decaying-trace families of algorithms, energy is governed by the area under the activity curve rather than by the structure of the algorithm.

\subsubsection{Activity variability}
Our final analysis metric looks at the within-time variability of activity, or more specifically how predictable activity can be across inputs? This is the trace-based analogue to the structural homogeneity measure described above. For SIMD architectures in particular, having activity threads diverge (visit different parts of an algorithm graph) offsets many of their  parallelization benefits, however this activity-variability cost is not necessarily realized in idealized neuromorphic systems.

\begin{definition}[Trace-based variability]
\label{def:activity_variability}
Let \(A_S := \sum_{t=0}^{T-1}|\Delta S(t)|\) denote cumulative synaptic activity over a horizon \(T\). When execution depends on inputs or intrinsic stochasticity, \(A_S\) is a random variable over an input distribution (and/or internal randomness). Define the \textbf{coefficient of variation} of cumulative synaptic activity as
\[
\mathrm{CV}_S(G_N;T) := \frac{\sqrt{\mathrm{Var}(A_S)}}{\mathbb{E}[A_S]},
\]
with the convention that \(\mathrm{CV}_S=0\) when \(A_S\) is deterministic.
\end{definition}

With regard to activity variability and stochasticity, we can state that the local nature of stochasticity in neuromorphic architectures is critical; if a neuromorphic system relies on a central source of randomness many of the potential neuromorphic advantages may be lost.

\begin{proposition}[Centralized randomness induces additional communication or neuron complexity]
\label{prop:central_randomness_cost}
Consider a neuromorphic algorithm family that requires \(q(t)\) random draws at step \(t\) to determine neuron-local updates or routing decisions. If randomness is provided locally as in Definition~\ref{def:local_stochasticity}, then these draws incur no additional inter-neuron communication cost beyond the existing activity \(\Delta S(t)\) (i.e., they can be absorbed into neuron-local constants).
In contrast, if randomness is provided by a shared generator external to the neuron subgraphs, then supplying \(q(t)\) random values at step \(t\) requires either:
\begin{enumerate}
    \item augmenting the inter-neuron communication structure \(S\) to distribute random values, increasing \(|S|\) and generally increasing \(|\Delta S(t)|\), or
    \item increasing neuron-subgraph complexity (e.g., additional state/edges in \(n_X,n_E\)) to buffer/pipeline shared random values, violating a fixed primitive class unless accounted for in constants.
\end{enumerate}
\end{proposition}

\begin{proof}
By definition of the framework, any information shared across neurons that influences their updates must be communicated through inter-neuron channels or be stored locally. Local random sources satisfy this by construction. A shared generator, however, produces values that must be made available to multiple neuron subgraphs; distributing those values requires communication (new edges/events) or additional local buffering/control logic, both of which are counted in the framework either as increases in \(|S|\), \(|\Delta S(t)|\), or neuron-subgraph size parameters.
\end{proof}

\subsubsection{Implications for compilation and algorithm design}
The graph-based landscape indicates whether an algorithm \emph{can} be realized with favorable structure (bounded degree/locality, recurrence, etc.) and whether it is naturally homogeneous (SIMD-friendly). The trace-based landscape determines whether the execution \emph{actually} realizes low energy by maintaining low activity intensity \(\Phi_S\), exhibiting activity decay, and/or exhibiting high variability (which can frustrate lockstep scheduling but is naturally handled by event-driven execution). Together, these landscapes provide a vocabulary for characterizing when an algorithm family is likely to be GPU-friendly, neuromorphic-friendly, both, or neither.
The results above suggest that neuromorphic advantage is most plausible when compilation (mapping) can produce neuromorphic algorithms with:
\begin{itemize}
    \item bounded fan-out and/or bounded-degree communication skeletons (supporting locality and limiting $|S|$),
    \item recurrent structure enabling high structural reuse $R_T(G_N)$,
    \item activity-decay or sparse activity over time, reducing $\sum_t a_S(t)$ and therefore energy.
\end{itemize}
Conversely, computations whose natural representations require dense connectivity and sustained activity (e.g., dense feed-forward layers) do not naturally exploit the event-driven energy model, and therefore are less likely to exhibit scalable neuromorphic advantages without substantial graph restructuring.

\paragraph{SIMD versus neuromorphic ``compressibility''.}
Definition~\ref{def:simd_homogeneity} formalizes a notion of within-step shared structure: large \(H(t)\) (or \(H^{\Delta}(t)\)) indicates abundant isomorphic operations at a common time slice and therefore favors SIMD-style accelerators. In contrast, neuromorphic architectures do not require such within-step homogeneity for correctness or parallelism, but can benefit strongly from \emph{across-time} compressibility via recurrence, captured by large structural reuse \(R_T(G_N)\) (Definition \ref{def:reuse}). This suggests a complementary landscape: dense homogeneous graphs (e.g., dense linear algebra) are typically SIMD-friendly, while sparse, irregular, recurrent, and activity-sparse computations are naturally aligned with neuromorphic execution.

\begin{remark}(Structure vs. trace)
The quantities \(H(\cdot)\), \(R_T(\cdot)\), and \(K(\cdot)\) (or \(\mathrm{Glob}(\cdot)\)) are structural properties of an instantiated graph and can be assessed without executing the algorithm. In contrast, the energy model depends on the execution trace via \(\sum_t|\Delta S(t)|\), so activity sparsity/decay is an execution property that depends on inputs and dynamics. Nevertheless, for several important families (e.g., convergent iterative methods and frontier-based propagations), structural constraints together with standard dynamical assumptions can imply activity concentration or decay.
\end{remark}

\subsubsection{Two illustrative algorithm families}

We now formalize two algorithm families: (i) a positive example of iterative bounded-degree dynamics (mesh-like problems), and (ii) a negative example of dense feed-forward layers.

\begin{definition}[Iterative bounded-degree state-update family (mesh-like)]
An \textbf{iterative bounded-degree state-update problem} is characterized by:
\begin{itemize}
    \item $M_S$: the number of state locations (sites);
    \item $M_T$: the number of discrete time steps (iterations);
    \item $K$: a constant bound on the number of neighbors influencing any site per iteration;
    \item a per-site local update computation with total work $T_{1S}$ and depth $T_{\inf S}$ (per iteration).
\end{itemize}
The corresponding conventional unrolled computational graph $G_{\text{mesh}}$ has
\[
T_{1,\text{mesh}} = \Theta(M_S M_T T_{1S}),\qquad
T_{\inf,\text{mesh}} = \Theta(M_T T_{\inf S}),
\]
and its communication skeleton has bounded degree $K$.
\label{def:meshFamily}
\end{definition}

\begin{definition}[Dense feed-forward layer family]
A \textbf{dense feed-forward layer} is characterized by input size $N_i$ and output size $N_j$ with all-to-all connectivity between the layer input and output. The induced computational graph has
\[
T_{\inf,\text{ff}} = \Theta(1),\qquad T_{1,\text{ff}} = \Theta(N_i N_j),
\]
and the corresponding neuromorphic communication skeleton is dense with $|S|=\Theta(N_i N_j)$.
\label{def:denseFF}
\end{definition}

We now state two high-level propositions that summarize the implications of Sections \ref{sec:time_space} and \ref{sec:energy} for these two algorithm families.

\begin{proposition}[Iterative bounded-degree problems admit activity-driven energy scaling]
Consider an iterative bounded-degree state-update problem (Definition \ref{def:meshFamily}) and let $G_N$ denote a neuromorphic implementation that instantiates one neuron subgraph per state site and uses bounded fan-out $K$ for inter-site communication. Then:
\begin{itemize}
    \item (\textbf{Time}) In an ideal fully-parallel neuromorphic execution model, the runtime satisfies
    \[
    C_T(G_N;M_T) = \Theta(M_T T_{\inf S}),
    \]
    matching the depth of the time-unrolled computation.
    \item (\textbf{Footprint}) The instantiated resource footprint satisfies
    \[
    |N| = \Theta(M_S),\qquad |S| = \Theta(M_S),
    \]
    up to constant factors determined by the neuron subgraph template and $K$.
    \item (\textbf{Energy}) Under the event-count energy model (Definition \ref{def:energyComplexity}),
    \[
    C_E(G_N;M_T) = O\!\left(\sum_{t=0}^{M_T-1} a_S(t) + M_S M_T\right),
    \]
    and if $G_N$ is activity-decaying (Definition \ref{def:activityDecay}), then the cumulative synaptic activity term may be substantially smaller than the worst-case $\Theta(M_S M_T)$ scaling.
\end{itemize}
\label{prop:meshPositive}
\end{proposition}

\begin{proof}
The time statement follows by applying Definition \ref{def:timeComplexity} to the time-unrolled graph $G_N^{(M_T)}$ (Definition \ref{def:unroll}) and taking the ideal fully-parallel limit in which sufficient physical parallelism exists to match the unrolled dependencies, yielding runtime proportional to depth. The footprint statement follows from instantiating $O(1)$ neuron-owned resources per state site and at most $K$ outgoing communication channels per site. The energy statement follows directly from Section \ref{sec:energy}, noting that $a_S(t)=|\Delta S(t)|$ and that, per Lemma \ref{lemma:upper_DeltaN}, neuron-owned activity is bounded by a linear function of synaptic activity plus a constant times the number of neurons.
\end{proof}

\begin{proposition}[Dense feed-forward layers do not admit an activity-sparsity advantage under sustained activation]
Consider the dense feed-forward layer family (Definition \ref{def:denseFF}). Let $G_N$ be a neuromorphic implementation with $|S|=\Theta(N_i N_j)$. Suppose that for each step $t$ in a forward evaluation, a constant fraction of synapses are active (in expectation), i.e.,
\[
\mathbb{E}[a_S(t)] \ge \alpha |S| \quad \text{for some constant } \alpha>0.
\]
Then, under the event-count energy model,
\[
\mathbb{E}[C_E(G_N;T)] = \Omega(T\,|S|) = \Omega(T\,N_i N_j),
\]
and therefore no asymptotic energy scaling advantage should be expected from event-driven execution for this dense layer under sustained activation.
\label{prop:denseNegative}
\end{proposition}

\begin{proof}
Under the stated assumption, each step incurs expected synaptic activity at least $\alpha|S|$. The event-count energy model (Definition \ref{def:energyComplexity}) implies expected energy is lower bounded by a constant times $\sum_t \mathbb{E}[a_S(t)] \ge T\alpha|S|$, yielding the result.
\end{proof}

\newpage

\subsection{Implications of Digital and Analog Architectures}
\label{sec:DigitalAnalog_Extension}

The previous two sections aim to formulate an NMC framework that is agnostic to implementation to maximize generality. For this reason, the algorithm metrics defined in Section \ref{sec:NMCadvantage} are defined purely in terms of algorithm structure $G$ and execution trace $\Delta G_N(t)$ and are independent of device physics. In practice, the manner that execution trace maps to energy depends on the physical implementation of neuron and synapse state, such as digital versus analog. Ultimately, as this section demonstrates, the underlying primitives central to NMC implementation are central to understanding how algorithm dynamics, such as the time-derivative of an algorithm's state evolution, relates to execution trace and, accordingly, required energy.  

In this section we provide non-normative extensions of the framework that connects the trace model to digital switching with a microstate-derivative interpretation and analog intrinsic dynamics with a residual-to-flow interpretation.  

\subsubsection{Derivative-based energy complexity}
\label{sec:energyComplexityDigital}
The event-count model (Definition \ref{def:energyComplexity}) treats neuromorphic energy as proportional to realized activity in the execution trace.
For \emph{digital} neuromorphic systems, this trace-based view can be grounded more directly in the standard abstraction that dynamic energy is dominated by \textit{bit flip} state transitions rather than by static state. This motivates a refinement in which ``activity'' is defined not as ``a component was processed,'' but as ``a component's represented state has changed.'' Under these transition-based semantics, the activity trace can be interpreted as a discrete-time derivative of the algorithm graph's global microstate. Note that this analysis is not entirely general: if either neuron or synapse dynamics are analog, then the relationship between energy and state evolution must be modeled differently.

\begin{definition}[Digital microstate of a neuromorphic algorithm]\label{def:microstate}
Let $G_N=(N,S)$ be a neuromorphic algorithm executed for $T$ discrete steps on a \emph{digital} substrate.

A \emph{digital microstate} is a binary vector
\[
X(t)\in\{0,1\}^{B},
\]
that encodes all state held by the instantiated algorithm at time $t$, including neuron-owned state within $N$ 
and any communication or router microstate necessary to represent events within $S$.

We write $X(t)=(X_N(t),X_S(t))$ for a partition into neuron-owned and synapse-associated bits.

Define the discrete-time \emph{microstate derivative support} as the set of bits that toggle between $t$ and $t+1$:
\[
\Delta X(t) := \{ i\in\{1,\dots,B\} : X_i(t+1)\neq X_i(t)\}.
\]
\end{definition}

\begin{lemma}[Derivative switching interpretation of activity in digital NMC]
\label{lem:derivative-trace-equivalence}
Let $G_N=(N,S)$ be a neuromorphic algorithm with digital microstate $X(t)$ as in Definition~\ref{def:microstate}.
Define a \emph{transition-based activity trace} $\Delta G_N^{\oplus}(t)=(\Delta N^{\oplus}(t),\Delta S^{\oplus}(t))$ by letting
$\Delta N^{\oplus}(t)$ be the set of neuron-owned microstate bits that toggle between $t$ and $t+1$ and
$\Delta S^{\oplus}(t)$ be the set of edge-/synapse-associated microstate bits that toggle between $t$ and $t+1$.
Then, for every $t$,
\[
|\Delta X(t)| \;=\; |\Delta N^{\oplus}(t)| + |\Delta S^{\oplus}(t)|.
\]
Consequently, any energy model proportional to the number of microstate transitions satisfies
\[
C_E(G_N;T)\ \propto\ \sum_{t=0}^{T-1} |\Delta X(t)|
\;=\;
\sum_{t=0}^{T-1}\Big( |\Delta N^{\oplus}(t)| + |\Delta S^{\oplus}(t)| \Big),
\]
i.e., energy is proportional to the discrete-time derivative of the computational graph's global state.
\end{lemma}

\begin{proof}
By Definition~\ref{def:microstate}, the global microstate $X(t)$ is the disjoint union of $N$-owned bits and $S$-owned bits, and a bit belongs to $\Delta X(t)$ if and only if it changes between $t$ and $t+1$.
Partitioning the toggling bits by ownership yields the claimed equality of cardinalities; summing over time gives the cumulative result.
\end{proof}

\begin{corollary}[Bounding event traces by transition-derivative traces]\label{cor:update-trace-bounds-derivative}
Let $\Delta G_N(t)=(\Delta N(t),\Delta S(t))$ denote the activity trace used in Definitions~\ref{def:deltaG}--\ref{def:energyComplexity}, where
$\Delta N(t)$ is the set of neuron-owned components updated/processed at time $t$ and
$\Delta S(t)\subseteq S$ is the set of synaptic transmission events at time $t$.

Assume a fixed digital primitive class such that there exist architecture-dependent constants $c_N,c_S>0$ with:
\begin{enumerate}
\item each processed neuron-owned update in $\Delta N(t)$ induces at most $c_N$ microstate bit flips, and
\item each synaptic event in $\Delta S(t)$ induces at most $c_S$ microstate bit flips (including any routing and buffering state changes represented within the model).
\end{enumerate}
Then for all $t$,
\[
|\Delta X(t)| \;\le\; c_N\,|\Delta N(t)| + c_S\,|\Delta S(t)|.
\]
Therefore,
\[
\sum_{t=0}^{T-1}|\Delta X(t)| \;=\; O\!\left(\sum_{t=0}^{T-1}|\Delta N(t)| + \sum_{t=0}^{T-1}|\Delta S(t)|\right),
\]
so the event-count model of Definition~\ref{def:energyComplexity} forms an upper bound (up to constants) on switching/derivative-based energy model.

If, additionally, each processed neuron update and each synaptic event induces at least one microstate transition (no ``null'' updates), then a matching lower bound holds up to constants and the two traces are $\Theta(\cdot)$-equivalent in aggregate over $T$.
\end{corollary}

This digital refinement clarifies one aspect of mechanisms such as leak and decay in digital NMC systems: if implemented via synchronous ticking of state each step, it forces dense $|\Delta X(t)|$ even when $|\Delta S(t)|$ is sparse, whereas event-driven implementations preserve sparsity by limiting microstate transitions to touched components.

\subsubsection{Relationship of microstate derivative to an algorithm's target function}

For a dynamical system (or ODE) there is an intrinsic notion of derivative in the \emph{target computation}.
Consider
\[
\dot{z}(t)=f(z(t),u(t)),
\]
and a discrete-time simulation (numerical integrator) with step size $\Delta t$,
\[
z_{k+1}=F_{\Delta t}(z_k,u_k)\approx z_k+\Delta t\,f(z_k,u_k).
\]
A digital neuromorphic implementation of this update maintains a global microstate $X(k)\in\{0,1\}^B$,
whose evolution induces a per-step switching activity $a_X(k)=|\Delta X(k)|$ (Definition~\ref{def:microstate}).
If the represented state $\hat z_k$ is encoded in a subset of bits $Y(k)\subseteq X(k)$, then any change
in the represented state necessarily causes microstate switching:
\[
d_H\!\left(Y(k+1),Y(k)\right)\le a_X(k).
\]
For a reasonable encoding, the number of output-bit changes grows with the magnitude of the
step-to-step change in the simulated state, e.g.,
\[
d_H\!\left(Y(k+1),Y(k)\right)\ \lesssim\ L_{\mathrm{enc}}\,\|z_{k+1}-z_k\|
\ \approx\ L_{\mathrm{enc}}\,\Delta t\,\|f(z_k,u_k)\|.
\]
where $L_{enc}$ is an encoding-dependent constant that upper bounds how much the digital representation can change in Hamming distance for a given change in the simulated state. 

Together these relations provide an intuitive link between the derivative of the modeled dynamics and the
minimum switching (and thus energy) required by any digital implementation: when the simulated state
changes rapidly, the represented state must change, and bit transitions are unavoidable. At the same
time, the \emph{total} switching $a_X(k)$ can be much larger than this ``necessary'' amount due to internal
bookkeeping, routing, and auxiliary variables. This motivates a compilation objective for digital
neuromorphic simulation of dynamical systems: preserve the desired update map $F_{\Delta t}$ while
minimizing \emph{toggle amplification} beyond what is required to represent the evolution of $z_k$.

\subsubsection{Digital refinement: how the microstate-derivative relates to activity decay}

The relationship between the microstate-derivative of an algorithm and the execution trace provides more than a refinement on the energy requirements of a NMC algorithm; it also helps us improve on the activity decay metric from the NMC algorithm analysis in section \ref{sec:NMCadvantage}. The baseline trace-based quantities (e.g., $a_S(t)=|\Delta S(t)|$) define activity in terms of which
synaptic channels and neuron components are \emph{active} under the chosen execution semantics. While this is appropriate as a technology-agnostic abstraction, for \emph{digital} neuromorphic systems we can strengthen the decay notion by leveraging the switching-based view from Section~\ref{sec:energyComplexityDigital}: dynamic energy is driven by bit flip \emph{state transitions}, i.e., by the discrete-time derivative
of the algorithm's global digital microstate. This refinement is most important when internal dynamics (e.g., leak or decay) are implemented via explicit state updates, since such mechanisms can induce switching even when spike traffic is sparse. 

\begin{definition}[Microstate-derivative activity process (digital)]\label{def:microstate-derivative-process}
Let $G_N=(N,S)$ be a neuromorphic algorithm executed on a digital substrate, with digital microstate $X(t)\in\{0,1\}^B$ and derivative support $\Delta X(t)$ as in Definition~\ref{def:microstate}.

Define the \emph{microstate-derivative activity process} as
\[
a_X(t) := |\Delta X(t)|.
\]
Define the corresponding \emph{normalized microstate-derivative intensity} over horizon $T$ as
\[
\Phi_X(G_N;T) := \frac{1}{BT}\sum_{t=0}^{T-1}|\Delta X(t)|.
\]
\end{definition}

\begin{definition}[Microstate-derivative-decaying execution]\label{def:microstate-derivative-decay}
A digital neuromorphic execution is \emph{microstate-derivative-decaying} if there exists a time $t_0$
such that for all $t\ge t_0$,
\[
\mathbb{E}[a_X(t+1)] \le \mathbb{E}[a_X(t)],
\]
and there exists a limiting regime in which
\[
\limsup_{t\to\infty}\mathbb{E}[a_X(t)] \le a_\infty,
\]
for some (typically small) constant $a_\infty$.
\end{definition}

\subsubsection{How to extend the framework to analog components with intrinsic dynamics}\label{sec:M4star1}
The formal analysis in Sections~\ref{sec:complexityAnalysis}--\ref{sec:NMCadvantage} intentionally adopts discrete-time, event-count models that are
well aligned with digital neuromorphic platforms and with algorithm-level comparison to conventional
stored-program execution.
This section sketches a \emph{non-normative} extension that incorporates analog components with intrinsic
continuous-time dynamics.
The purpose is not to change the formal scaling results above, but to illustrate how the same
$G_N=(N,S)$ formalism can support more advanced physical energy models in which some computation arises
``for free'' from natural device relaxation while other computation incurs explicit energetic cost to
drive or hold state.

In the core framework, all computation is localized within neuron-owned components
$N = (N_S,N_X,N_Y,N_E)$ and inter-neuron communication is represented by discrete channels $S$.
This representation remains compatible with mixed-signal implementations, but the energy model must be
refined when some neuron-owned components are implemented as analog dynamical elements.
In particular, unlike digital state which is static when left unclocked, an analog state variable may
exhibit intrinsic drift or relaxation toward an equilibrium or an attractor state.

\paragraph{Analog component model.}
We start by considering a specific class of analog devices common to NMC hardware. Let $u$ denote a neuron-owned component (e.g., $u\in N_X$ for membrane-like state or $u\in N_S$ for post-synaptic filtering) that is implemented as an analog dynamical element with continuous-time state
$x_u(t_{\tau})\in\mathbb{R}^{d_u}$, where $t_{\tau}$ is continuous time.
We posit that when the component is left unforced it follows natural and unforced dynamics
\begin{equation}
\dot{x}_u(t_{\tau}) = F_u(x_u(t_{\tau});\theta_u(t_{\tau})), \label{eq:analog_natural_dynamics}
\end{equation}
where $\theta_u(t_{\tau})$ denotes parameters that may be modified by over time (e.g., synaptic conductance or somatic leak).
In many cases $F_u$ has a stable equilibrium $x_u^\star(\theta_u)$ such that $x_u(t_{\tau})\to x_u^\star$ as
$t\to\infty$ under constant $\theta_u$.

To model energetic intervention, we introduce an additive \emph{drive} input $a_u(t_{\tau})$:
\begin{equation}
\dot{x}_u(t_{\tau}) = F_u(x_u(t_{\tau});\theta_u(t_{\tau})) + B_u a_u(t_{\tau}), \label{eq:analog_forced_dynamics}
\end{equation}
where $B_u$ maps drive to state.
The intent is not to imply centralized control; rather $a_u(t_{\tau})$ abstracts any energy-consuming action
required to steer, refresh, or hold the component away from its natural evolution (e.g., bias currents,
refresh pulses, explicit write operations, or sustained amplification). Effectively, $a_u(t_{\tau})$ represents any input that pushes the device away from its natural dynamics.

\paragraph{Events as state deflections or equilibrium shifts.}
Discrete spike events (arriving via $S$) can influence an analog component in two conceptually distinct
ways that are both expressible within the framework:
\begin{enumerate}
\item \emph{State deflection (kick):} an event produces an instantaneous perturbation
$x_u(t_{\tau,k}^+) = x_u(t_{\tau,k}^-) + \Delta x_u$, after which the component relaxes according to
\eqref{eq:analog_natural_dynamics} with $a_u(t_{\tau})\equiv 0$.
\item \emph{Equilibrium shift (parameter update):} an event changes $\theta_u(t_{\tau})$ (e.g., synaptic
conductance or somatic leak), thereby changing the natural vector field $F_u$ and the equilibrium
$x_u^\star(\theta_u)$, after which the state relaxes to the new equilibrium under $a_u(t_{\tau})\equiv 0$.
\end{enumerate}
In both cases, a substantial portion of computation may occur through the intrinsic dynamics of the
analog component itself (``free'' in the sense of not requiring explicit drive), though the event that
caused the kick or parameter update may have a nonzero energetic cost.

\paragraph{Residual (paid) derivative and analog activity trace.}
The digital refinement above related energy to the discrete-time derivative of a binary microstate.
For analog components, a useful analogue is the \emph{residual derivative}---the deviation from its natural
evolution:
\begin{equation}
r_u(t_{\tau}) := \dot{x}_u(t_{\tau}) - F_u(x_u(t_{\tau});\theta_u(t_{\tau})). \label{eq:analog_residual}
\end{equation}
Under \eqref{eq:analog_forced_dynamics}, $r_u(t_{\tau})=B_u a_u(t_{\tau})$.
Thus, $r_u(t_{\tau})\equiv 0$ corresponds to letting the device do what it naturally does (relax, integrate,
filter), whereas $r_u(t_{\tau})\neq 0$ corresponds to energy-consuming intervention to steer or hold state.

This motivates an analog extension of trace-based activity in which we distinguish:
(i) \emph{event activity} (spike deliveries and discrete parameter writes) and
(ii) \emph{drive activity} (time intervals during which $r_u(t_{\tau})$ is non-negligible).
For a threshold $\varepsilon>0$, define the set of analog components requiring drive at time $t$ as
\begin{equation}
\Delta N_{\mathrm{drive}}(t_{\tau}) := \{u : \|r_u(t_{\tau})\| > \varepsilon\}. \label{eq:drive_activity_set}
\end{equation}
Then the analog analogue of the event-count model is that energy is governed by a combination of
discrete events and the duration/extent of drive:
\begin{equation}
C_E^{\mathrm{analog}} \;\approx\;
\underbrace{\sum_{k\in \mathcal{E}} e_k}_{\text{discrete event costs}}
\;+\;
\underbrace{\int_0^T \sum_{u\in N} \rho_u(r_u(t_{\tau}))\,dt}_{\text{drive/holding costs}},
\label{eq:analog_energy_functional}
\end{equation}
where $\mathcal{E}$ denotes the set of discrete events (e.g., spike transmissions, parameter writes) and
$\rho_u(\cdot)$ is a nonnegative per-component dissipation/effort functional (architecture dependent).
Equation~\eqref{eq:analog_energy_functional} is consistent with the trace-based view of Section~\ref{sec:complexityAnalysis}:
energy is still execution-dependent, but ``activity'' is now a functional of continuous-time trajectories,
not only the cardinality of event sets.

Notably, the digital switching model above is consistent with this analog extension when the microstate is finite and static under
no events (no intrinsic drift), i.e., when $F_u\equiv 0$ and energy is dominated by discrete transitions.
In that case, the residual derivative reduces to the usual discrete-time microstate derivative used in
Section~\ref{sec:energyComplexityDigital}.

\subsubsection{Implications for algorithm metrics under analog realizations}\label{sec:M4star2}

The structural metrics in Section~\ref{sec:NMCadvantage} are defined on the instantiated graph $G_N=(N,S)$ and therefore do
not fundamentally change with analog; however, the \emph{trace-based} metrics require reinterpretation because energy is not determined solely by spike events when intrinsic analog dynamics and holding costs are present.

\paragraph{Trace intensity.}
For hybrid mixed-signal implementations in which inter-neuron communication remains event-based (e.g., AER),
the synaptic activity intensity $\Phi_S(G_N;T)$ remains a first-order predictor of the \emph{discrete event}
portion of energy.
In addition, analog components introduce a complementary intensity associated with drive/holding:
\begin{equation}
\Phi_{\mathrm{drive}} := \frac{1}{|N|\,T}\int_0^T |\Delta N_{\mathrm{drive}}(t_{\tau})|\,dt_{\tau},
\label{eq:drive_intensity}
\end{equation}
or, more generally, an amplitude-weighted version
$\int_0^T \sum_u \rho_u(r_u(t_{\tau}))\,dt$ (cf.~\eqref{eq:analog_energy_functional}).
Algorithms that rely on long-lived deviations from equilibrium (``holding information in analog state'')
may have low $\Phi_S$ but high $\Phi_{\mathrm{drive}}$, whereas relaxation-based algorithms may have modest
event rates but low drive costs.

\paragraph{Activity decay and convergence.}
The decay notion in Section~\ref{sec:NMCadvantage} becomes two-dimensional in analog settings:
\begin{itemize}
\item \emph{Event decay:} $|\Delta S(t_{\tau})|$ decreases (as before).
\item \emph{Drive decay:} $|\Delta N_{\mathrm{drive}}(t_{\tau})|$ and/or $\sum_u \rho_u(r_u(t_{\tau}))$ decreases.
\end{itemize}
In particular, analog ``free computation'' corresponds to regimes where useful computation is embedded in
the natural relaxation $F_u$ and drive is needed only transiently (small $\Phi_{\mathrm{drive}}$ and
drive decay). Conversely, mechanisms that require maintaining state far from equilibrium for long durations introduce
holding costs that may dominate energy even if spike traffic is sparse.

\paragraph{Variability.}
Trace variability measures (e.g., $CV_S$) remain meaningful for event activity, but in analog components input-dependent variability can also arise through amplitude and time-constant differences that affect $\rho_u(r_u(t_{\tau}))$.
Thus, a complete analog analysis may require a variability measure for drive energy
(e.g., the coefficient of variation of $\int \sum_u \rho_u(r_u(t_{\tau})) dt$), not only event counts.

\subsubsection{Digital vs.\ analog separation perspectives}\label{sec:M4star3}

A general, algorithm-independent separation theorem between digital and analog neuromorphic energy models
is difficult without committing to a device-level dissipation model (i.e., specifying the functionals
$\rho_u$ and the discrete event costs in \eqref{eq:analog_energy_functional}).
Nevertheless, the framework suggests conditional separations that clarify where analog intrinsic dynamics
\emph{could} change scaling relative to purely digital switching models.

Consider a class of algorithms whose neuromorphic formulation is based on relaxation to equilibrium (or
fixed points), where events serve primarily to initialize or perturb state and then the system is allowed
to evolve under natural dynamics. If the digital realization of the same algorithm requires explicit state updates each tick (e.g., synchronous
leak/decay or iterative relaxation implemented by repeated digital updates), then its switching derivative
is $\Theta(BT)$ in the worst case. In contrast, an analog realization with negligible holding/drive cost after initialization may have energy
dominated by a number of discrete events that is sublinear in $T$ (e.g., $O(B)$ or $O(B\log T)$ depending on
how perturbations are applied), yielding an asymptotic advantage \emph{conditional} on low baseline dissipation.

Conversely, for algorithm classes that require maintaining analog state far from equilibrium for a duration
proportional to $T$ (e.g., representing memory by sustained deviation), analog holding costs may scale as
$\Omega(T)$ even when event activity is sparse. In such regimes, a digital implementation using static memory (with near-zero holding cost) may be
energetically preferable, illustrating that analog advantages depend on whether the computation can be
embedded in natural dynamics rather than in sustained forcing.

In short, the present framework can accommodate analog intrinsic dynamics by replacing the digital
microstate-derivative activity with a residual-to-natural-flow notion of activity and an energy functional
that includes both discrete event costs and continuous drive/holding costs.
This suggests that analog neuromorphic substrates are most promising for algorithm families that can
express useful computation through relaxation and natural dynamics with minimal sustained drive, whereas
purely event-count interpretations may be misleading when holding costs or baseline dissipation dominate.

\newpage
\subsection{Framework Validation}
\label{sec:FrameworkValidation}

We next look to validate the suitability of our framework against previously published neuromorphic applications and broader algorithm classes. As we are treating neuromorphic architectures as a type of parallel architecture, the question is which algorithm classes, if any, are preferentially well-suited for neuromorphic compared to alternative parallel SIMD architectures like GPUs.

\subsubsection{Algorithm classes through the two-landscape view}
\label{sec:class_landscape_mapping}

Before mapping specific published neuromorphic algorithms, it is useful to position broad algorithm classes in the graph-based and trace-based landscapes of Section~\ref{sec:NMCadvantage}. Tables~\ref{tab:landscape_structural} and \ref{tab:landscape_trace} provide a qualitative summary. The intent is not to assign a single scalar ``score'' to each class, but to highlight which structural and trace properties dominate time/space/energy outcomes across architectures.

\hyphenpenalty=10000
\exhyphenpenalty=10000

\begin{table*}[t]
\centering
\caption{Graph-based (structure-dependent) landscape coordinates for representative algorithm classes discussed in Section~M4.2 (and the dense ANN example in Section~5.2), sorted roughly from more neuromorphic-aligned to less neuromorphic-aligned based on recurrence/reuse and communication structure.}
\label{tab:landscape_structural}
\renewcommand{\arraystretch}{1.15}
\setlength{\tabcolsep}{6pt}
\small
\begin{tabularx}{\textwidth}{C{4.8cm} C{2.4cm} C{1.7cm} C{5.9cm}}
\toprule
\textbf{Algorithm class} &
\textbf{Within-step homogeneity \(H\)} &
\textbf{Reuse \(R_T\)} &
\textbf{Communication structure (fan-out/locality)} \\
\midrule

Iterative solvers&
Low--moderate &
High &
Sparse; mesh-induced locality; bounded degree common \\
\addlinespace

Wavefront / dynamic programming propagation  &
Low--moderate &
High &
Sparse; often local / bounded degree \\
\addlinespace

DTMC random walks / Monte Carlo sampling &
Low--moderate &
High &
Bounded degree (local transitions); event-proportional communication \\
\addlinespace

Message passing graph algorithms  &
Low--moderate &
High &
Bounded degree (pre-defined input graph); locality / embedding dependent \\
\addlinespace

Event-based vision pipelines &
Moderate &
High &
Local receptive fields; bounded fan-out at fixed kernel size \\
\addlinespace

Reservoir computing / liquid state machines &
Low--moderate &
High &
Typically sparse recurrent connectivity; locality depends on design \\
\addlinespace

State-space sequence models &
Moderate &
High &
Structured recurrence; bounded degree if designed/sparse \\
\addlinespace

Dense linear algebra  &
High &
Low &
Dense / global connectivity; \(|S|=\Theta(N_iN_j)\) \\
\bottomrule
\end{tabularx}
\end{table*}

\begin{table*}[t]
\centering
\caption{Trace-based (activity-dependent) landscape coordinates for the same algorithm classes as in Table~\ref{tab:landscape_structural}. These properties depend on inputs, operating regime, and execution semantics.}
\label{tab:landscape_trace}
\renewcommand{\arraystretch}{1.15}
\setlength{\tabcolsep}{6pt}
\small
\begin{tabularx}{\textwidth}{C{4.8cm} C{2.8cm} C{2.8cm} C{4.2cm}}
\toprule
\textbf{Algorithm class} &
\textbf{Activity intensity \(\Phi_S\)} &
\textbf{Trajectory (decay / convergence)} &
\textbf{Variability \(\mathrm{CV}_S\)} \\
\midrule

Iterative solvers &
Low--moderate&
Often decaying (with convergence) &
Moderate \\
\addlinespace

Wavefront / dynamic programming propagation  &
Low &
Often decaying (with termination) &
High (instance-dependent) \\
\addlinespace

DTMC random walks / Monte Carlo sampling &
Low &
Not necessarily decaying &
High (stochastic / input-dependent) \\
\addlinespace

Message passing graph algorithms  &
Regime dependent &
Often terminating (model-dependent) &
Moderate--high (instance-dependent) \\
\addlinespace

Event-based vision pipelines  &
Low &
Scene-dependent; often bursty &
High (input-dependent) \\
\addlinespace

Reservoir computing / liquid state machines &
Moderate &
Often stationary (by design) &
Moderate--high (input-dependent) \\
\addlinespace

State-space sequence models &
Low--moderate &
Sometimes (if gated/selective) &
Moderate \\
\addlinespace

Dense linear algebra &
High &
Stationary &
Low \\
\bottomrule
\end{tabularx}
\end{table*}

GPU-friendly workloads are typically characterized by large within-step homogeneity (high $H$; i.e., many isomorphic operations per time step), while neuromorphic-friendly workloads are typically characterized by high reuse (high $R_T$; i.e., recurrence) and bounded-fan out and local communication structure (low $K$). Similarly, from an activity perspective, algorithms with low activity intensity (low $Phi_S$), significant activity decay ($a_S(t) \rightarrow 0$), and / or activity variability (high $CV(G_N)$) will be more neuromorphic friendly, whereas such algorithms may not see comparable benefits (and may even be detrimental, in the case of high $CV(G)$, on GPUs.

\subsubsection{Framework validation via representative algorithm mappings}

The definitions and scaling results developed above are meant to be generic with respect to neuron and synapse models while still being sufficiently structured to support comparative complexity arguments. To validate that the framework is neither overly restrictive nor vacuous, we map several representative neuromorphic algorithm families from the literature into the neuromorphic algorithm structure ($G_N=(N,S)$) and summarize what each contributes to the theory. These results are expanded on in detail in the Appendix.

\begin{itemize}
    \item \textbf{Iterative solvers (activity-driven convergence and structural reuse).}
    Iterative solvers, particularly those that target sparse linear systems ($\textit{A}\textbf{x} = \textbf{b}$), check many of the neuromorphic boxes. A recent example known as NeuroFEM \cite{theilman2025solving} instantiates a fixed recurrent spiking graph that solves a sparse linear system by iterative convergence. When mapped at the granular level (one spiking unit per neuron subgraph), NeuroFEM directly realizes the framework assumptions that (i) inter-neuron communication occurs only via discrete spike events on $S$, (ii) all synaptic dynamics (fast/slow integration) and solver dynamics (PI-like internal state) are post-synaptic and neuron-local (captured within $n_S$ and $n_X$), and (iii) recurrence enables structural reuse without unrolling iterations into additional graph resources. This mapping provides concrete support for Section \ref{sec:energy}'s event-count energy model and Section \ref{sec:NMCadvantage}'s claim that iterative, mesh-/sparsity-structured computations can exhibit energy scaling governed by cumulative activity $\sum_t|\Delta S(t)|$ rather than by the static graph size $|S|$.

    \item \textbf{Distributed Monte Carlo sampling (localized computation and event-proportional work).}
    Smith et al. \cite{smith2022neuromorphic, severa2016spiking} implement discrete-time Markov chain (DTMC) random walks by representing walkers as spikes and realizing each DTMC state as a small spiking circuit that counts incoming walkers and stochastically routes them to outgoing neighbors. This algorithm family validates the framework's separation of neuron-owned computation $N$ from inter-neuron communication $S$, and further motivates treating energy as a functional of the execution trace (Section \ref{sec:energy}). Specifically, the algorithm's energy use is naturally proportional to the number of walker spike events rather than to the total size of the state graph. In addition, this mapping highlights that execution semantics (e.g., synchronous DTMC stepping versus asynchronous random walk dynamics) can materially change the activity sets $\Delta N(t)$ and $\Delta S(t)$ without changing the underlying structural graph, reinforcing the need to distinguish structural sparsity ($|S|$) from activity sparsity ($|\Delta S(t)|$).

    \item \textbf{Reservoir computing (high structural reuse).}
    Reservoir computing (e.g., liquid state machines and echo state networks) is a canonical recurrent computation paradigm \cite{maass2002real, jaeger2001echo, jaeger2023toward} in which a fixed recurrent ``reservoir'' is driven by a streaming input and only a low-dimensional readout is trained. In our framework, the reservoir corresponds to a fixed instantiated recurrent graph $G_N=(N,S)$ executed over a long horizon $T$, yielding high structural reuse \(R_T(G_N)\) by construction. Reservoir connectivity is often sparse and local (or designed to be sparse), implying bounded or slowly growing fan-out in \(S\). Unlike convergent iterative solvers, however, reservoir dynamics are typically designed to remain active (near a ``critical'' regime) rather than to converge to a quiescent state; consequently the activity trajectory \(a_S(t)=|\Delta S(t)|\) need not decay and may be approximately stationary. This exemplar therefore clarifies an important distinction made by the framework: recurrence (high \(R_T\)) is structurally important for neuromorphic alignment, but an energy advantage depends on the realized trace regime. As such, reservoir computing is most likely to be energy-favorable when inputs are sparse/event-driven and the operating point yields low \(\Phi_S\), whereas dense high-rate reservoirs may not exhibit activity-sparsity advantages despite their structural recurrence.
    
    \item \textbf{Event-based vision (bounded fan-out with sparse and variable activity).}
    Event-camera processing pipelines (e.g., event-based object classification, tracking, or motion estimation; see, \cite{benosman2012asynchronous, benosman2013event, orchard2021efficient} and related event-vision literature) provide a natural counterpoint to the claim that ``deep networks are not neuromorphic-friendly.'' In the present framework, an event-vision pipeline is represented as a neuromorphic graph \(G_N=(N,S)\) whose connectivity is typically local (e.g., convolutional receptive fields or locally connected spatiotemporal filters), implying bounded fan-out/degree in \(S\) at fixed kernel size. The key distinction is trace-based: event cameras emit sparse, asynchronous updates, so the induced activity trace \(\Delta G_N(t)=(\Delta N(t),\Delta S(t))\) can be highly sparse in typical scenes, yielding low activity intensity \(\Phi_S(G_N;T)\ll 1\) even for multi-layer networks. Moreover, activity is highly input dependent, leading to large variability \(\mathrm{CV}_S\) across scenes and motion regimes. This exemplar supports the framework prediction that neural network computations can be neuromorphic-aligned when (i) input is event-driven, (ii) the network preserves sparsity (avoids dense fully-active layers), and (iii) computation is streaming/recurrent (high effective reuse over time).

    \item \textbf{Threshold gate simulation (fan-in limitations).}
    Parekh et al. \cite{parekh2018constant} construct low-depth threshold circuits for matrix multiplication under a $TC^0$ model that assumes high fan-in threshold gates and bounded precision. Mapping each threshold gate to a neuron subgraph yields a feed-forward neuromorphic graph $G_N^\ast$, and serves as a concrete instance of the ``neuromorphic can simulate conventional circuits'' direction of the theory. At the same time, it makes the architectural caveats explicit: the depth advantages in $TC^0$ rely on primitive power (fan-in and precision) and on wiring assumptions that may not hold on realizable neuromorphic substrates. In other words, these constructions are not a generic recipe for neuromorphic advantage; they are a reminder that depth/work results always come with assumptions about primitives and interconnect.

    \item \textbf{Dynamic programming (spike-time encoding with sparse and variable activity).}
    Aimone et al. \cite{aimone2019dynamic} map broad classes of dynamic programs (DPs) to spiking circuits by representing subproblems as neuron subgraphs and dependencies as synaptic communication channels. In our framework this fits naturally as $G_N=(N,S)$, and it highlights a capability that is awkward in many conventional formulations: in some DP families, solution values can be encoded in timing (e.g., time-of-first-spike) rather than in explicit multi-bit state variables. This motivates allowing delay-augmented communication while preserving discrete spike events as the primitive.

    From an energy perspective, many DP and shortest-path constructions induce \emph{wavefront} traces: activity propagates outward from an initial condition and often terminates once a goal condition is reached. In those regimes, $\Delta S(t)$ is concentrated near a moving frontier rather than being spread uniformly over $S$. Under bounded fan-out/local DP dependency graphs, $|\Delta S(t)|$ can scale with the frontier size rather than with $|S|$, yielding $\Phi_S(G_N;T)\ll 1$ even though the full dependency structure is instantiated. Because frontier evolution is instance dependent, these executions also tend to have high variability in cumulative activity (large $\mathrm{CV}_S$), which is natural for event-driven execution but can induce divergence and overhead in lockstep accelerators.

    \item \textbf{Message passing graph algorithms (local processing with activity sparsity and variability).}
    Aimone et al.~\cite{aimone2020provable,aimone2021provable} develop spiking formulations of shortest-path problems and analyze them under explicit conventional baselines that account for data movement. In our terms, these algorithms instantiate a $G_N=(N,S)$ in which each input-graph vertex maps to a constant-size neuron subgraph and each input-graph edge maps to one or more spike channels. They also make two practical refinements in our abstraction concrete: (i) multi-bit messages of width $\lambda$ correspond to spike bundles, i.e., $\Theta(\lambda)$ binary channels per logical edge (a direct cost in $|S|$ and potentially in $|\Delta S(t)|$), and (ii) embedding the logical graph into restricted interconnect topologies can dominate realized costs, motivating locality-aware refinements beyond the base event-count model. These works also reinforce the separation in emphasis between the present paper and the ``provable advantage'' line of results: our framework gives conditional structure/trace-based regimes, while stronger separations require adopting explicit conventional data-movement models.

    \item \textbf{State space models (recurrence with potentially sparse activity).}
    State-space sequence models \cite{voelker2019legendre, gu2024mamba} provide an ML-motivated exemplar that is recurrent by construction and therefore naturally yields high reuse $R_T(G_N)$ over long sequences: a fixed circuit runs for many token steps rather than being unrolled into a large feed-forward graph. In our terms, these models correspond to bounded-depth per-step updates of neuron-local state $n_X$ driven by structured recurrent communication in $S$ (the state transition) and by input-dependent gating/modulation. The same structural distinction appears here as in the rest of the paper: if the transition structure is bounded-degree and the computation is selectively gated so that $|\Delta S(t)|$ and $|\Delta N(t)|$ scale with event rate rather than with $|S|$, then the event-count model predicts potentially favorable energy scaling. If, instead, token processing induces sustained dense activation each step (a constant fraction of edges active), then energy scales as $\Omega(T|S|)$ up to constants and neuromorphic advantage is unlikely without additional sparsification or approximation.
\end{itemize}

\section*{Acknowledgments}

The author thanks Ojas Parekh, Darby Smith, Brad Theilman, and Craig Vineyard for comments and discussions. The DOE Advanced Simulation and Computing program provided support for this work. This article has been authored by an employee of National Technology \& Engineering Solutions of Sandia, LLC under Contract No. DE-NA0003525 with the U.S. Department of Energy (DOE). The employee owns all right, title and interest in and to the article and is solely responsible for its contents. The United States Government retains and the publisher, by accepting the article for publication, acknowledges that the United States Government retains a non-exclusive, paid-up, irrevocable, world-wide license to publish or reproduce the published form of this article or allow others to do so, for United States Government purposes. The DOE will provide public access to these results of federally sponsored research in accordance with the DOE Public Access Plan https://www.energy.gov/downloads/doe-public-access-plan .

\bibliographystyle{IEEEtran}


\end{document}